\documentclass{article}

\usepackage{arxiv}
\usepackage[utf8]{inputenc} 
\usepackage[T1]{fontenc}    
\usepackage{hyperref}       
\usepackage{url} 
\usepackage{amssymb}
\usepackage{latexsym}
\usepackage{amsmath}
\usepackage{enumitem}
\usepackage{cite}
\usepackage{amsthm}

\newtheorem*{remark*}{Remark}
 
\usepackage{multirow}
\usepackage{bm}

\usepackage{booktabs}      
\usepackage{amsfonts}      
\usepackage{nicefrac}    
\usepackage{microtype}     
\usepackage{lipsum}
\usepackage{graphicx}
\usepackage{subcaption}
\usepackage{xcolor}
\usepackage{algorithm}
\usepackage{algorithmic}
\graphicspath{ {./images/} }

\title{Convolution-weighting method for the physics-informed neural network: A Primal-Dual Optimization Perspective}

\author{
  Chenhao Si \\
  School of Data Science\\
  The Chinese University of Hong Kong, Shenzhen\\
  Shenzhen, China  \\
  \texttt{222042011@link.cuhk.edu.cn} \\
  \And
  Ming Yan* \\
  School of Data Science\\
  The Chinese University of Hong Kong, Shenzhen \\
  Shenzhen, China  \\
  \texttt{yanming@cuhk.edu.cn} \\
}

\begin{document}
\maketitle
\begin{abstract}
Physics-informed neural networks (PINNs) are extensively employed to solve partial differential equations (PDEs) by ensuring that the outputs and gradients of deep learning models adhere to the governing equations. However, constrained by computational limitations, PINNs are typically optimized using a finite set of points, which poses significant challenges in guaranteeing their convergence and accuracy. In this study, we proposed a new weighting scheme that will adaptively change the weights to the loss functions from isolated points to their continuous neighborhood regions. The empirical results show that our weighting scheme can reduce the relative $L^2$ errors to a lower value.
\end{abstract}


\section{Introduction}
\label{Introduction}
Physics-Informed Neural Networks (PINNs)~\cite{raissi2019physics, karniadakis2021physics} have established themselves as a universal framework for integrating physical laws with neural networks. By encoding governing Partial Differential Equations (PDEs) into neural network loss functions through automatic differentiation, PINNs enable solving both forward and inverse problems without requiring dense observational data. This paradigm leverages two synergistic components: the expressivity of neural networks to approximate complex solution spaces, and the regularization provided by PDE residuals and boundary/initial conditions. Its effectiveness has been demonstrated across diverse domains, from heat transfer~\cite{xu2023physics,cai2021physics,majumdar2025hxpinn} and solid mechanics~\cite{hu2024physics,faroughi2024physics} to stochastic systems~\cite{zhang2020learning,chen2021solving} and uncertainty quantification~\cite{yang2019adversarial,zhang2019quantifying,yang2021b,tan2024physics,tan2025plug}. 

The PDE-involved loss formulation remains the cornerstone of PINNs, whether in canonical PINNs~\cite{raissi2019physics, karniadakis2021physics} or their recent extensions through transformers~\cite{zhao2023pinnsformer}, diffusion models~\cite{huang2024diffusionpde,bastek2024physics}, and operator learning frameworks~\cite{lu2019deeponet,xu2025velocity,gao2024coordinate,cao2024laplace}. While these advancements demonstrate improved handling of stochastic systems and high-dimensional parameter spaces, they universally confront a fundamental challenge: the inherent imbalance between competing loss components—governing equations, boundary/initial conditions, and optional observational data. This imbalance manifests as gradient pathologies~\cite{wang2021understanding} during optimization, where dominant loss terms dictate parameter updates while critical physical constraints remain under-enforced~\cite{RBA-PINN}.

Recent advances in PINN training strategies primarily address loss imbalance through two complementary approaches. The first involves adaptive spatiotemporal resampling of collocation points, where regions exhibiting sharp gradients or high residuals—such as shock interfaces~\cite{abbasi2025challenges,liu2024discontinuity} or turbulent boundary layers~\cite{arzani2023theory,bararnia2022application}—are progressively allocated denser sampling points. For example, Lu et al.~\cite{lu2021deepxde} developed a threshold-dependent method for selecting new training points, whereas Wu et al.~\cite{wu2023comprehensive} devised a probability density function derived from residuals to enhance sampling efficiency. Moreover, building upon Wu et al.’s methodology, Gao et al.~\cite{gao2023failure} define a “failure region” within PINN and perform resampling within this region. Gao et al.~\cite{gao2023active} further extend this line of inquiry by employing active learning for resampling, with a particular focus on high-dimensional cases.

The second and more widely studied approach focuses on dynamic loss weighting, initially inspired by multi-task learning frameworks that assign scalar weights to distinct loss components~\cite{bischof2025multi} (e.g., PDE residuals versus boundary conditions and initial conditions). For example, Wang et al.~\cite{wang2021understanding} uses an learning rate annealing strategy to balance the loss functions during the training. Wang et al.~\cite{wang2022NTK} utilized the Neural Tangent Kernel (NTK) to analyze the dynamics of gradients during the training of PINNs and leveraged insights from NTK to strategically tune the weights assigned to each loss component. Xiang et al.~\cite{xiang2022self} adjusted the weights via Gaussian likelihood estimation, whereas Hua et al.~\cite{hua2023physics} employed the reciprocal of the loss variance during training as the weighting scheme. However, such component-level weighting proves insufficient for PINNs due to two inherent limitations: (1) spatially heterogeneous solution features necessitate point-wise rather than global weight adjustments within individual loss terms, as optimal weights for stiff regions differ drastically from smooth domains~\cite{song2024loss,LUO2025114010}; (2) hard-constrained formulations that embed boundary conditions via network architectures collapse all physical constraints into a single residual loss, rendering component-level weighting inapplicable~\cite{lu2021physics,sukumar2022exact,dong2021method}.

These limitations have spurred the development of point-wise adaptive weighting schemes. Pioneering work by McClenny et al.~\cite{SA-PINN} assigns trainable weights to individual training points, allowing the network to autonomously focus on difficult regions. This is achieved via a soft attention mask mechanism, where weights increase with loss magnitude, trained using gradient descent/ascent to minimize losses while maximizing weights. Moreover, Song et al.~\cite{song2024loss} used LANs as adversarial networks for weight training to achieve adaptive weighting of point errors and improve PINN performance. Furthermore, besides utilizing additional training parameters or axillary networks, Anagnostopoulos et al.~\cite{RBA-PINN} proposed a residual-based attention (RBA) scheme that updates residual weights according to the absolute magnitude of residuals. The scheme's efficiency lies in its gradient-free design, imposing negligible computational overhead while dynamically focusing on problematic regions. 

Fundamental limitations persist in existing point-wise weighting schemes due to their discrete treatment of inherently continuous PDE systems. While current methods successfully prioritize individual collocation points with high residuals~\cite{SA-PINN,song2024loss,RBA-PINN}, they neglect the spatial correlation inherent in PDE solutions—a critical oversight given that local stiffness regions inherently influence neighboring domains~\cite{wu2024ropinn}. This atomized weighting approach risks overfitting to isolated points while failing to ensure global solution consistency. Furthermore, the decoupled development of weighting and resampling strategies creates suboptimal synergies; adaptive sampling identifies critical regions but lacks mechanisms to propagate this spatial awareness into loss balancing. 

To bridge the gap between discrete weighting heuristics and continuous PDE dynamics, we propose a convolution-weighting framework that synergistically unifies spatial resampling with physics-aware weight regularization. By convolving point-wise residuals with desired kernels, our method transforms isolated weight adjustments into spatially coherent regularization operators, effectively addressing the paradox of applying discrete optimization to continuum physical systems. The main contributions of this work are threefold:
\begin{itemize}
    \item \textbf{Optimization-Based Analysis of PINN Weighting Mechanisms}: We reinterpret adaptive weighting through a min-max optimization lens, providing mechanistic insights into the basic reasoning of the weighting schemes for PINNs. This analysis motivates our spatially correlated weighting strategy as a necessary remedy for the discrete-continuum mismatch in PINN training.
    \item \textbf{Continuum Weighting via Convolutional Operators}: Departing from point-wise weight updates that disregard solution smoothness, we develop convolutional weighting. This innovation ensures weight fields inherit the PDE’s intrinsic continuity, suppressing oscillatory artifacts while maintaining compatibility with adaptive resampling. 
    \item \textbf{Integrated Training-Weighting-Resampling Architecture}: We create an end-to-end trainable architecture that dynamically couples three traditionally isolated components: (i) neural PDE solver, (ii) physics-informed convolutional weighting, and (iii) residual-driven resampling.
\end{itemize}

The organization of the rest of the paper is as follows. We review the PINN framework in Section~\ref{PINN}, followed by introducing our methods in Section~\ref{Motivation} starting with a primal-dual optimization problem, then outlining the modified problem incorporating convolution techniques and showing its relation to the weighting methods in PINN training as well as the rationale behind the need for the resampling strategy. The empirical results of our method across various PDEs compared with baseline models~\cite{SA-PINN,song2024loss,RBA-PINN} are presented in Section~\ref{Experiment}. Finally, Section~\ref{Discussion} summarizes the findings and discusses our future work.

\section{Physics-Informed Neural Network}
\label{PINN}
Denote the spatial domain as $\Omega \subset \mathbb{R}^n$ with boundary $\partial \Omega$, and let $T$ represent the time domain. The spatial-temporal variable is given by $(\mathbf{x}, t) \in \Omega \times T$. A time-dependent partial differential equation (PDE) over this domain is defined as follows:
\begin{subequations}
\begin{align}
    \mathcal{F}[u](\mathbf{x}, t) &= 0, \label{(1)} \quad (\mathbf{x}, t) \in \Omega \times T, \\
    \mathcal{B}[u](\mathbf{x}, t) &= 0, \label{(2)} \quad (\mathbf{x}, t) \in \partial \Omega \times T, \quad \text{(boundary condition)} \\
    \mathcal{I}[u](\mathbf{x}, 0) &= 0, \quad \mathbf{x} \in \Omega, \hspace{51pt} \text{(initial condition)}
\end{align}
\end{subequations}
where $\mathcal{F}$, $\mathcal{B}$, and $\mathcal{I}$ are differential operators, and $u(\mathbf{x}, t)$ is the solution to the PDE, subject to boundary and initial conditions.

A PINN parameterized by $\theta$ approximates the solution $u(\mathbf{x}, t)$. The input to the neural network is $(\mathbf{x}, t)$, and the approximation is denoted by $\hat{u}(\theta)(\mathbf{x}, t)$. The PINN minimizes the following objective function:
\begin{align}
    \mathcal{L}(\theta) = \lambda_F \mathcal{L}_F(\theta) + \lambda_B \mathcal{L}_B(\theta) + \lambda_I \mathcal{L}_I(\theta), \label{PINN loss}
\end{align}
where
\begin{align}
    \mathcal{L}_F(\theta) &= \frac{1}{N_f} \sum_{(\mathbf{x}, t) \in \Omega_F} \big| \mathcal{F}[\hat{u}(\theta)](\mathbf{x}, t) \big|^2, \label{residual loss} \\
    \mathcal{L}_B(\theta) &= \frac{1}{N_b} \sum_{(\mathbf{x}, t) \in \Omega_B} \big| \mathcal{B}[\hat{u}(\theta)](\mathbf{x}, t) \big|^2, \label{boundary loss} \\
    \mathcal{L}_I(\theta) &= \frac{1}{N_0} \sum_{(\mathbf{x}, 0) \in \Omega_I} \big| \mathcal{I}[\hat{u}(\theta)](\mathbf{x}, 0) \big|^2. \label{initial loss}
\end{align}
Here, $\Omega_F$, $\Omega_B$, and $\Omega_I$ are the training sets for the PDE residual, boundary condition, and initial condition, respectively, with cardinalities $N_f$, $N_b$, and $N_0$. The weights $\lambda_F$, $\lambda_B$, and $\lambda_I$ are hyperparameters tuning the contributions of each loss component. Notably, $\Omega_F$ may include points on the boundary or at the initial time, allowing $\Omega_F \cap \Omega_B$ and $\Omega_F \cap \Omega_I$ to be non-empty.

\subsection{Inverse Problem}
In the inverse problem, we aim to infer unknown parameters $\lambda$ of the PDE system (e.g., material properties, source terms, or diffusion coefficients) using observed data from the solution field $u(\mathbf{x}, t)$. The PDE operators $\mathcal{F}$, $\mathcal{B}$, and $\mathcal{I}$:
\begin{subequations}
\begin{align}
    \mathcal{F}[u;\lambda](\mathbf{x}, t) &= 0,  \quad (\mathbf{x}, t) \in \Omega \times T, \\
    \mathcal{B}[u;\lambda](\mathbf{x}, t) &= 0,  \quad (\mathbf{x}, t) \in \partial \Omega \times T, \quad \text{(boundary condition)} \\
    \mathcal{I}[u;\lambda](\mathbf{x}, 0) &= 0, \quad \mathbf{x} \in \Omega, \hspace{51pt} \text{(initial condition)}
\end{align}
\end{subequations}
where we should note that the unknown parameter $\lambda$ can either be a scaler or a function. Moreover, in many cases, we do not have the exact forms of the boundary and initial conditions, rather, we have the observed data $u_{\text{obs}}(\mathbf{x}_{\text{obs}}, t_{\text{obs}})$ where $(\mathbf{x}_{\text{obs}}, t_{\text{obs}})\in \Omega$ is not necessary on the boundaries or at $t=0$. 

In this case, the objective function is modified into:
\begin{align}
    \mathcal{L}(\theta) = \lambda_F \mathcal{L}_F(\theta) + \lambda_{\text{obs}} \mathcal{L}_{\text{obs}}(\theta), \label{PINN inverse loss}
\end{align}
where
\begin{align}
    \mathcal{L}_{\text{obs}}(\theta) &= \frac{1}{N_{\text{obs}}} \sum_{(\mathbf{x}, t) \in \Omega_{\text{obs}}} \big|\hat{u}(\theta)(\mathbf{x}, t) - u_{\text{obs}}(\mathbf{x}, t)\big|^2. \label{inverse data loss}
\end{align}
$\Omega_{\text{obs}}$ is the training set of the observed data and $\mathcal{L}_F(\theta)$ is the same as Eq.~\eqref{residual loss}. We would like to note that, in the inverse case, measuring the observed data is expensive and prone to noise; therefore, $N_{\text{obs}}$ is usually a small number in practice.

\section{Convolution-weighting method}
\label{Motivation}
Effective training of PINNs is crucial for the network's ability to accurately capture the solution's behavior across the entire domain. One significant challenge arises from the way the loss function is traditionally computed—the mean squared residual evaluated at a set of collocation points. This averaging process can obscure large residuals at specific points or regions, causing the optimizer to overlook areas where the PDE constraints are poorly satisfied. As a result, despite an overall reduction in loss during training, the network may not learn critical features of the solution, particularly those associated with complex spatial or temporal variations. One idea is to assign different weights to each residual training point. In other words, the residual loss in Eq.~\eqref{residual loss} becomes:
\begin{align}
    \mathcal{L}_F(\theta) &= \sum_{(\mathbf{x}, t) \in \Omega_F} \lambda_{F}(\mathbf{x}, t)\big| \mathcal{F}[\hat{u}(\theta)](\mathbf{x}, t) \big|^2,\label{residual loss 2}
\end{align}
where the weight $\lambda_F(\mathbf{x},t)$ depends on $\mathbf{x}$ and $t$, and it can be updated or learned during the training.

\subsection{Adaptive weighting from a min-max optimization perspective}

As mentioned earlier, there are several approaches to update the weights $\lambda_F(\mathbf{x},t)$ at each sampled point~\cite{RBA-PINN,SA-PINN,song2024loss}. The central idea is to increase the relative weight at a point if its residual loss is large. In this subsection, we establish a connection between RBA~\cite{RBA-PINN} and a primal-dual update arising from a saddle-point formulation.

For convenience, we denote a training point as $x_i$ (including both spatial and temporal variables), with corresponding residual $r_i$ and weight $\lambda_i$. Ideally, $\lambda_i$ should be large when $r_i$ is large. However, it is also important to prevent any $\lambda_i$ from becoming excessively large, as this would cause the PINN to focus disproportionately on a small subset of the training data. To balance these objectives, we consider the following min-max optimization problem:
\begin{align}
    \min_{\theta}\max_{\lambda}~\mathcal{L}(\lambda,\theta) = \lambda^Tr(\theta) - \frac{1}{2}\lambda^T\lambda+\mathcal{L}_2(\theta), \label{OP}
\end{align}
where $r = [r_1, r_2,\ldots, r_{N_f}]$ and $\lambda = [\lambda_1, \lambda_2, \ldots, \lambda_{N_f}]$ are the vectors of residuals and their associated weights at different points $x_i$. The term $\mathcal{L}_2(\theta)$ aggregates the remaining loss components, such as $\lambda_B \mathcal{L}_{B}(\theta) + \lambda_I \mathcal{L}_I(\theta)$ in~\eqref{PINN loss}, or $\lambda_{\text{obs}} \mathcal{L}_{\text{obs}}(\theta)$ in~\eqref{PINN inverse loss}.

We aim to increase $\lambda_i$ when $r_i$ is large, while the regularization term $\lambda^T \lambda$ discourages overly large weights. The resulting alternating gradient descent-ascent updates for $\theta$ and $\lambda$ are given by:
\begin{align}
    \theta^{(k+1)} &=\theta^{(k)}- \eta_{\theta}\nabla_{\theta}\mathcal{L}(\lambda^{(k)},\theta)= \theta^{(k)} - \eta_{\theta}\sum_{i=1}^{N_f}\lambda_i^{(k)}\nabla_{\theta}r_i(\theta^{(k)}),\\
    \lambda^{(k+1)}&=\lambda^{(k)} +\eta_{\lambda}
    \nabla_{\lambda}\mathcal{L}(\lambda,\theta^{k+1})=(1-\eta_{\lambda})\lambda^{(k)} +\eta_{\lambda}r^{(k+1)},
    \label{GDA}
\end{align}
where $\eta_{\theta}, \eta_{\lambda} > 0$ are the learning rates for $\theta$ and $\lambda$, respectively. Note that RBA~\cite{RBA-PINN} additionally normalizes $r^{(k)}$ and introduces a separate parameter for its influence in the update of $\lambda^{(k+1)}$.

\subsection{Leveraging Spatiotemporal Smoothness via Convolution}
Physical systems often exhibit spatiotemporal smoothness, implying that residuals at neighboring points are typically correlated~\cite{wu2024ropinn}. Traditional pointwise residual weighting neglects this structure, potentially amplifying noise and high-frequency fluctuations. To better exploit this inherent continuity, we first consider an idealized formulation in which residuals are convolved over the entire domain using a symmetric positive semidefinite (SPD) linear operator $W$:
\begin{align}
\bar{r}(\cdot) := W r(\cdot),\label{convolution-ideal}
\end{align}
where $W$ denotes a global convolution matrix defined over all grid points. We adopt the discrete convolution setting for its intuitive interpretation and its direct connection to the algorithm that will be introduced. To ensure symmetry and stability in the resulting min-max optimization, we employ the square root of the SPD operator, denoted as $\sqrt{W}$, which is also symmetry and satisfies $\sqrt{W}\sqrt{W}=W$. This leads us to the following modified optimization problem, which parallels Eq.~\eqref{OP}:
\begin{align}
\min_{\theta}\max_{\lambda}~\mathcal{L}(\lambda,\theta) = \lambda^T\sqrt{W} r(\theta) - \frac{1}{2}\lambda^T\lambda + \mathcal{L}_2(\theta). \label{OP-full-grid}
\end{align}

The gradient descent-ascent updates for this formulation are given by:
\begin{align}
    \theta^{(k+1)} &= \theta^{(k)} - \eta_{\theta} \left( \sum_{i=1}^N (\sqrt{W} \lambda)_i^{(k)} \nabla_{\theta} r_i^{(k)}(\theta^{(k)}) + \nabla_{\theta} \mathcal{L}_2(\theta^{(k)}) \right), \label{GDA-sample-primal}\\
    \lambda^{(k+1)} &= (1 - \eta_{\lambda}) \lambda^{(k)} + \eta_{\lambda} \sqrt{W} r^{(k+1)}. \label{GDA-sample-dual}
\end{align}

To simplify the expressions, we introduce a new variable $\tilde{\lambda} := \sqrt{W} \lambda$, which transforms the updates into the following form:
\begin{align}
    \theta^{(k+1)} &= \theta^{(k)} - \eta_{\theta} \left (\sum_{i=1}^{N}\tilde\lambda_i^{(k)}\nabla_{\theta}r_i^{(k)}(\theta^{(k)})+ \nabla_{\theta}\mathcal{L}_2(\theta^{(k)})\right),\label{GDA-sample-primal-2}\\
    \tilde\lambda^{(k+1)}&=(1-{{\eta_{\lambda}}})\tilde\lambda^{(k)} +{\eta_{\lambda}}Wr^{(k+1)},\label{GDA-sample-dual-2}
\end{align}
where Eq.~\eqref{GDA-sample-dual-2} follows directly from multiplying both sides of the dual update in Eq.~\eqref{GDA-sample-dual} by $\sqrt{W}$.


To address the computational challenges associated with full-grid optimization, we employ stochastic sampling to approximate the primal update in Eq.~\eqref{GDA-sample-primal-2}. Specifically, at each iteration, we select a subset of $N_f$ collocation points ($N_f \ll N$), leading to the following approximation:
\begin{align}
\theta^{(k+1)} = \theta^{(k)} - \eta_{\theta}\left (\sum_{i=1}^{N_f}\tilde{\lambda}_i^{(k)}\nabla_{\theta}r_i^{(k)}(\theta^{(k)})+ \nabla_{\theta}\mathcal{L}_2(\theta^{(k)})\right),\label{GDA-sample-primal-approx}
\end{align}
and the update of $\tilde{\lambda}$ is applied only to the sampled points.

Note Eq.~\eqref{GDA-sample-primal-approx} demonstrates that the primal update does not depend on $Wr(\theta)$, which helps keep the overall computational cost low. The update of each $\tilde{\lambda}_i$ in Eq.~\eqref{GDA-sample-dual-2} only requires evaluating the convolution at the sampled points, and this involves only a forward computation. Notably, the forward pass is significantly less expensive than the backward computation required for updating $\theta$.

To efficiently compute the action of the convolution operator $W$ at a point $x_i$, we propose using a localized average over a neighborhood defined by $\mathcal{N}(x_i, \epsilon) = \{x \in \Omega \mid \|x - x_i\| < \epsilon\}$. The resulting smoothed residual is given by:
\begin{align}
\bar{r}(x_i) = \frac{1}{M+1}\left(r(x_i) + \sum_{j=1}^{M} r(x_j)\right), \quad x_j \in \mathcal{N}(x_i, \epsilon), \label{average}
\end{align}
where $M$ is the number of points randomly sampled from the neighborhood. This approach reduces the computational complexity to $\mathcal{O}(M N_f)$, making the method scalable while still capturing the benefits of convolution. Notably, Eq.~\eqref{average} corresponds to applying a sparse approximation of $W$, where each row of $W$ has at most $M+1$ non-zero entries.



\subsection{Normalization for Stable Weight Dynamics}
In the min-max optimization framework (Eq.~\eqref{OP}), the gradient ascent update for $\lambda$ (Eq.~\eqref{GDA}) dynamically scales the residuals to prioritize critical regions. However, without proper normalization, the effective contribution of the residuals to the loss function can become disproportionately large or small, which in turn adversely affects the update of $\theta$.


The RBA method~\cite{RBA-PINN} applies normalization by scaling residuals with their $\ell^\infty$ norm, ensuring that the maximum weight remains bounded. The update rule for each weight $\lambda_i$ is given by:
\begin{align}
    \lambda_i^{(k+1)}=(1-\eta_\lambda)\lambda_i^{(k)} +\eta^*\frac{r_i^{(k+1)}}{\|r\|_{\infty}}.\label{RBA}
\end{align}
which guarantees that $\lambda_i \in \left(0, \frac{\eta^*}{\eta_\lambda}\right]$. In particular, when $\eta^* = \eta_\lambda$, each weight remains within the interval $(0, 1]$. While this approach prevents individual weights from growing arbitrarily large, it does not control their global distribution: the total sum $\sum_i \lambda_i$ can still vary significantly across iterations and problem scales.

To address this, we adopt an alternative normalization strategy that explicitly constrains the sum of the weights. Specifically, we propose the following update:
\begin{align}
    \lambda_i^{(k+1)}=(1-\eta_{\lambda})\lambda_i^{(k)} +\eta_{\lambda}\frac{\bar{r}_i^{(k+1)}}{\sum_{j=1}^{N_f}\bar{r}_j^{(k+1)}},\label{CWP}
\end{align}
where $\bar{r}_i^{(k+1)}$ is a smoothed or locally averaged residual. This formulation guarantees that $\sum_{i=1}^{N_f} \lambda_i = 1$ at every iteration, provided the weights are initialized uniformly as $\lambda_i^{(0)} = 1/N_f$. 

This sum-to-one normalization offers two key advantages. First, it prevents the total weight magnitude from drifting, thereby preserving a consistent influence on the overall loss landscape. Second, it inherently balances the effect of the learning rate $\eta_\lambda$: while increasing $\eta_\lambda$ amplifies the contribution of the current residuals, the normalization term adapts dynamically to their total scale. As a result, the method is more robust to the specific choice of $\eta_\lambda$ and less sensitive to variations in the residual magnitudes—an important property when dealing with multiscale or heterogeneous PDE problems.

\subsection{Adaptive Resampling for High-Residual Regions}
\label{sec: re-sampling}
To enhance the model’s focus on critical spatial regions and leverage the spatiotemporal smoothness prior, we introduce an adaptive resampling mechanism that dynamically updates the collocation points based on local residual magnitudes. This strategy addresses a central challenge in large-scale PINNs: efficiently directing computational effort toward regions where the model most significantly violates physical constraints.

Our resampling approach mimics the effect of the global convolution operator $W$ by concentrating training on regions where the smoothed residual $\bar{r}(\theta)$ (as defined in Eq.~\eqref{average}) is dominant. This is consistent with prior adaptive sampling techniques in the PINN literature that emphasize high-residual regions~\cite{wu2023comprehensive, lu2021deepxde, gao2023failure}. Since the residuals $r(x_j)$ over each local neighborhood $\mathcal{N}(x_i, \epsilon)$ are already computed during the localized weighting step, we propose a highly efficient reuse of this data for resampling. Specifically, for each training point $x_i$, we identify the point in its neighborhood with the highest residual and replace $x_i$ accordingly in the next iteration.

This resampling process incurs no additional cost in terms of function evaluations, as all necessary residuals are already available from the convolution-based smoothing step. To ensure training stability and avoid premature shifts in focus, we perform this update only every $K$ training steps (empirically, $K=200$ in our experiments). This periodic schedule allows residuals to stabilize before updating the training set, improving both robustness and convergence behavior.


An important design consideration is how to manage the neighborhoods when sample points are updated. Specifically, we consider two variants:
\begin{itemize}
\item  \textbf{CWP-fix:} After replacing $x_i$ with a new point, we retain the original neighborhood $\mathcal{N}(x_i, \epsilon)$ for both weighting and residual evaluation. This ensures continuity and local consistency, as the region of influence remains unchanged.

\item \textbf{CWP:} After replacement, the neighborhood is re-centered at the new point, i.e., we redefine $\mathcal{N}(x_i, \epsilon)$. This allows the model to explore new spatial regions and adaptively track evolving error hotspots.
\end{itemize}
Both strategies preserve the overall computational complexity of $\mathcal{O}(M N_f)$, as they leverage residuals already computed during localized smoothing. In summary, this adaptive resampling mechanism efficiently reallocates training resources toward regions of high residual error, aligning model training with the physical structure of the problem while maintaining scalability.

\section{Experiments}
\label{Experiment}

In this section, we evaluate the proposed convolution-based weighting methods across a variety of PDE benchmarks. We compare our approach against several representative dynamic weighting strategies developed for PINNs. As discussed in Section~\ref{Introduction}, we consider a range of established techniques, including the self-adaptive weighting method~\cite{SA-PINN}, the loss-attentional weighting method~\cite{song2024loss}, and the residual-based attention weighting method~\cite{RBA-PINN}, referred to as SA, LA, and RBA, respectively. We include three variants of our method in the comparison: CW, which applies convolution-based weighting without resampling, and two methods with adaptive resampling—CWP-fix, which retains fixed neighborhoods, and CWP, which updates neighborhoods dynamically.

Unless otherwise specified, we randomly sample $M = 4$ points within the neighborhood $\mathcal{N}(x, \epsilon = 0.01)$ of each training point $x$. The hyperparameter $\eta_{\lambda}$ in Eq.~\eqref{CWP} is set to a default value of $0.001$ throughout all experiments.

For all problem setups, the testing dataset consists of 90,000 randomly distributed points, ensuring consistency across experiments. Training samples are drawn independently from the relevant domains, with the quantities denoted as \(N_f\) for residual terms, \(N_b\) for boundary conditions, and \(N_0\) for initial conditions, where applicable. The training and testing sets are strictly non-overlapping to preserve the independence of evaluation data. This sampling strategy is designed to balance computational efficiency and memory usage while accommodating varying levels of problem complexity.

For each PDE, every model is trained and evaluated over three independent trials using distinct random initialization seeds. The best-performing result among the three trials is reported as the representative performance. Model accuracy is assessed using two metrics: the relative \( L^2 \) error and the \( L^{\infty} \) norm, defined as follows:

\[
\text{Relative } L^2 \text{ error} = \frac{\sqrt{\sum_{k=1}^N \left| \hat{u}(\mathbf{x}_k, t_k) - u(\mathbf{x}_k, t_k) \right|^2}}{\sqrt{\sum_{k=1}^N \left| u(\mathbf{x}_k, t_k) \right|^2}},
\]  
\[
L^{\infty} \text{ norm} = \max_{1 \leq k \leq N} \left| \hat{u}(\mathbf{x}_k, t_k) - u(\mathbf{x}_k, t_k) \right|,
\]  
where \( u \) denotes the exact PDE solution, \( \hat{u} \) is the predicted output of the tested model, and \( N \) is the total number of points in the testing dataset.

\subsection{1D Heat Equation with high frequencies}
\begin{figure}[!htb]
\centering
    \begin{minipage}{0.6\textwidth}
     \centering
     \includegraphics[width=\linewidth]{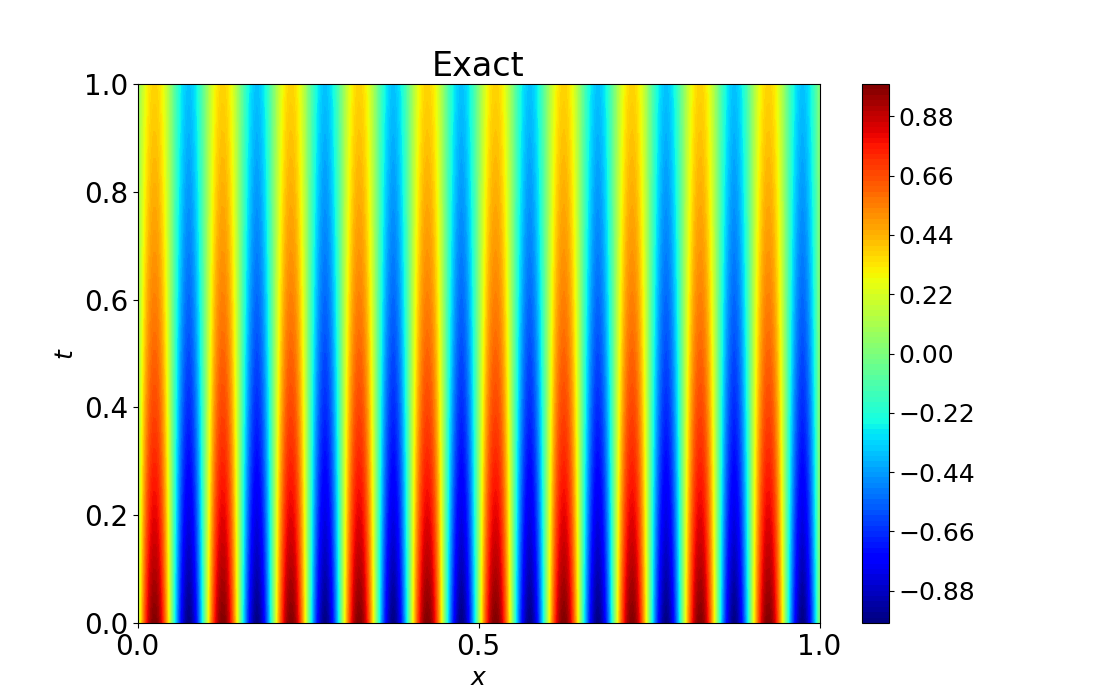} 
   \end{minipage}
    \caption{Exact solution of the 1D Heat Equation.}\label{Exact Heat 1}
\end{figure}

In this section, we evaluate our proposed algorithms on the 1D heat equation with high-frequency dynamics:
\begin{align}
    &u_t = \frac{u_{xx}}{400\pi^2}, \quad x\in[0,1],~t\in[0,1], \label{Heat PDE}\\
    &u(t,0) = u(t,1) = 0,\label{Heat PDE boundary}\\
    &u(0,x) = u_0(x)\quad x\in[0,1],\label{Heat PDE Initial}
\end{align}
where the initial condition \( u_0(x) \) is derived from the analytical solution:
\begin{align}
    u(t,x) = e^{-t}\sin(20\pi x),\label{Heat Sol 1}
\end{align}
as shown in Fig.~\ref{Exact Heat 1}.

All methods are implemented using the same neural network architecture for fair comparison. Each network consists of 4 hidden layers with 80 neurons per layer and employs the Adam optimizer with a learning rate of 0.0015. Training is carried out for 50,000 iterations, with an exponential learning rate decay factor of 0.8 applied every 2,000 iterations.
To enforce boundary and initial conditions, we adopt a hard constraint formulation:
\[
\hat{u}(t,x) = tx(1-x)\hat{u}_{\mathcal{NN}}(t,x) + \sin(20\pi x),
\]
where \( \hat{u}_{\mathcal{NN}}(t,x) \) denotes the raw neural network output and \( \hat{u}(t,x) \) is the final prediction. This formulation guarantees that the Dirichlet boundary conditions and initial conditions are satisfied exactly, allowing the training to focus solely on minimizing the residual loss. 


\begin{table}[!ht]
    \caption{Final performance of various algorithms on the 1D heat equation after 50,000 training iterations with \( N_f = 1{,}000 \).}
    \label{Heat 1000 Table}
    \centering
    \begin{tabular}{|c|c|c|c|c|c|c|}
        \hline
        & CWP & CWP-fix & CW & RBA & SA & LA \\ \hline
        Rel. $L^2$ error & \bm{$1.04\times 10^{-3}$} & $1.97\times 10^{-3}$ & $3.26\times 10^{-3}$ & $1.63\times 10^{-2}$ & $6.15\times 10^{-3}$ & $7.39\times 10^{-3}$ \\ \hline
        $L^{\infty}$ norm & \bm{$5.73\times 10^{-3}$} & $6.01\times 10^{-3}$ & $8.23\times 10^{-3}$ & $3.24\times 10^{-2}$ & $1.03\times 10^{-2}$ & $2.05\times 10^{-2}$ \\ \hline
    \end{tabular}
\end{table}

\begin{figure}[!htb]
\centering
    \begin{minipage}{0.45\textwidth}
     \centering
     \includegraphics[width=\linewidth]{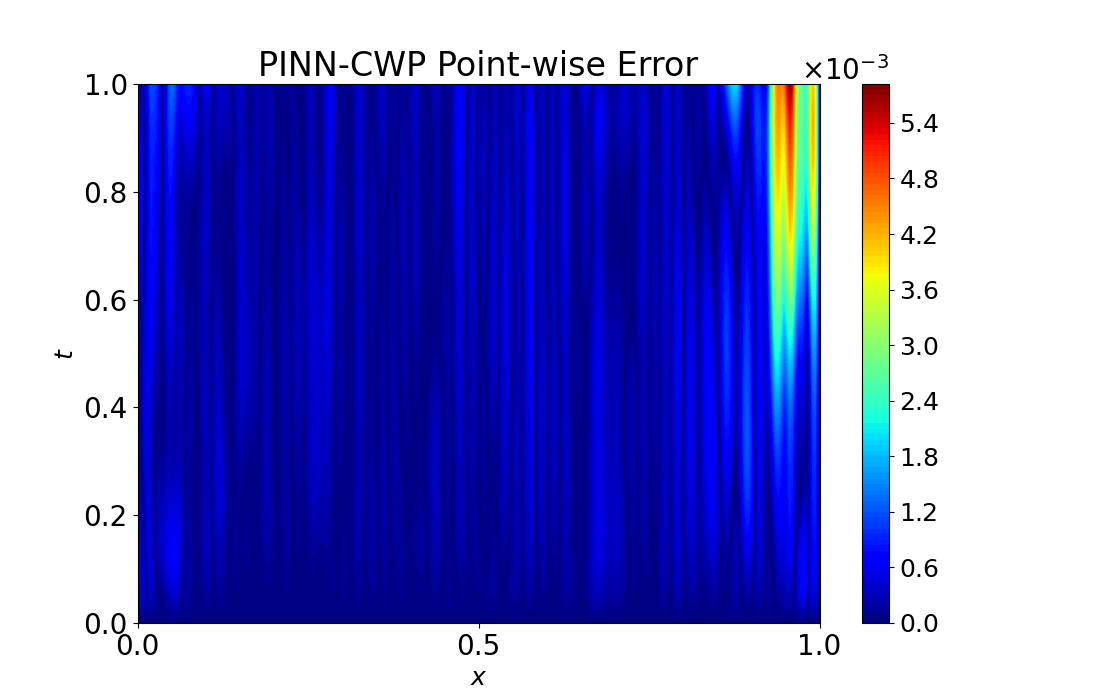}  
   \end{minipage}
    \begin{minipage}{0.45\textwidth}
     \centering
     \includegraphics[width=\linewidth]{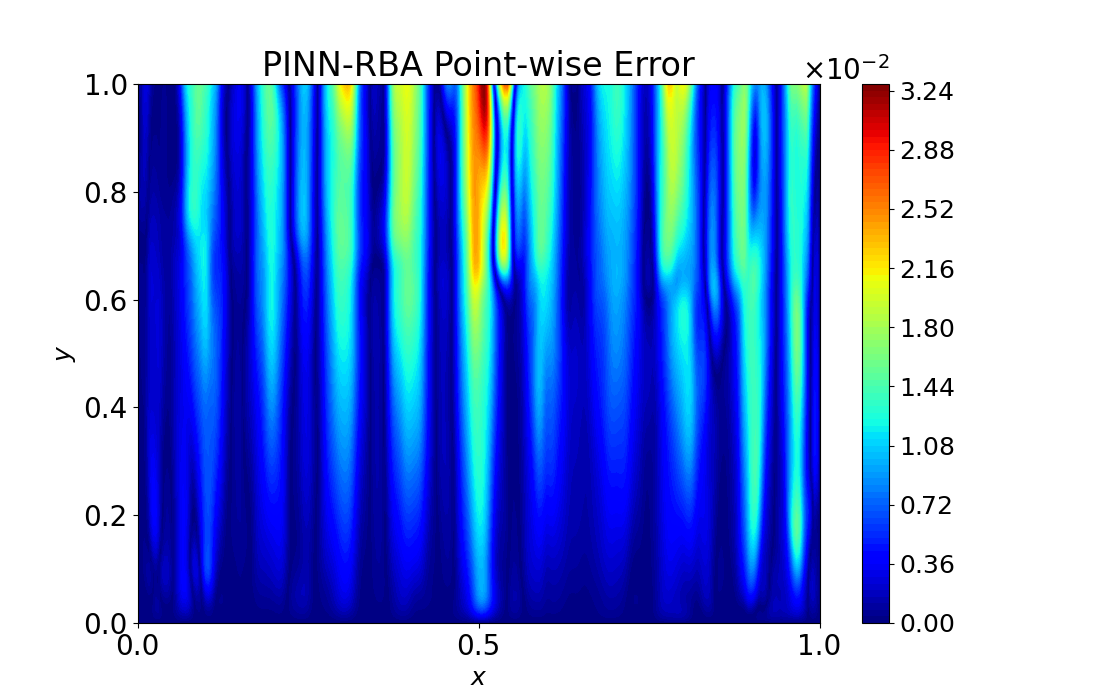}  
   \end{minipage}

    \begin{minipage}{0.45\textwidth}
     \centering
     \includegraphics[width=\linewidth]{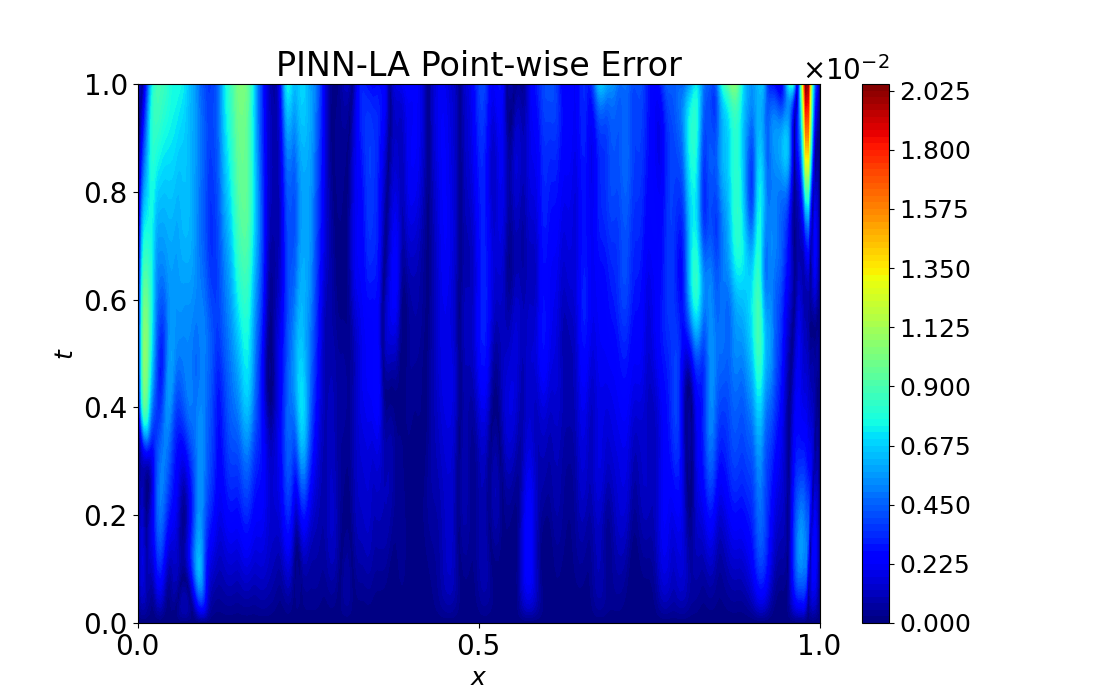}  
   \end{minipage}
    \begin{minipage}{0.45\textwidth}
     \centering
     \includegraphics[width=\linewidth]{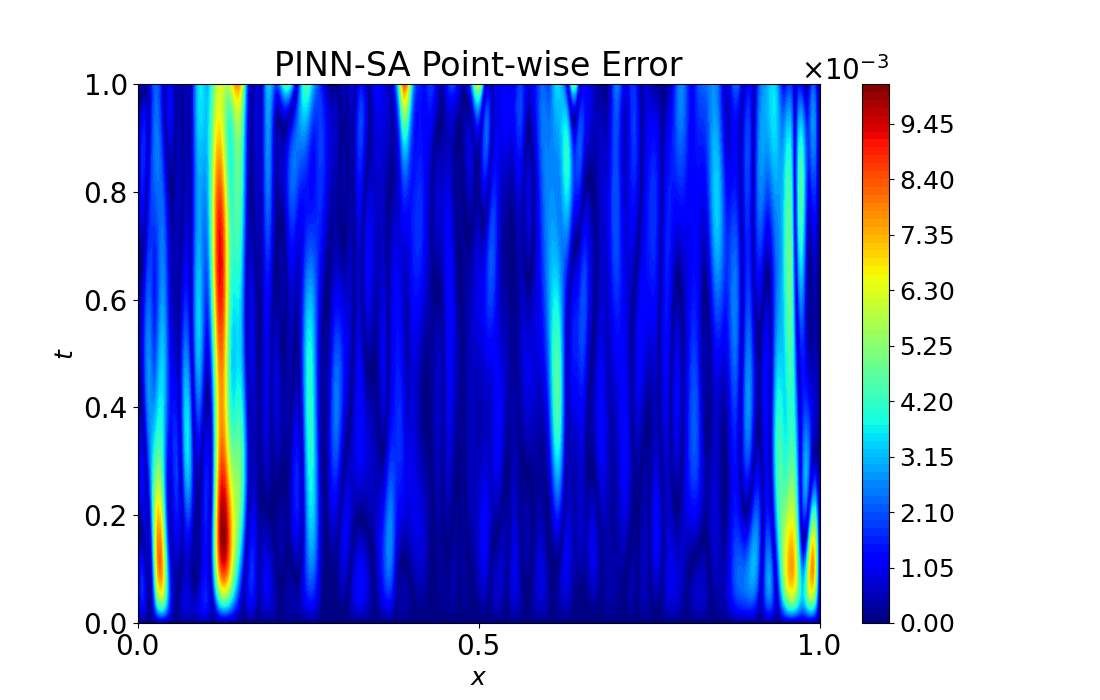}  
   \end{minipage}
    \caption{Point-wise error for the CWP (top-left), RBA (top-right), LA (bottom-left), and SA methods (bottom-right), respectively, on the 1D heat equation.}\label{pointwise error Heat 1000}
\end{figure}





\begin{figure}[!htb]
\centering
    \begin{minipage}{0.45\textwidth}
     \centering
     \includegraphics[width=\linewidth]{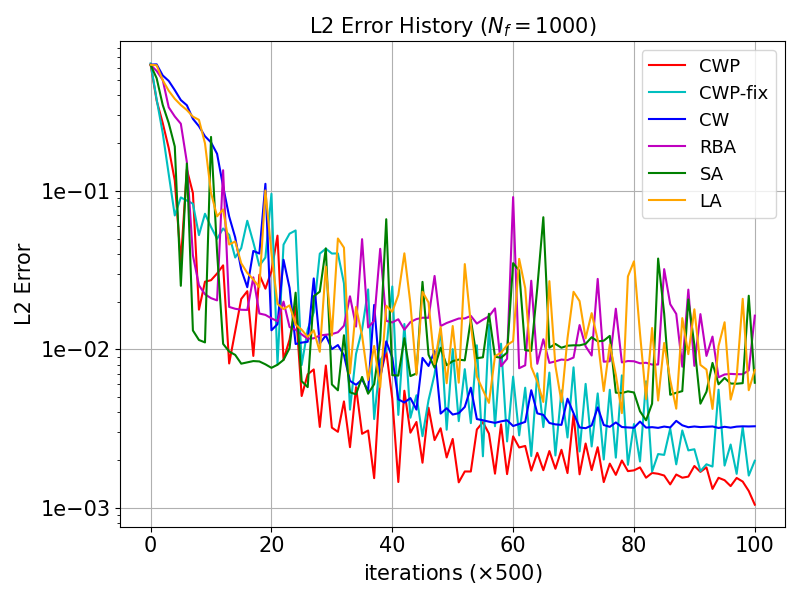} 
   \end{minipage}
    \caption{Training history of the relative \( L^2 \) error for the CWP, RBA, LA, and SA methods on the 1D heat equation with \( N_f = 1{,}000 \).}\label{History Heat 1000}
\end{figure}

Table~\ref{Heat 1000 Table} summarizes the performance of various methods under the same setting, and the empirical results for \( N_f = 1,000 \) are presented in Fig.~\ref{pointwise error Heat 1000} and Fig.~\ref{History Heat 1000}. Among all approaches, CWP achieves the lowest relative \( L^2 \) error and \( L^{\infty} \) norm, outperforming baseline methods such as RBA, SA, and LA. Notably, CWP computes residuals by averaging over four neighboring points for each collocation point, which theoretically increases the computational cost during the forward pass. However, in practice, the forward GPU time for CWP (\(1.679 \pm 0.21\) ms/iteration) remains comparable to that of RBA (\(1.621 \pm 0.22\) ms/iteration) and SA (\(1.612 \pm 0.21\) ms/iteration), despite requiring four times as many residual evaluations. This is because the dominant computational cost in PINN training arises from the backward pass (gradient computation). Even LA, which employs an auxiliary network for adaptivity, incurs higher latency (\(1.710 \pm 0.31\) ms/iteration) due to its larger parameter count. Therefore, the marginal increase in CWP’s forward time is a justified trade-off for its superior accuracy.

\begin{figure}[!htb]
\centering
    \begin{minipage}{0.45\textwidth}
     \centering
     \includegraphics[width=\linewidth]{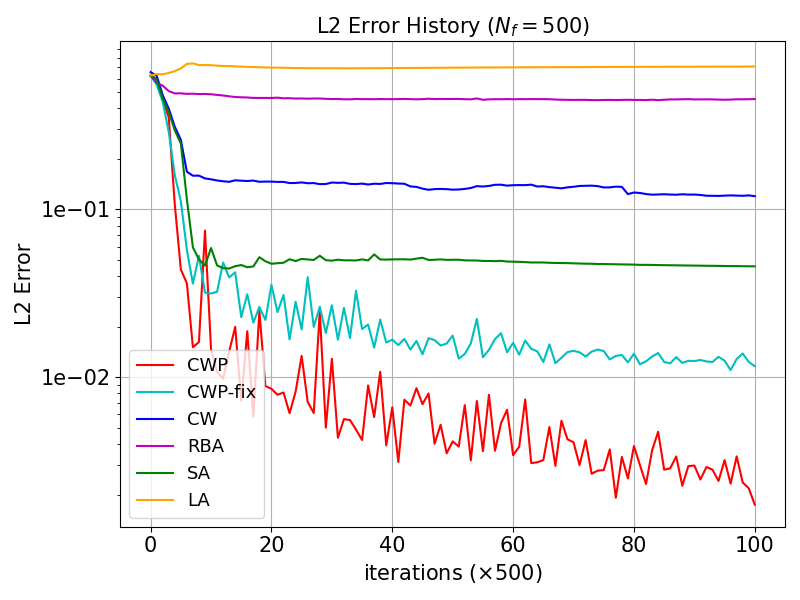}
   \end{minipage}
    \caption{Training history of the relative \( L^2 \) error for the CWP, RBA, LA, and SA methods on the 1D heat equation with \( N_f = 500 \).}\label{History Heat 500}
\end{figure}

The differences between CWP and its ablated variants (CWP-fix and CW) are less pronounced under the original training regime, indicating that the full advantages of our method emerge more clearly in more challenging scenarios. To rigorously assess the contributions of each component within CWP, we reduced the training dataset size to \( N_f = 500 \). Fig.~\ref{History Heat 500} presents the convergence history, while Table~\ref{Heat 500 Table} reports the final errors. In this low-data regime, CWP maintains a robust relative \( L^2 \) error of \(1.73 \times 10^{-3}\), whereas all baseline methods (RBA, SA, LA) fail to achieve errors below \(1 \times 10^{-2}\). Notably, the performance gap between CWP and its ablated variants widens considerably. Specifically, the variant without resampling (CW) exhibits marked performance degradation at \( N_f = 500 \), consistent with our theoretical motivation in Section~\ref{Motivation}, where resampling alleviates sparse data bias. Furthermore, although CWP-fix employs resampling, it underperforms compared to CWP, highlighting that dynamically incorporating neighborhood information via convolution enhances training stability and generalization.


\begin{table}[!h]
    \caption{Final performance of various algorithms on the 1D heat equation after 50,000 training iterations with \( N_f = 500 \).}
    \label{Heat 500 Table}
    \centering
    \begin{tabular}{|c|c|c|c|c|c|c|}
    \hline
    & CWP & CWP-fix & CW & RBA & SA & LA \\ \hline
    Rel. $L^2$ error & \bm{$1.73\times 10^{-3}$} & $1.16\times 10^{-2}$ & $1.13\times 10^{-1}$ & $4.50\times 10^{-1}$ & $4.57\times 10^{-2}$ & $7.07\times 10^{-1}$ \\ \hline
    $L^{\infty}$ norm & \bm{$6.17\times 10^{-3}$} & $2.89\times 10^{-2}$ & $6.23\times 10^{-1}$ & $8.82\times 10^{-1}$ & $1.09\times 10^{-1}$ & $1.31\times 10^{0}$ \\ \hline
\end{tabular}

\end{table}

\label{Section Heat}

\subsection{2D Klein-Gordon equation}
We evaluate all methods on a 2D time-dependent Klein–Gordon equation defined as
\begin{align}
    &u_{tt} - \Delta u + u^2 = f, \quad (x,y)\in \Omega, t\in T,\\
    &u(0,x, y) = x + y,\\
    &u(t, x,y) = g(t,x,y)\quad (x,y)\in\partial\Omega,
\end{align}
where the source term \( f \) and boundary condition \( g \) are constructed to match the analytical solution
\begin{align}
    u(t,x,y) = (x+y)\cos(t) + xy\sin(t).
\end{align}

To solve this problem, we use a fully connected neural network with 4 hidden layers and 80 neurons per layer for all models. The initial condition is enforced via a hard constraint, where the predicted solution is expressed as
\[
\hat{u}(t, x, y) = t \hat{u}_{\mathcal{NN}}(t, x, y) + x + y,
\]
ensuring that \( \hat{u}(0, x, y) = x + y \) exactly. As a result, the training loss only includes the residual term~\eqref{residual loss} and the boundary term~\eqref{boundary loss}, with the latter weighted by a factor of \( \lambda_B = 100 \).

All experiments use the same training configuration: \( N_f = 1{,}000 \) collocation points and \( N_b = 300 \) boundary points. The training process employs the Adam optimizer with an exponential learning rate scheduler (initial rate \( 5 \times 10^{-3} \), decay factor 0.8 every 1{,}000 iterations).

\begin{figure}[!htb]
\centering
    \begin{minipage}{0.45\textwidth}
     \centering
     \includegraphics[width=\linewidth]{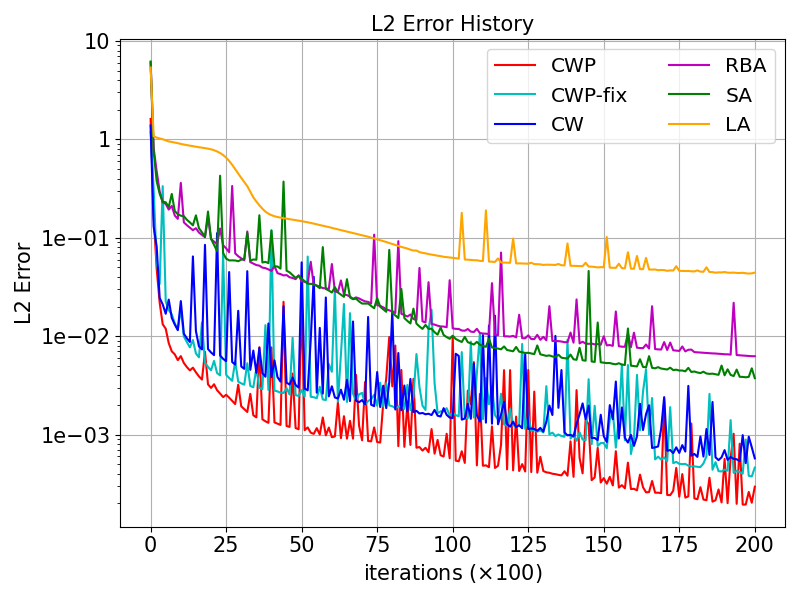} 
   \end{minipage}

    \caption{2D Klein-Gordon equation: Training history of the relative \( L^2 \) error for the CWP, RBA, SA, and LA methods with \( N_f = 1,000 \) and \(N_b=300\).}\label{fig: 2D KG HIST}
\end{figure}

We compare the proposed CWP method with two ablated variants (CWP-fix and CW) and three representative dynamic weighting strategies (RBA, SA, and LA). The training histories for all methods are shown in Fig.~\ref{fig: 2D KG HIST}. 
During training, the CWP method consistently achieves the best performance among all evaluated models. It outperforms both SA and RBA, with particularly pronounced improvements over LA. Final numerical results, reported in Table~\ref{table: KG}, show that CWP attains the lowest relative \( L^2 \) errors and \( L^{\infty} \) norms across the board. 



\begin{table}[!h]
    \caption{Final performance of various algorithms on the 2D Klein-Gordon equation after 20,000 training iterations with \( N_f = 1,000 \). }
    \label{table: KG}
    \centering
    \begin{tabular}{|c|c|c|c|c|c|c|}
        \hline
        & CWP & CWP-fix & CW & RBA & SA & LA \\ \hline
        Rel. $L^2$ error & \bm{$2.95\times 10^{-4}$} & $4.61\times 10^{-4}$ & $5.73\times 10^{-4}$ & $6.25\times 10^{-3}$ & $3.74\times 10^{-3}$ & $4.43\times 10^{-2}$ \\ \hline
        $L^{\infty}$ norm & \bm{$2.42\times 10^{-3}$} & $4.14\times 10^{-3}$ & $5.64\times 10^{-3}$ & $3.27\times 10^{-2}$ & $2.74\times 10^{-2}$ & $3.16\times 10^{-1}$ \\ \hline
    \end{tabular}
\end{table}

\begin{figure}[!htb]
\centering
    \begin{minipage}{0.33\textwidth}
     \centering
     \includegraphics[width=\linewidth]{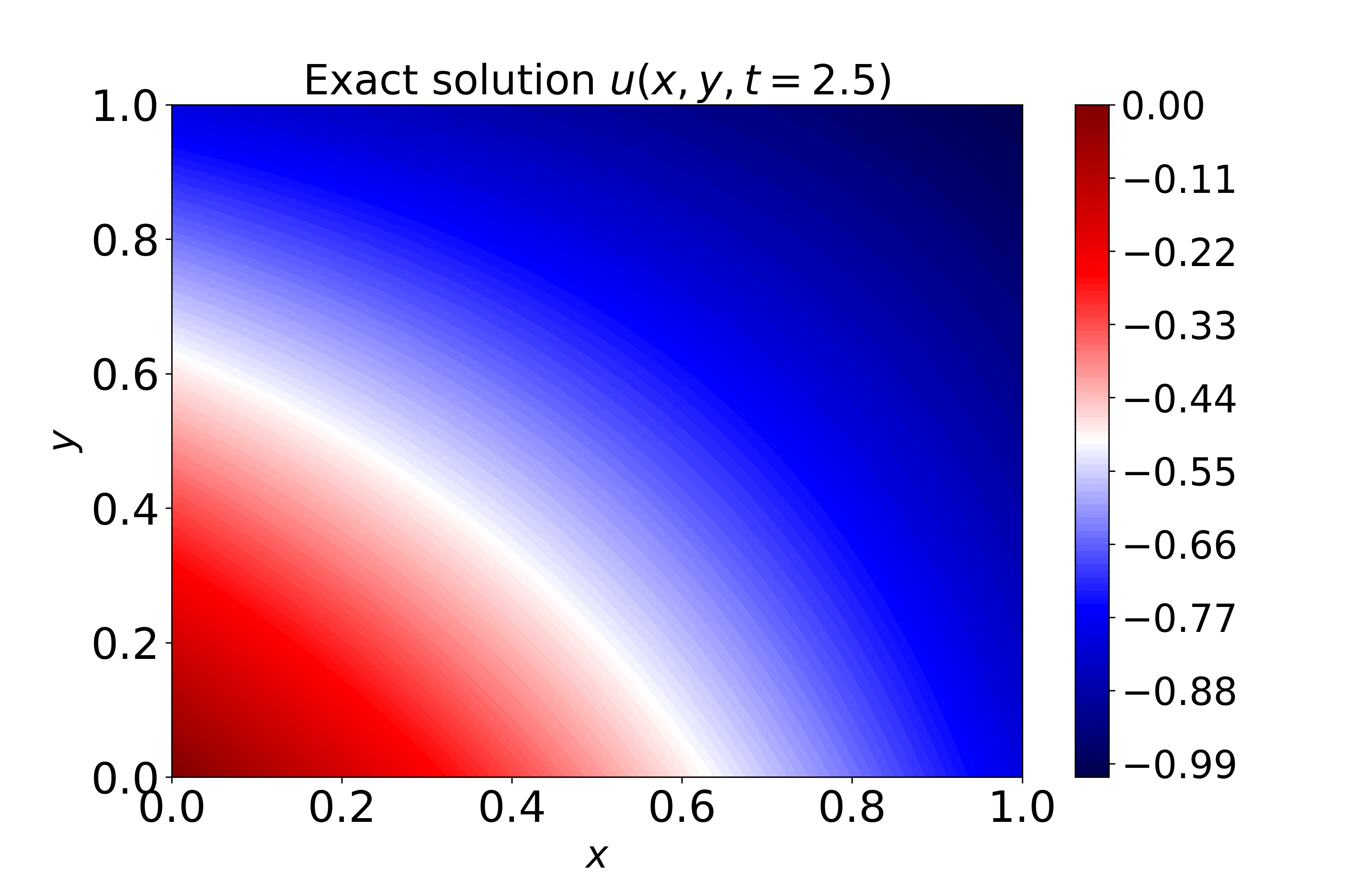} 
   \end{minipage}
    \begin{minipage}{0.33\textwidth}
     \centering
     \includegraphics[width=\linewidth]{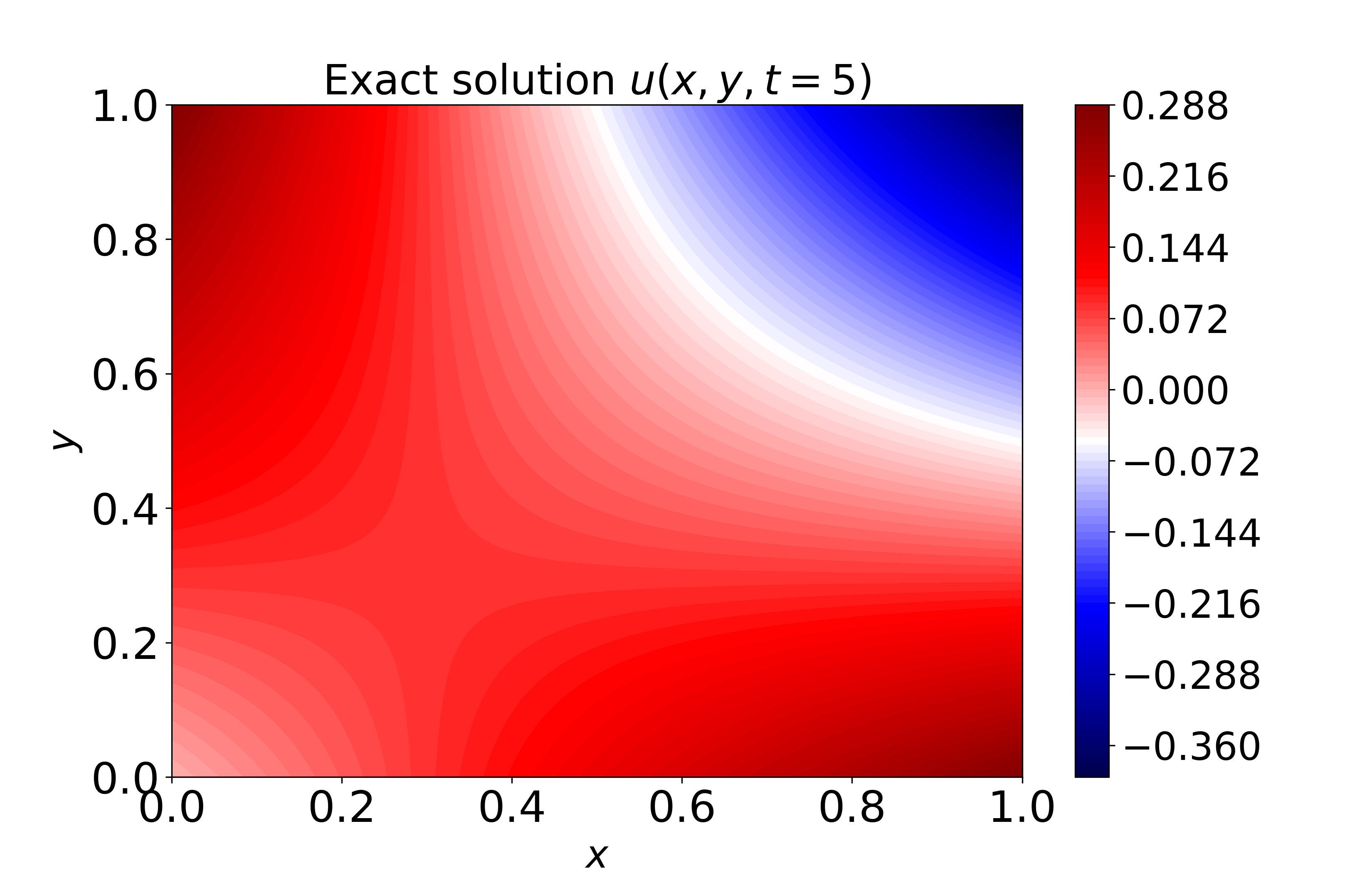}  
   \end{minipage}
   \begin{minipage}{0.33\textwidth}
     \centering
     \includegraphics[width=\linewidth]{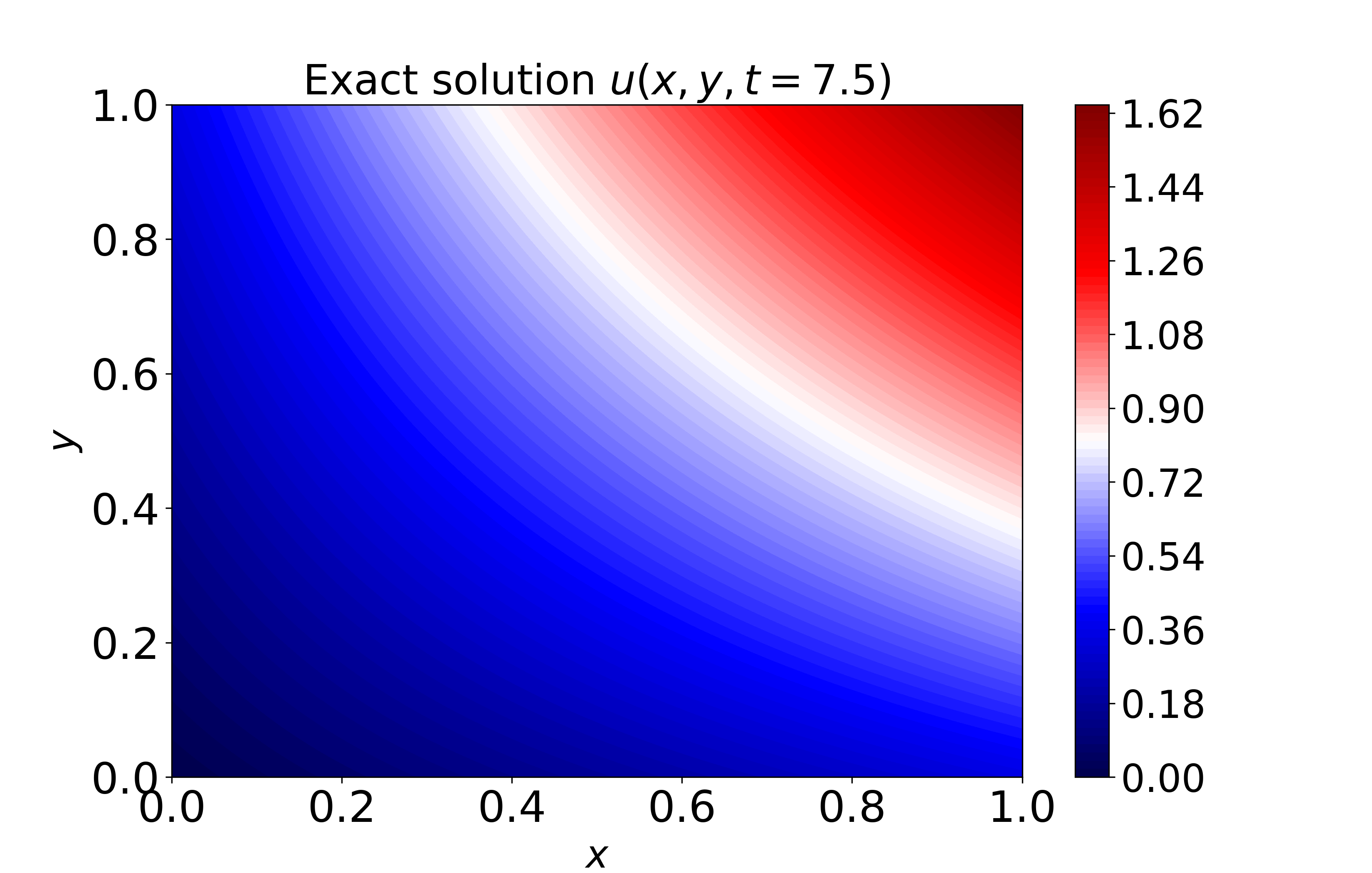}  
   \end{minipage}

   \begin{minipage}{0.33\textwidth}
     \centering
     \includegraphics[width=\linewidth]{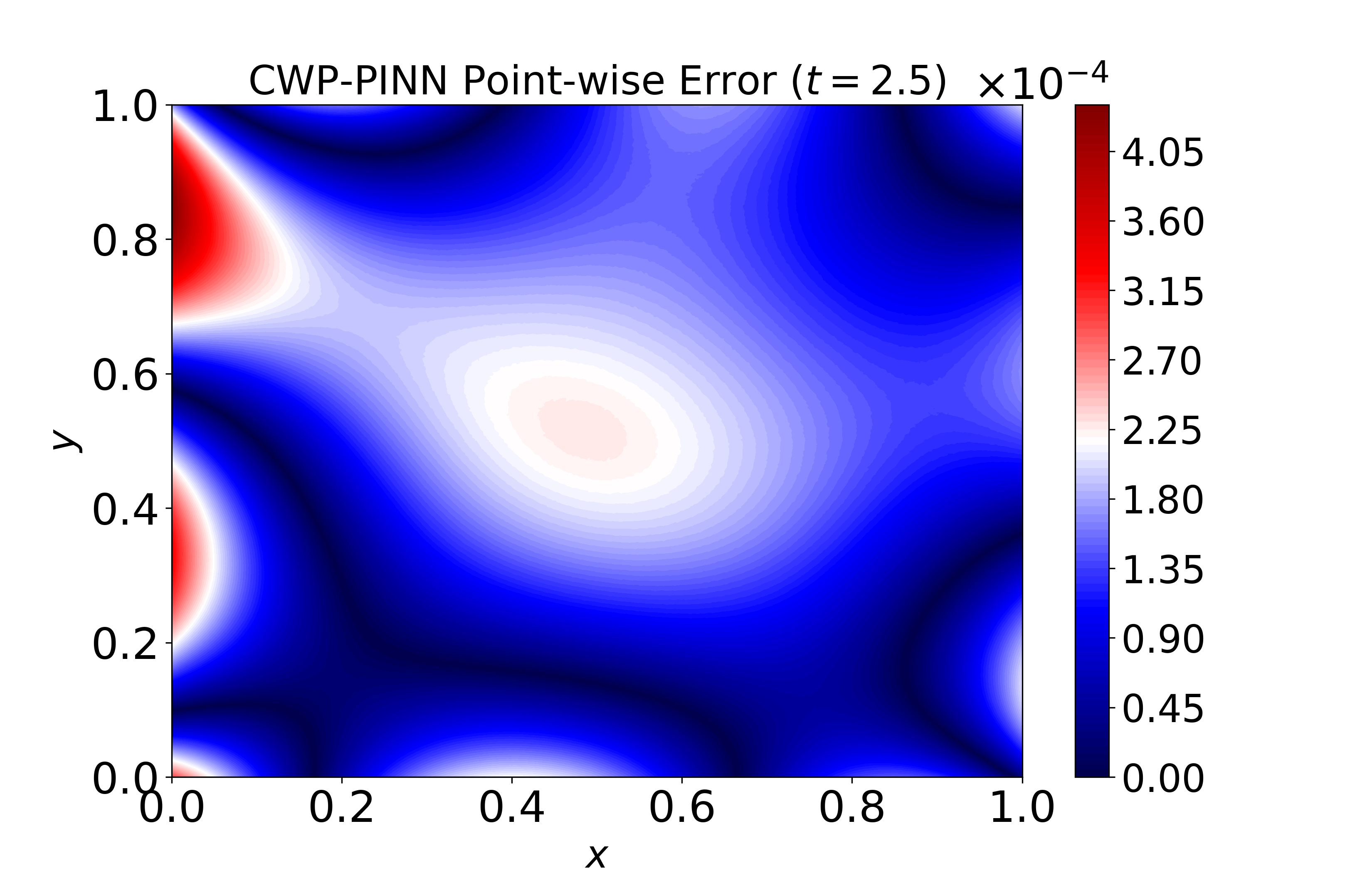} 
   \end{minipage}
    \begin{minipage}{0.33\textwidth}
     \centering
     \includegraphics[width=\linewidth]{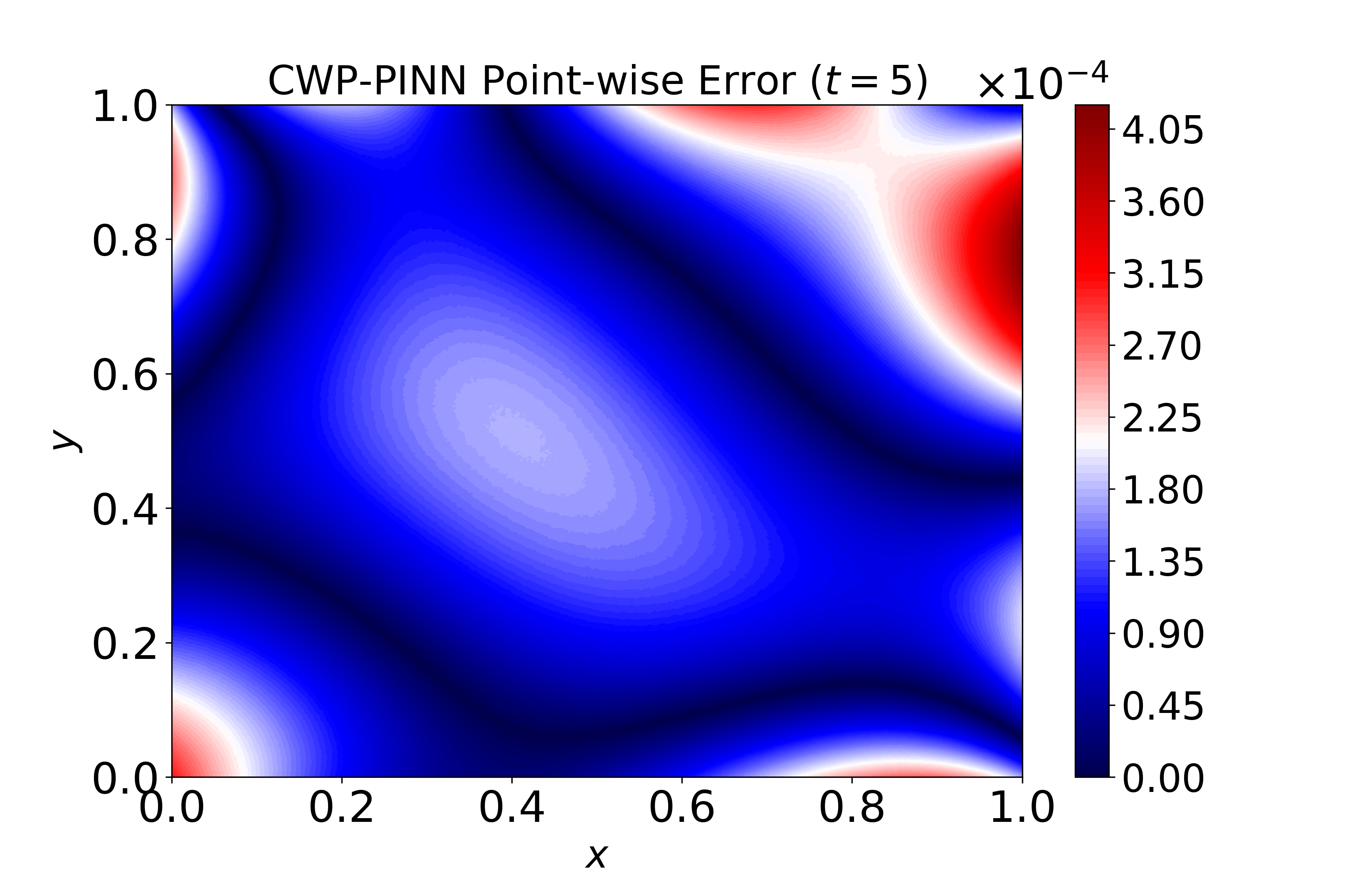}  
   \end{minipage}
   \begin{minipage}{0.33\textwidth}
     \centering
     \includegraphics[width=\linewidth]{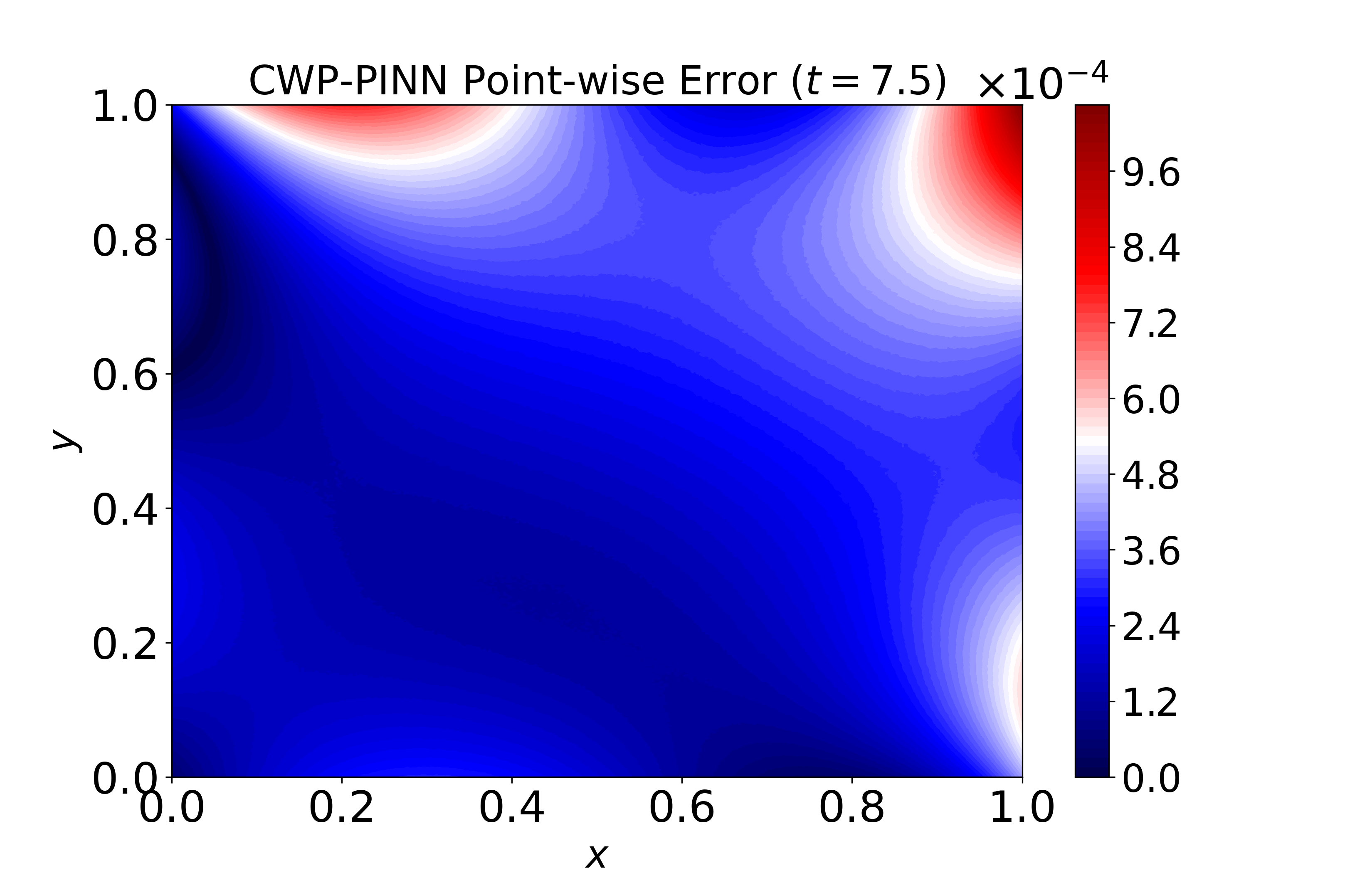}  
   \end{minipage}

   \begin{minipage}{0.33\textwidth}
     \centering
     \includegraphics[width=\linewidth]{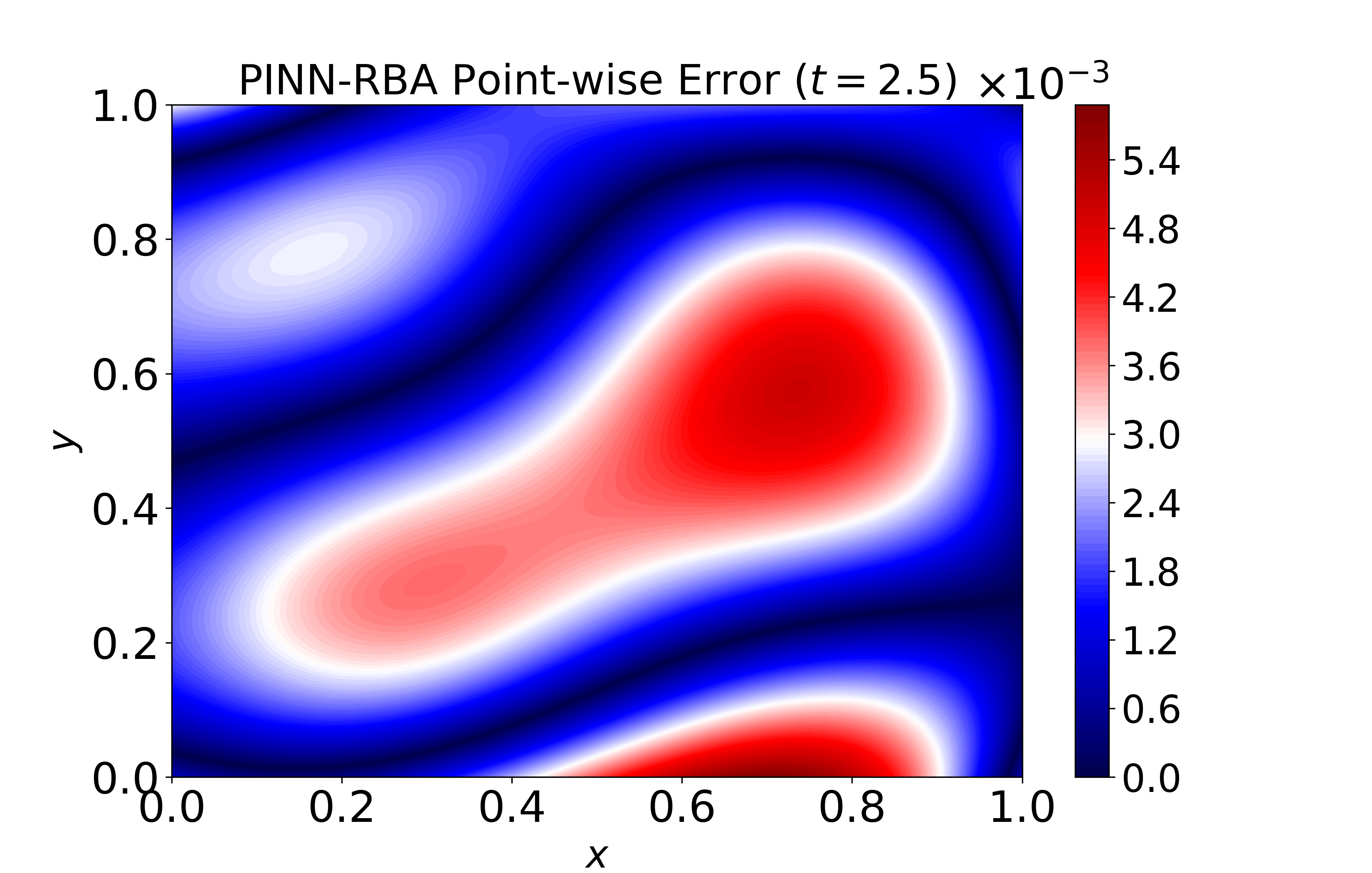} 
   \end{minipage}
    \begin{minipage}{0.33\textwidth}
     \centering
     \includegraphics[width=\linewidth]{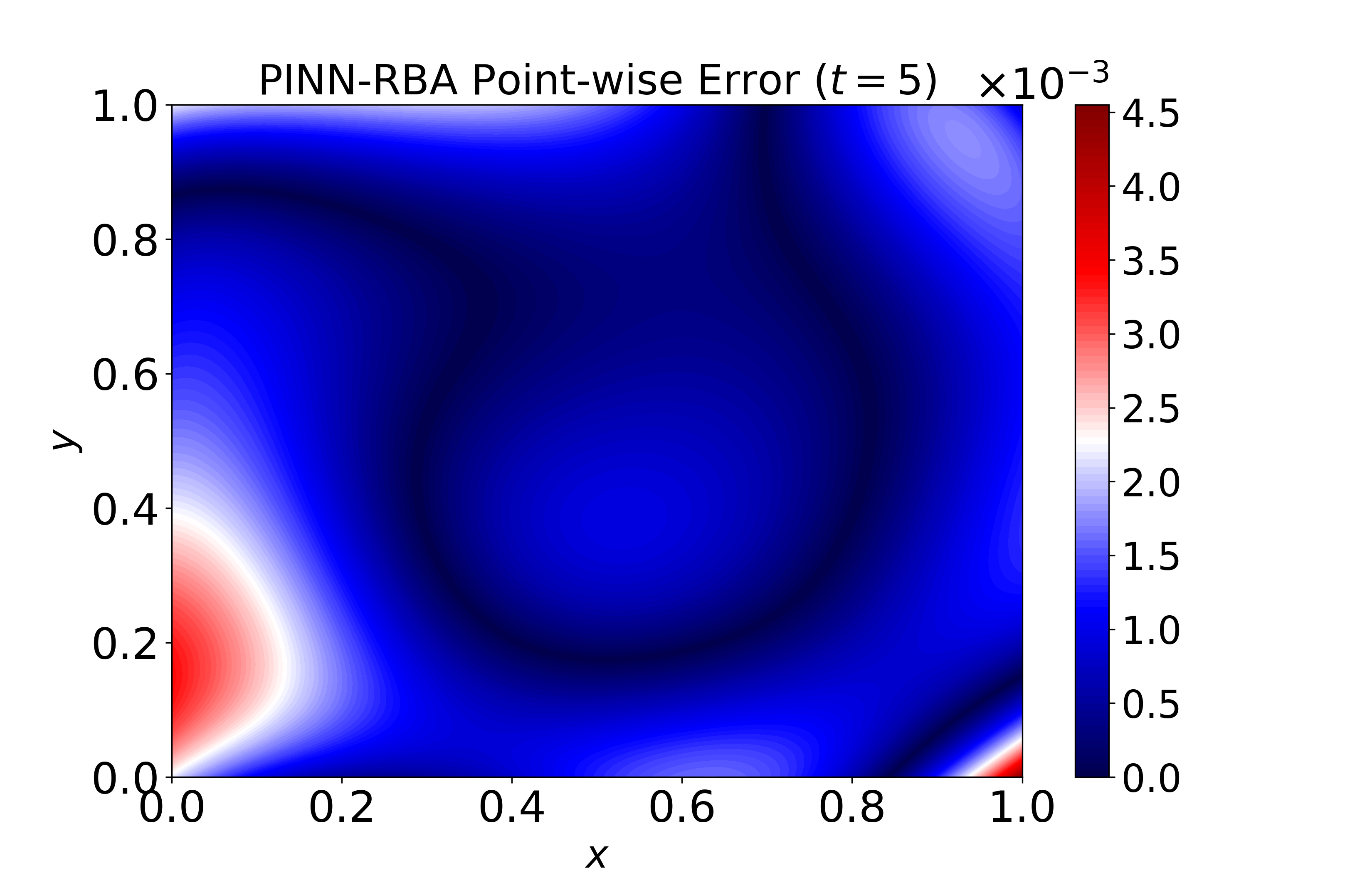}  
   \end{minipage}
   \begin{minipage}{0.33\textwidth}
     \centering
     \includegraphics[width=\linewidth]{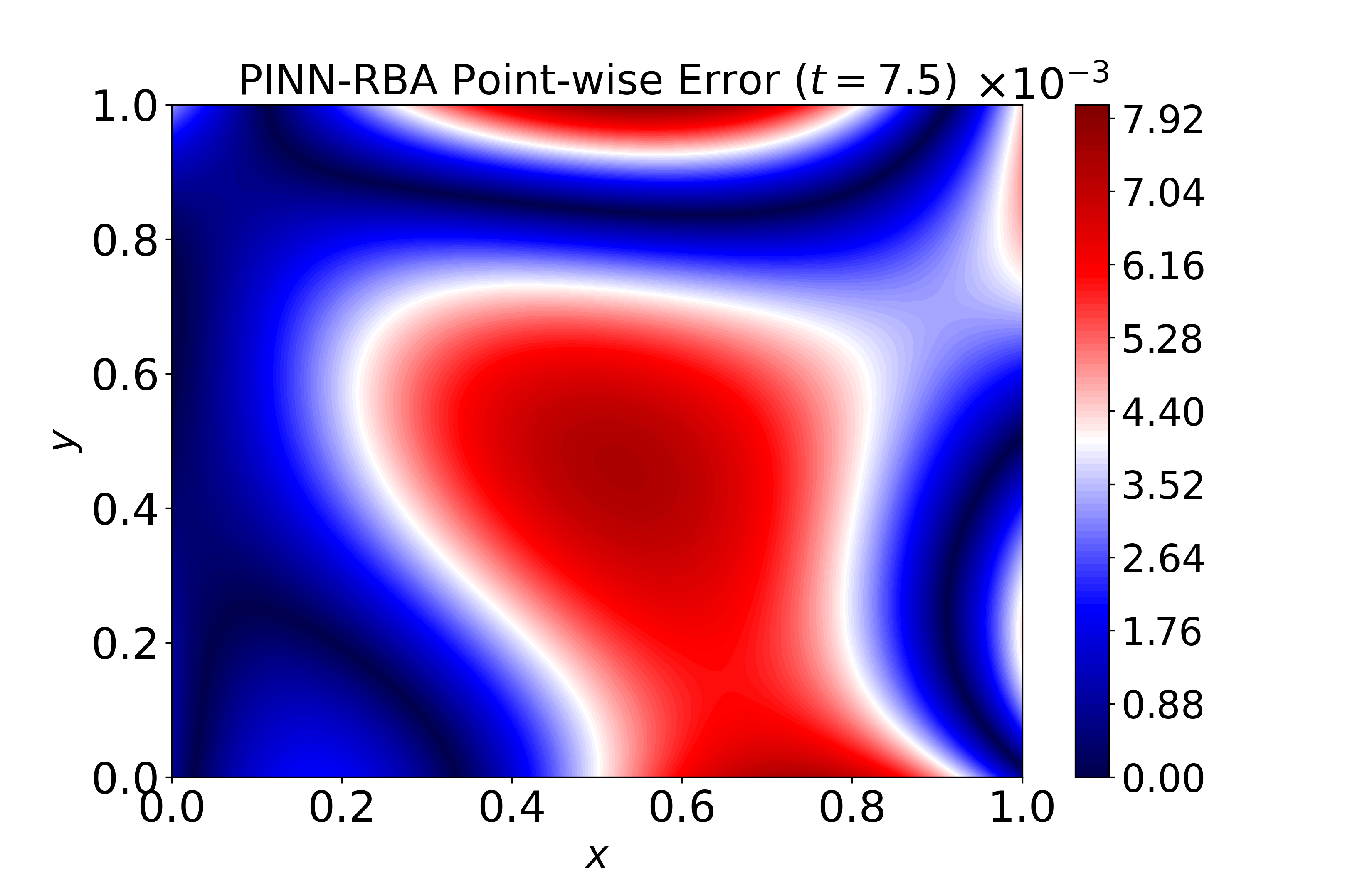}  
   \end{minipage}

   \begin{minipage}{0.33\textwidth}
     \centering
     \includegraphics[width=\linewidth]{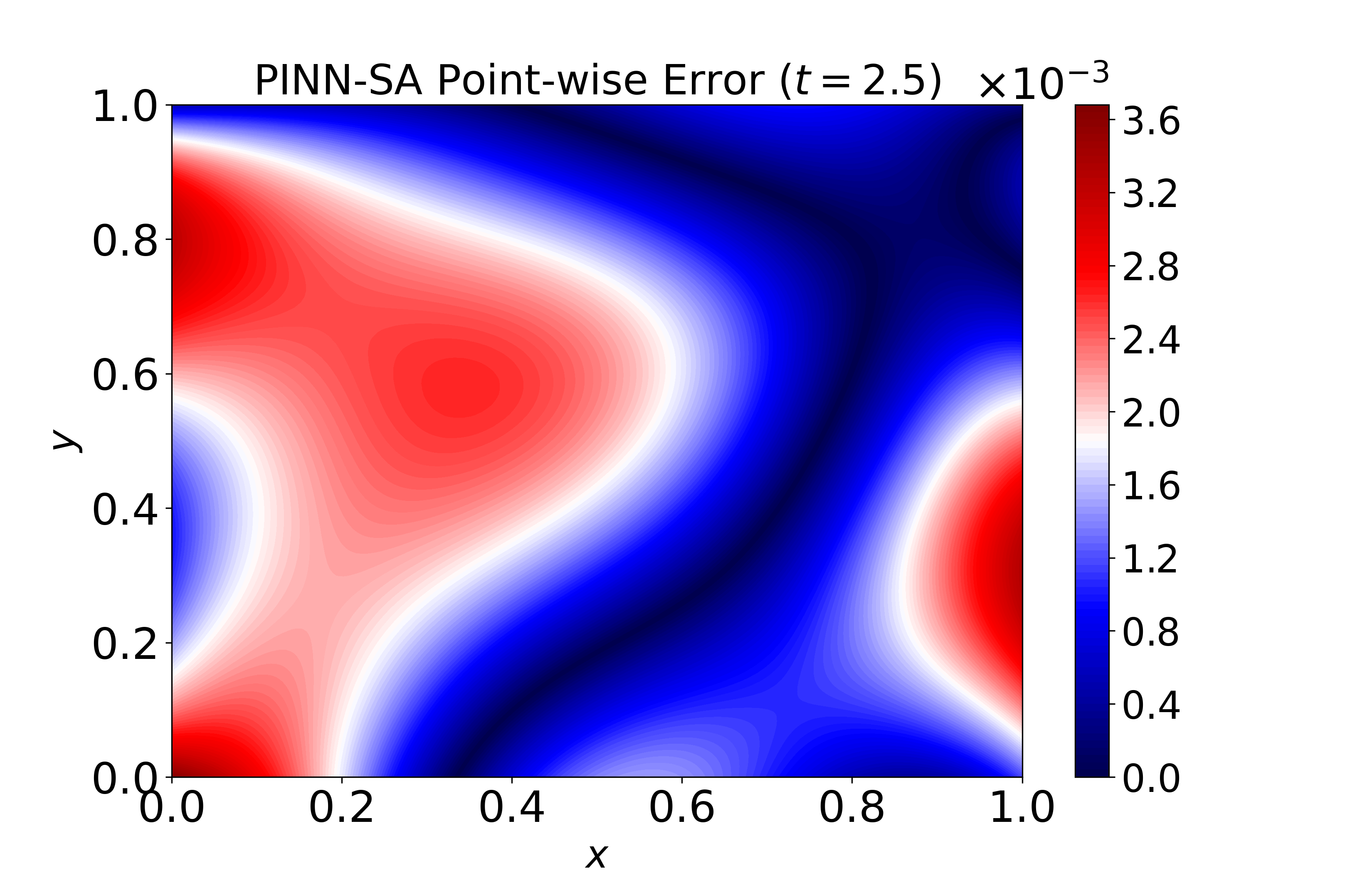} 
   \end{minipage}
    \begin{minipage}{0.33\textwidth}
     \centering
     \includegraphics[width=\linewidth]{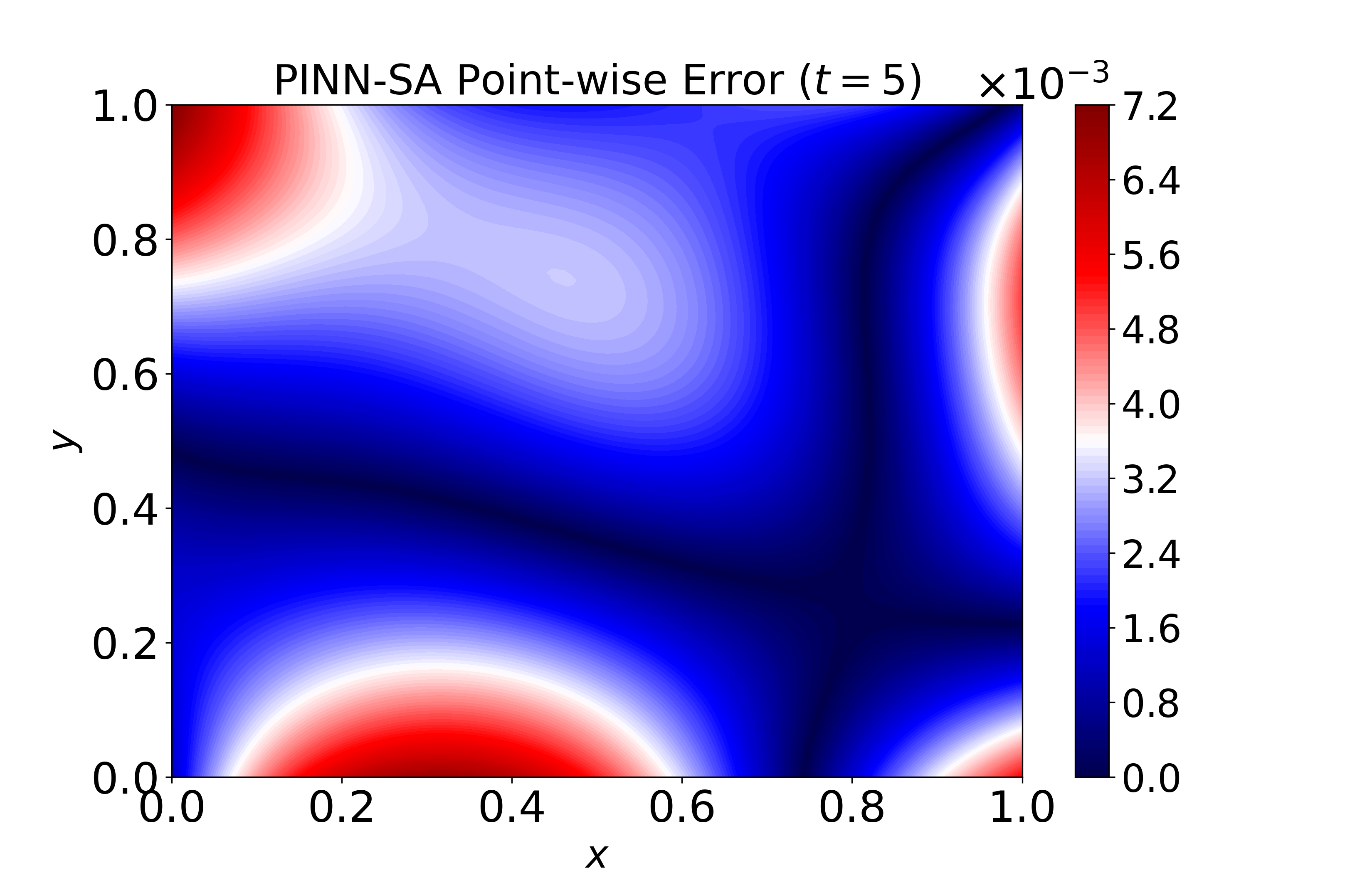}  
   \end{minipage}
   \begin{minipage}{0.33\textwidth}
     \centering
     \includegraphics[width=\linewidth]{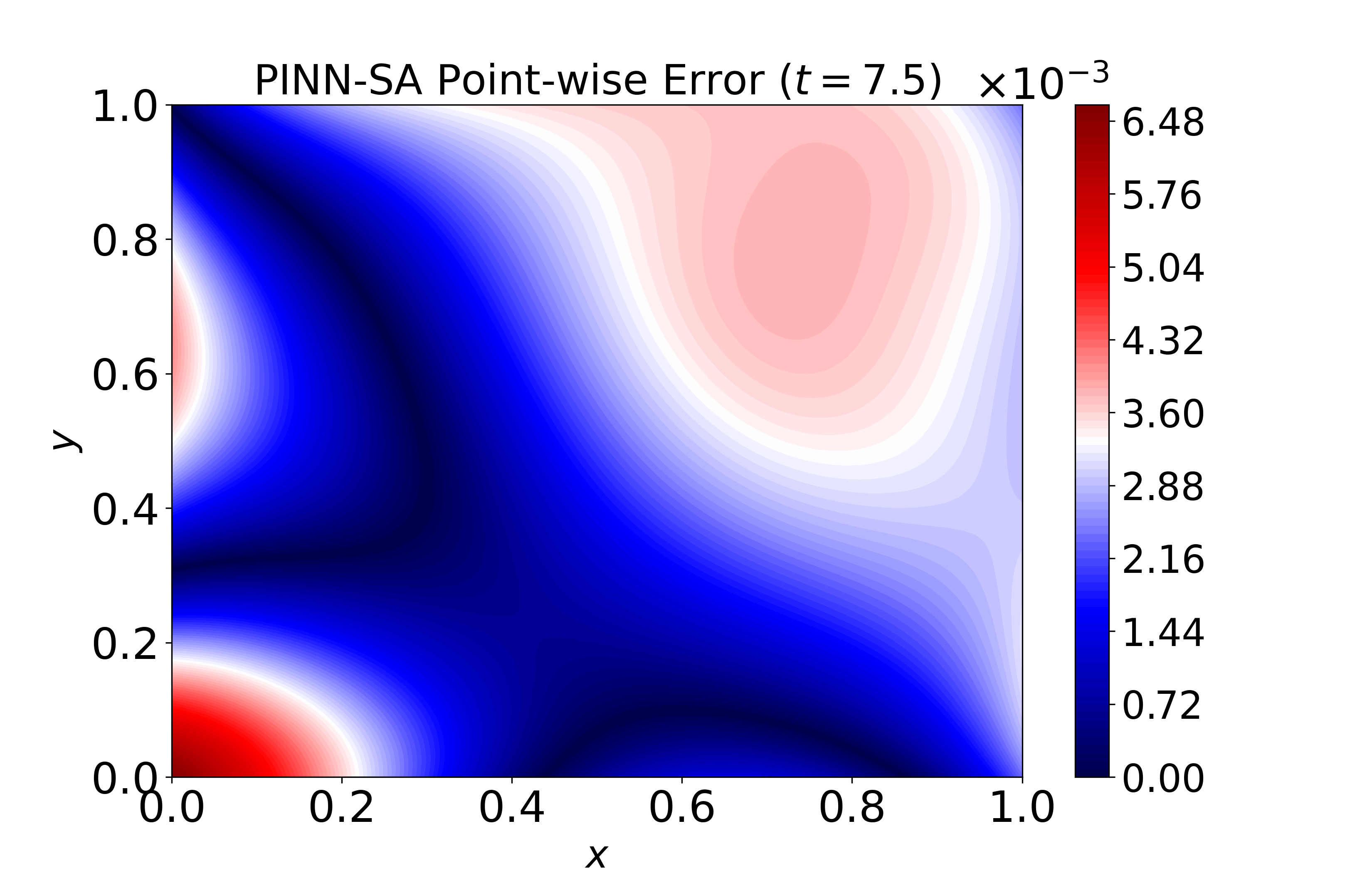}  
   \end{minipage}

   \begin{minipage}{0.33\textwidth}
     \centering
     \includegraphics[width=\linewidth]{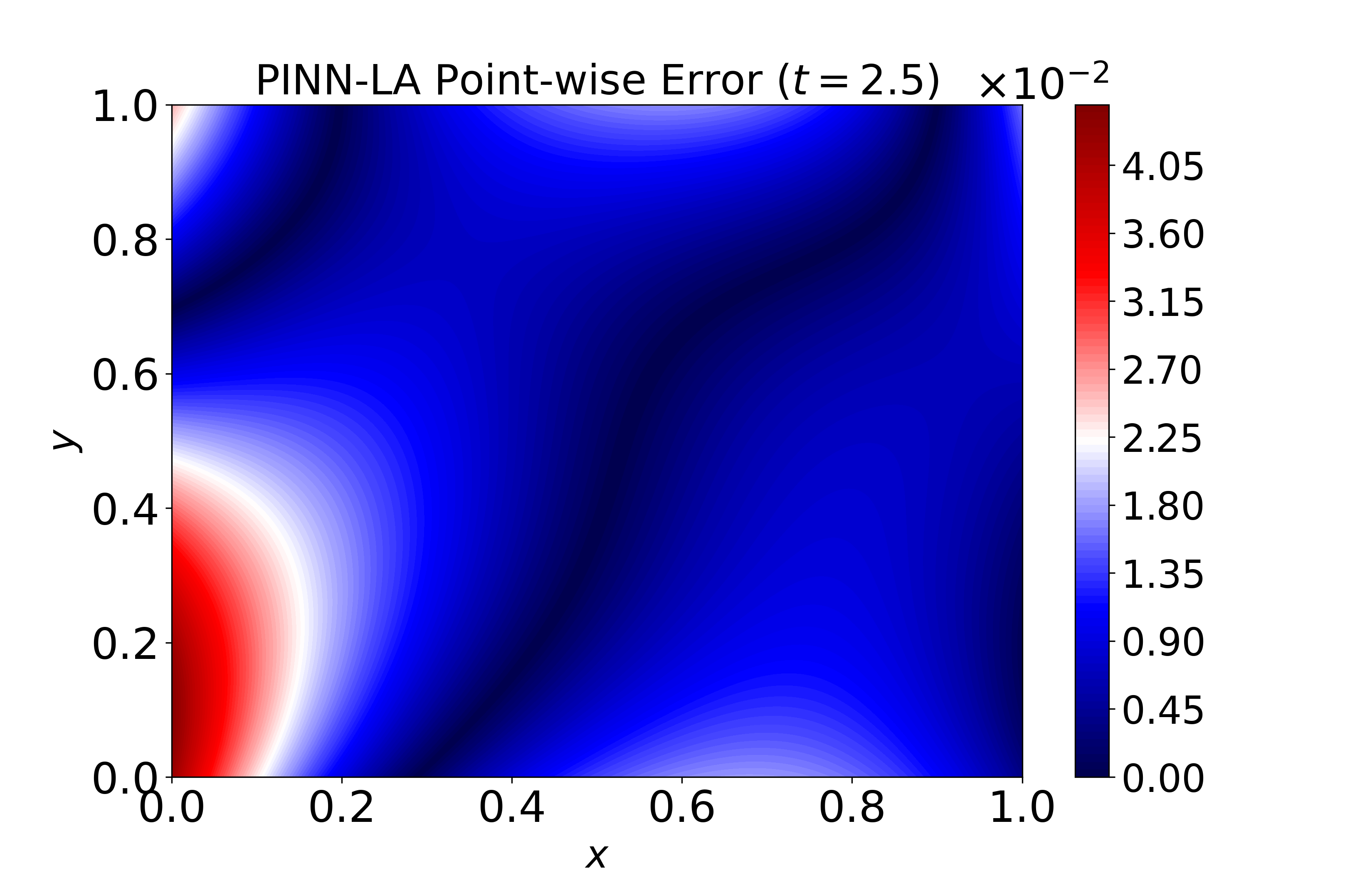} 
   \end{minipage}
    \begin{minipage}{0.33\textwidth}
     \centering
     \includegraphics[width=\linewidth]{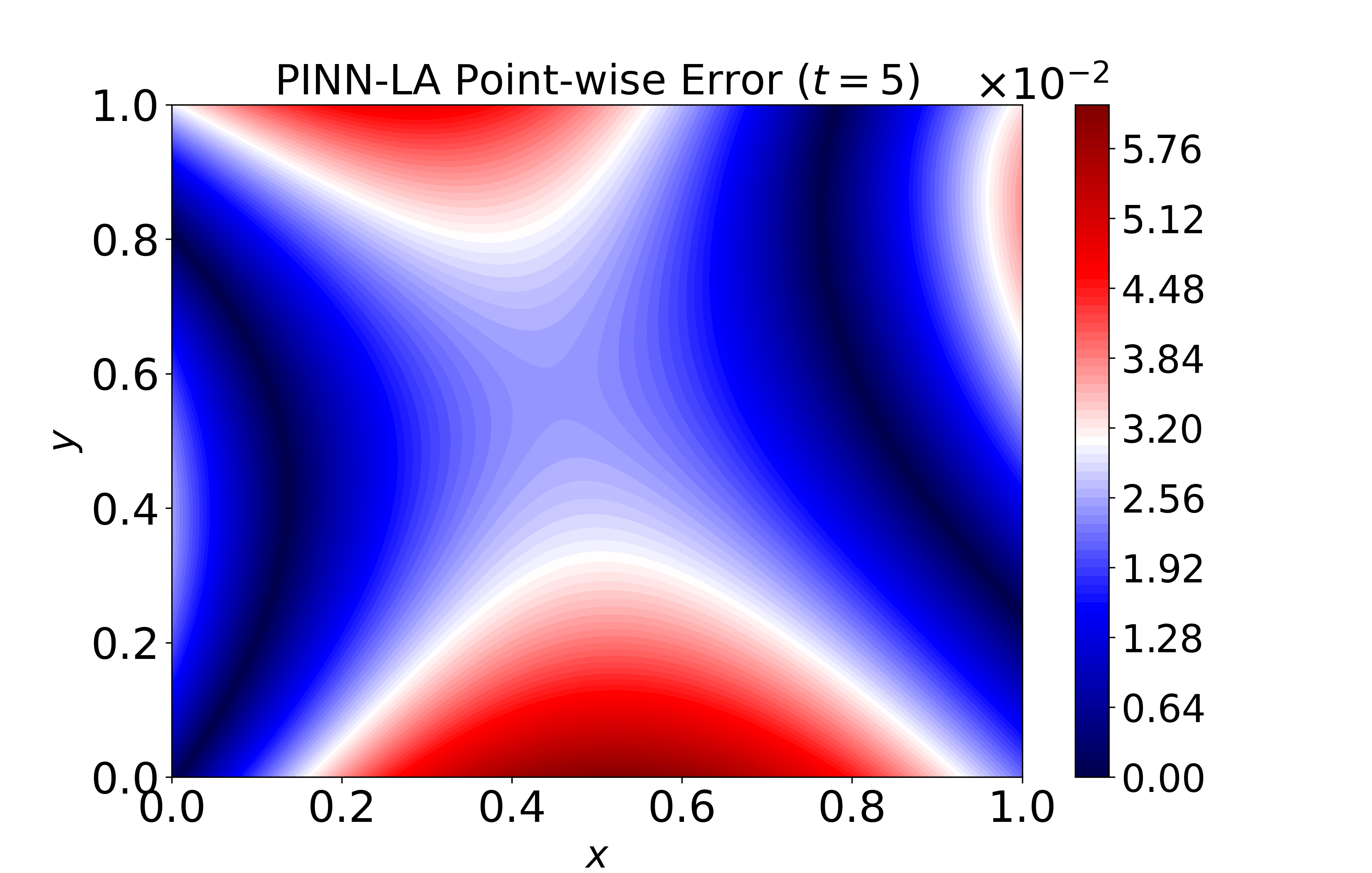}  
   \end{minipage}
   \begin{minipage}{0.33\textwidth}
     \centering
     \includegraphics[width=\linewidth]{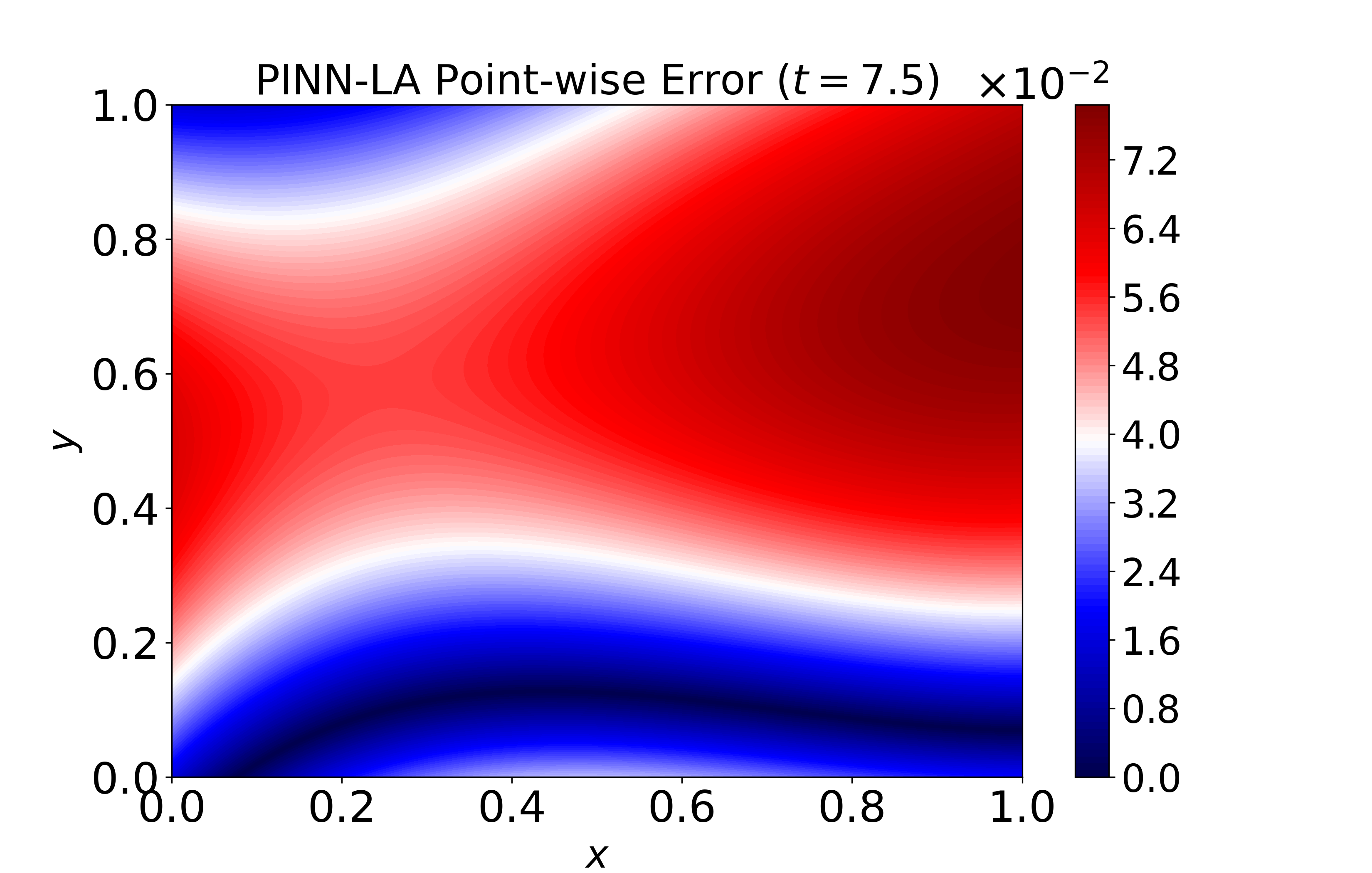}  
   \end{minipage}
    \caption{2D Klein–Gordon equation: Exact solution (top row) and point-wise error distributions for CWP (second row), RBA (third row), SA (fourth row), and LA (bottom row) at time slices \( t = 2.5 \) (left column), \( t = 5 \) (middle column), and \( t = 7.5 \) (right column).}\label{fig: 2D KG heatmap}
\end{figure}

To further illustrate the spatial distribution of model accuracy, Fig.~\ref{fig: 2D KG heatmap} presents point-wise error heatmaps for all methods at three representative time slices (\( t = 2.5 \), \( t = 5 \), and \( t = 7.5 \)). As shown, CWP (second row) consistently yields lower error magnitudes compared to RBA (third row), SA (fourth row), and LA (bottom row). While all methods show some discrepancies from the exact solution, CWP avoids the large localized error regions—visually indicated by saturated color spots—that are prevalent in the baseline methods. These visual patterns align well with both the error histories and the quantitative metrics, providing strong empirical support for CWP’s robustness and precision across time. Overall, the combination of quantitative and visual evidence confirms that CWP delivers significantly more accurate and stable predictions for this PDE.

  
\label{Sec 2D Klein-Gordon}

\subsection{Viscous Burgers equation}

This section evaluates the performance of the proposed CWP method on a PDE featuring a developing shock. Specifically, we consider the one-dimensional viscous Burgers equation:
\begin{subequations}
\begin{align}
&u_t + u u_x =\frac{0.01}{\pi} u_{xx}, \quad x\in\Omega,~t\in T,\\
&u(x, t) = 0,~~~~~~~~~~~~~~~\quad x\in\partial\Omega,~t\in T,\\
&u(x, 0) = -\sin(\pi x),~~~~~~~~~~~~~~\quad x\in\Omega,
\end{align}
\end{subequations}
where $\Omega = [-1,1]$ and $T = [0,1]$. 
As reported in~\cite{raissi2019physics}, a numerical solution to this problem reveals that while the solution remains smooth over much of the domain, a sharp shock forms near \(x=0\) as time progresses.
Accurately resolving this shock region is a known challenge for PINNs, due to the model's intrinsic bias toward smooth function approximation.


All models in this study share a consistent architecture of 7 hidden layers with 20 neurons per layer. To enforce the boundary and initial conditions via a hard constraint, we define the network output as
\[  
\hat{u}(t,x) = t(x-1)^2\hat{u}_{\mathcal{NN}}(t,x) - \sin(\pi x),  
\] 
where \(\hat{u}_{\mathcal{NN}}(t,x)\) is the raw neural network output. A total of \(N_f=10,000\) collocation points are used during training. All models are trained with the Adam optimizer using an initial learning rate of \( 5 \times 10^{-3} \), and an exponential decay scheduler with a decay factor of 0.7 applied every 1,000 iterations. The learning rate is held fixed once it reaches \( 1 \times 10^{-5} \), i.e., after 18,000 iterations, ensuring stable convergence during later training stages.

\begin{figure}[!htb]
\centering
    \begin{minipage}{0.45\textwidth}
     \centering
     \includegraphics[width=\linewidth]{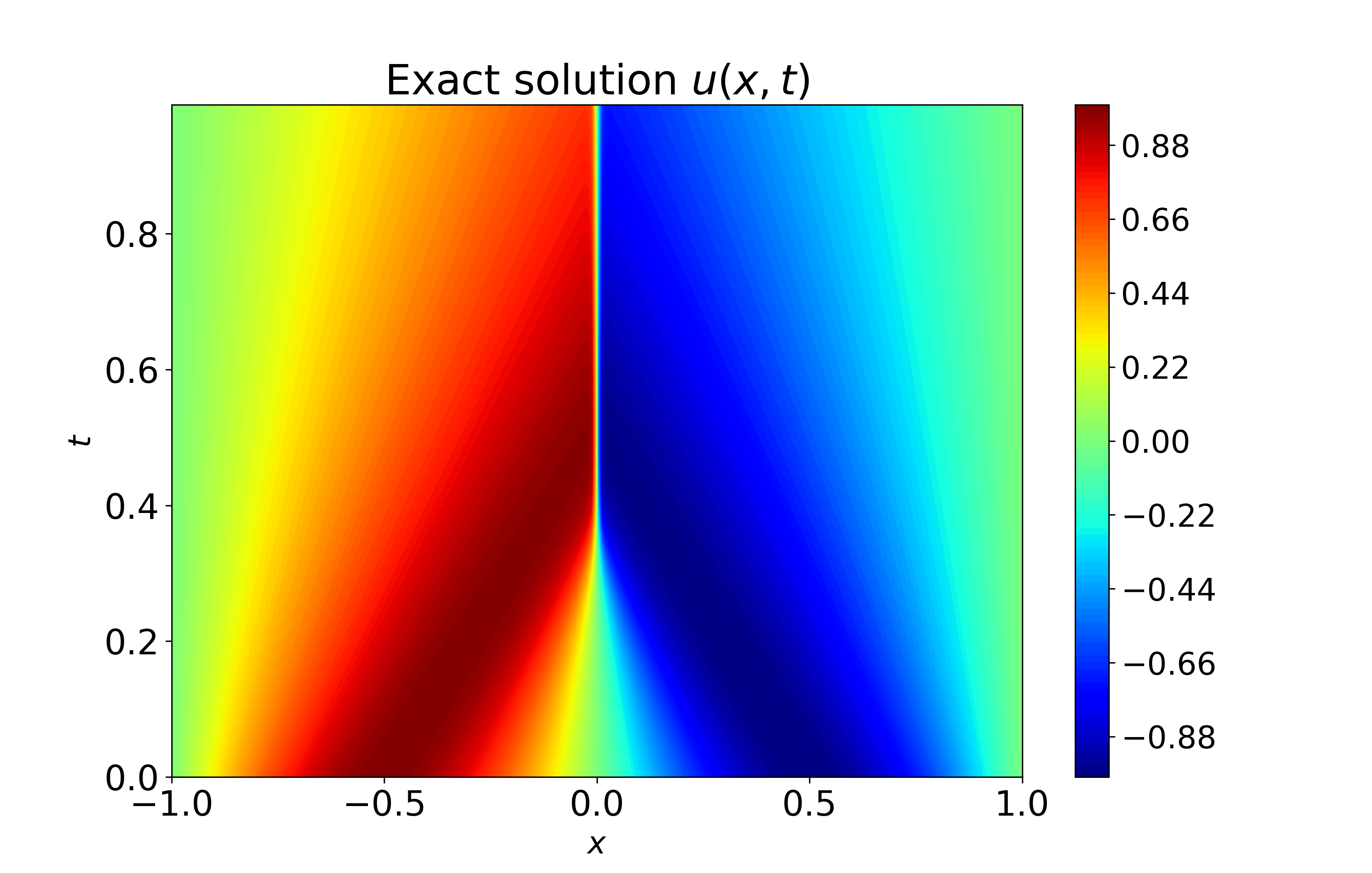} 
   \end{minipage}

   \begin{minipage}{0.45\textwidth}
     \centering
     \includegraphics[width=\linewidth]{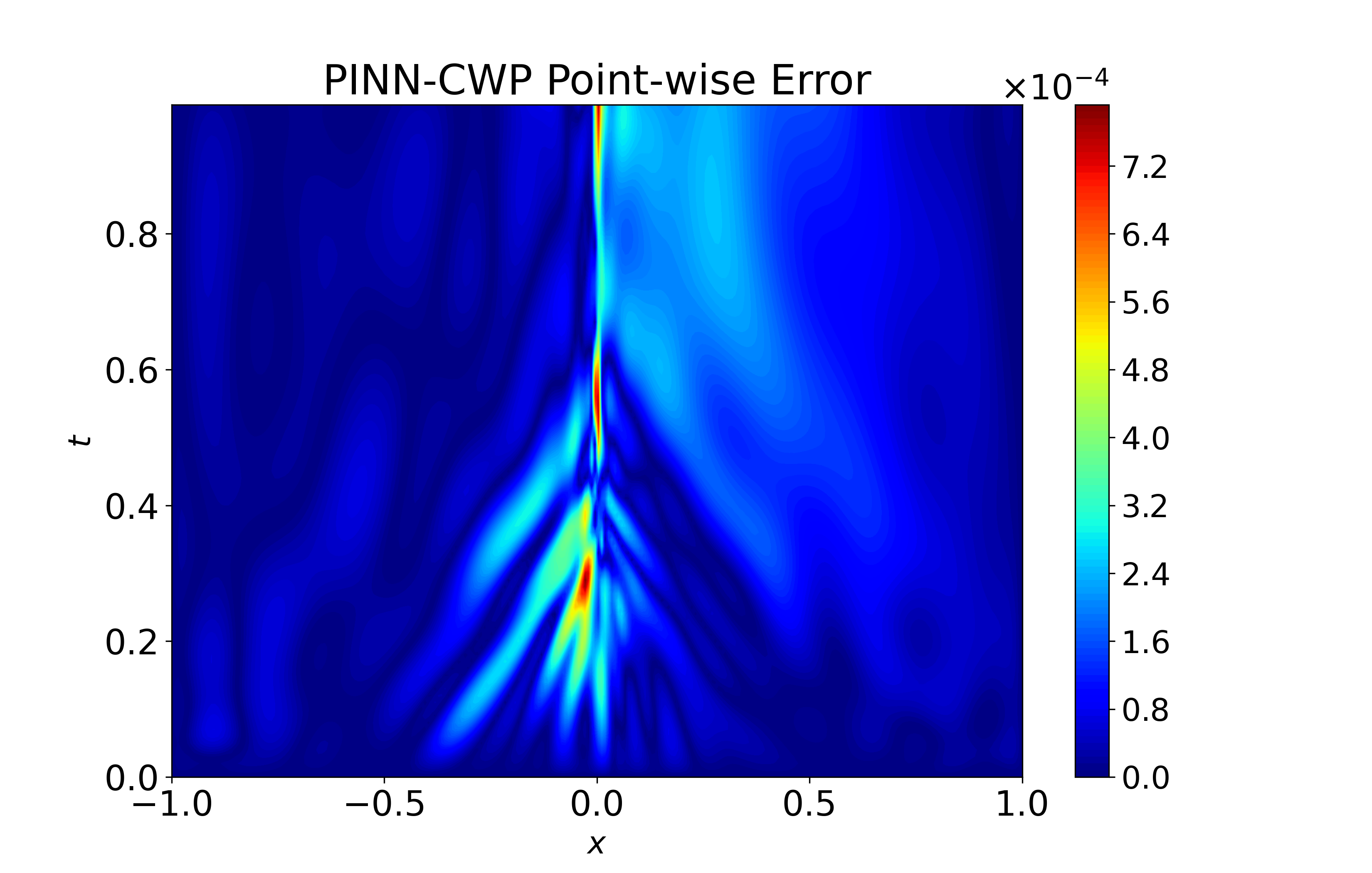} 
   \end{minipage}
    \begin{minipage}{0.45\textwidth}
     \centering
     \includegraphics[width=\linewidth]{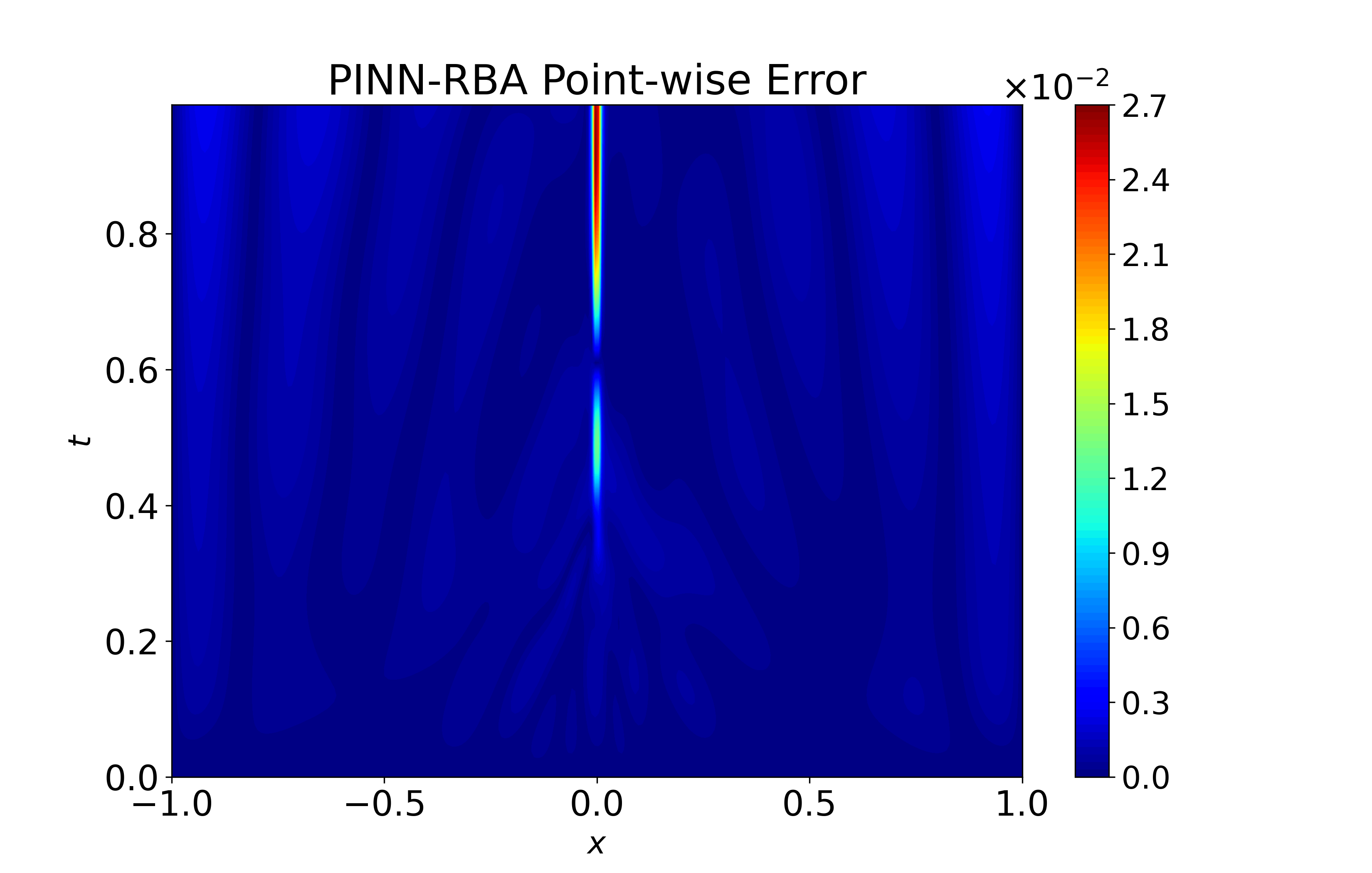}  
   \end{minipage}

   \begin{minipage}{0.45\textwidth}
     \centering
     \includegraphics[width=\linewidth]{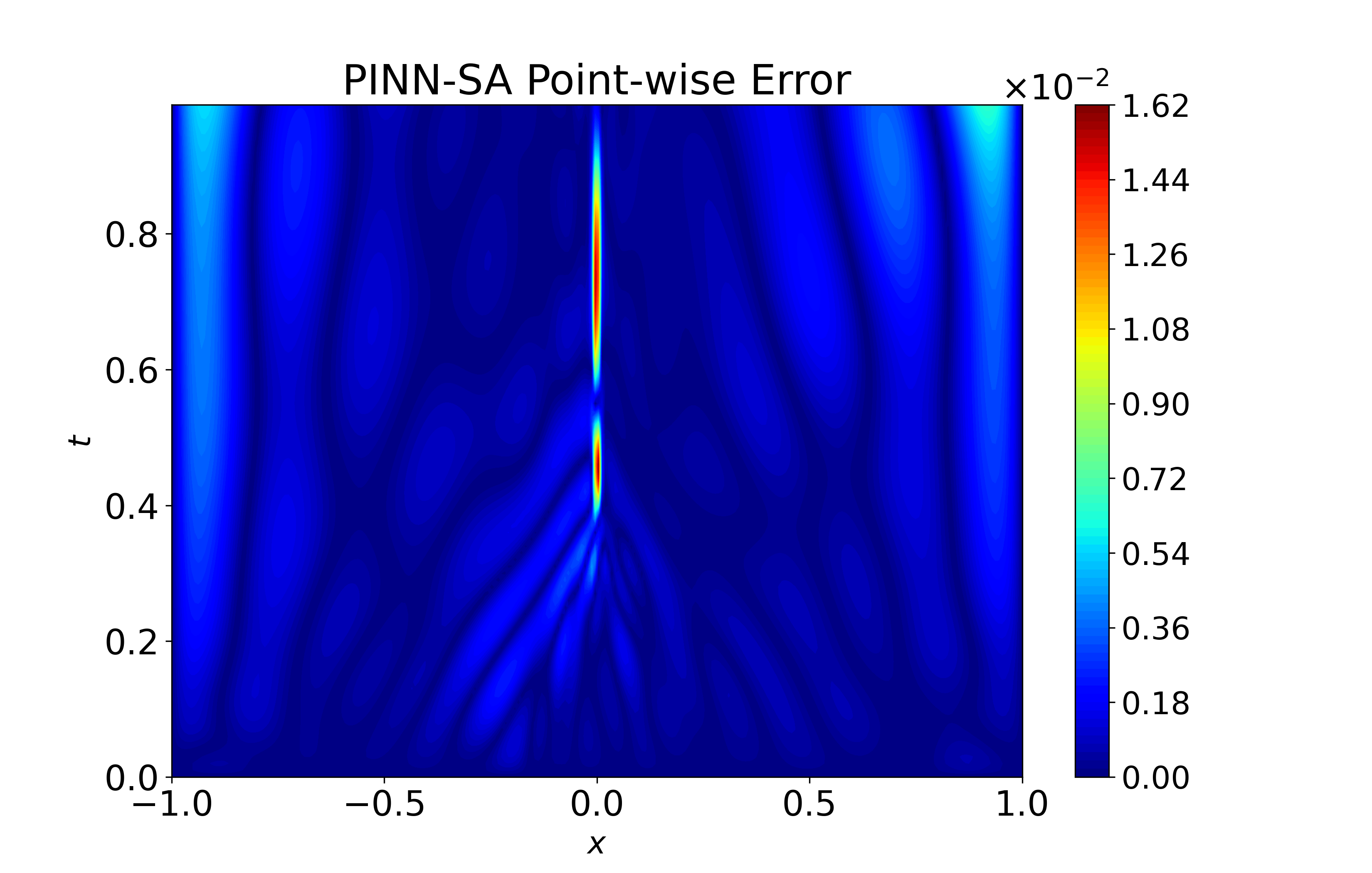} 
   \end{minipage}
    \begin{minipage}{0.45\textwidth}
     \centering
     \includegraphics[width=\linewidth]{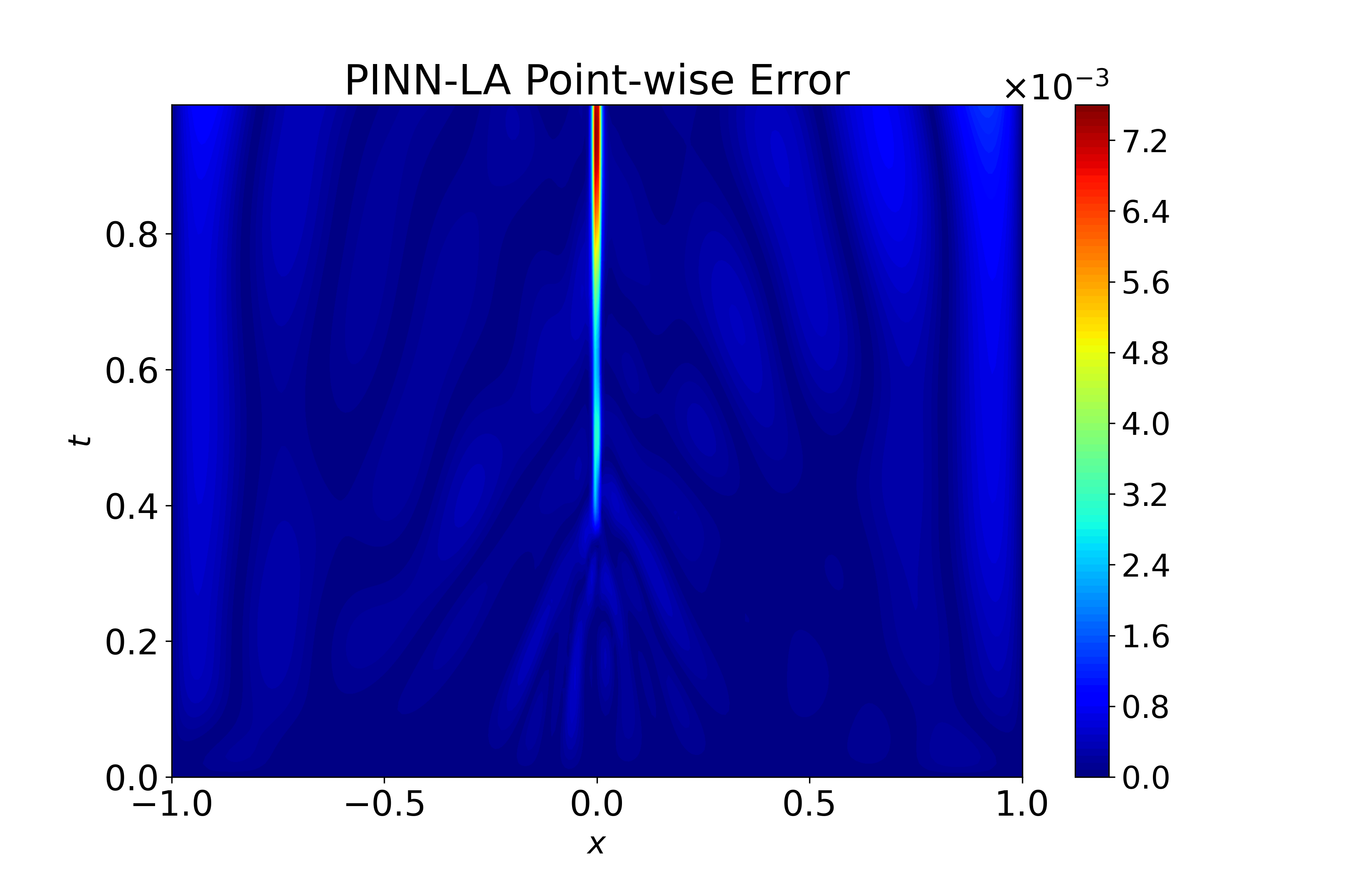}  
   \end{minipage}

    \begin{minipage}{0.45\textwidth}
     \centering
     \includegraphics[width=\linewidth]{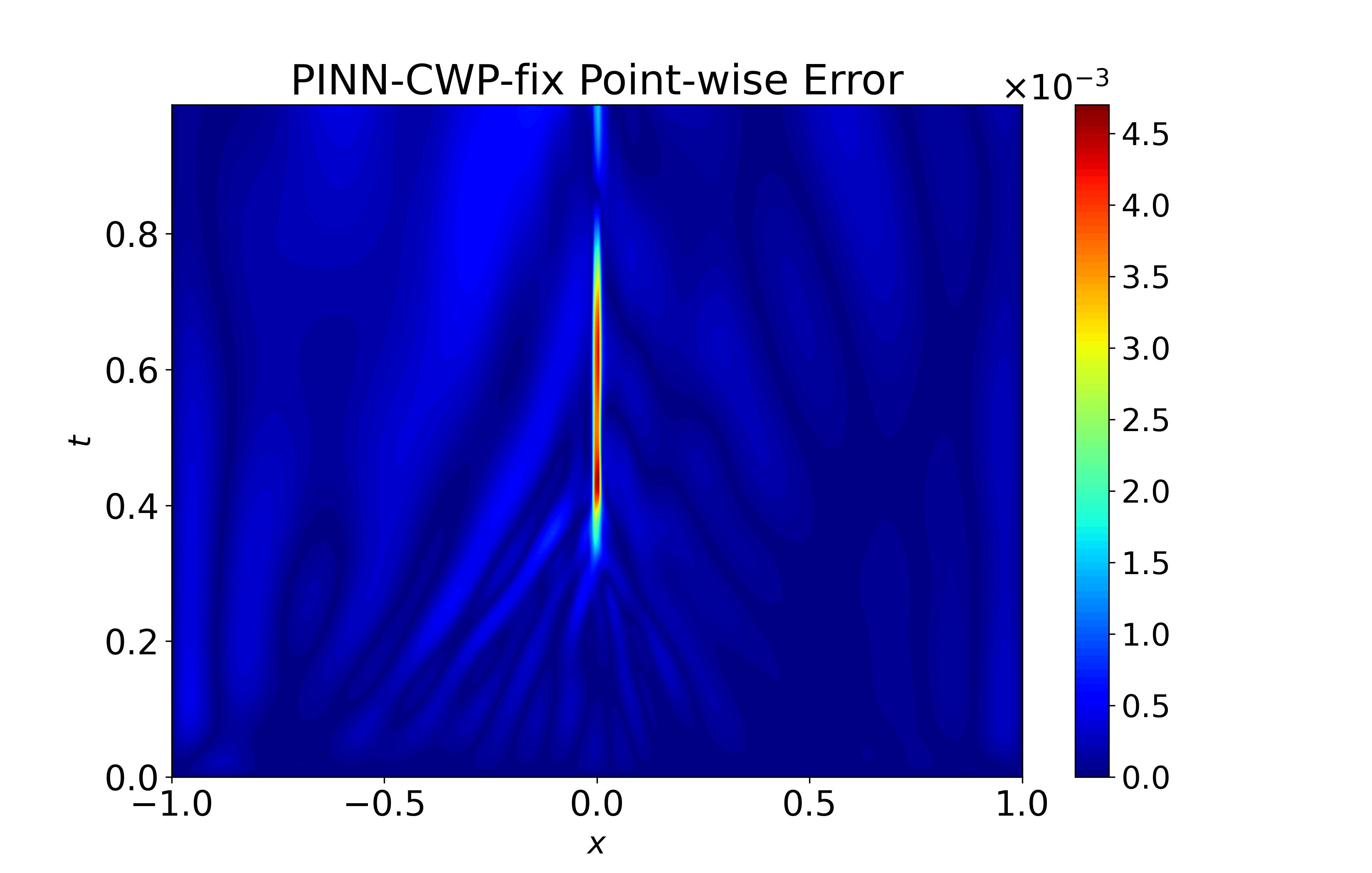} 
   \end{minipage}   
    \begin{minipage}{0.45\textwidth}
     \centering
     \includegraphics[width=\linewidth]{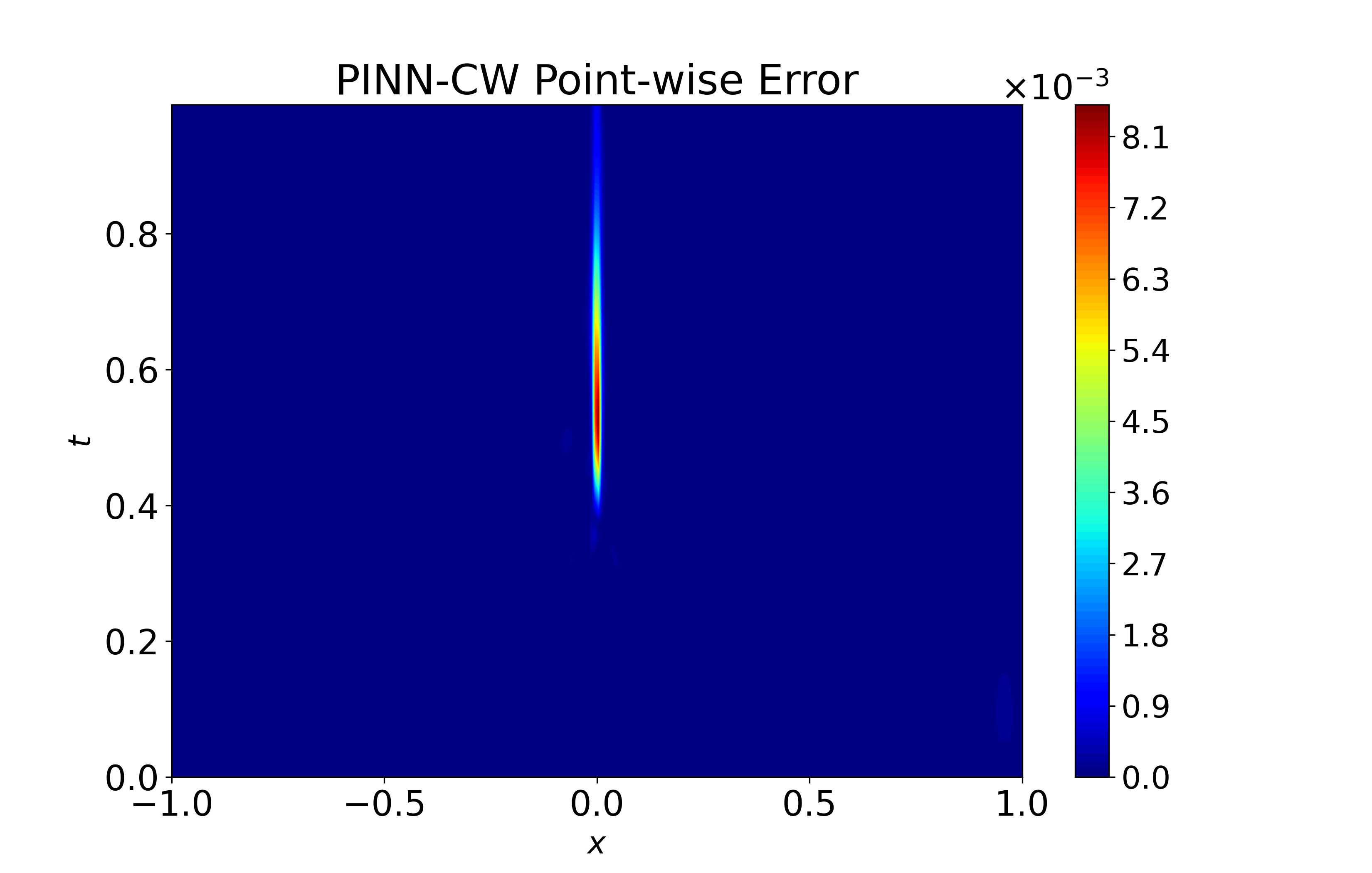}  
   \end{minipage}
    \caption{Exact solution (top row) and point-wise error maps for CWP, RBA, SA, and LA (second to fifth rows, respectively). CWP yields the lowest point-wise errors near the shock—a region characterized by sharp gradients and high solution complexity. In the ablation study, CW outperforms RBA, SA, and LA, underscoring the benefit of incorporating convolution-based weighting. Furthermore, CWP surpasses both CW and CWP-fix, demonstrating that the full integration of convolutional weighting and adaptive resampling is critical for achieving optimal accuracy.}\label{fig: Burgers heatmap}
\end{figure}

Fig.~\ref{fig: Burgers heatmap} presents the point-wise error distributions for each method. CWP (second row, first column) demonstrates markedly improved accuracy in the shock region compared to the baseline methods RBA, SA, and LA. While all methods perform adequately in smooth regions, the error heatmaps clearly show that RBA, SA, and LA struggle to capture the sharp transition near 
$x=0$, resulting in significantly larger errors. In contrast, CWP maintains lower error magnitudes throughout, particularly in the critical shock region, underscoring its robustness and effectiveness in handling solutions with discontinuities.

To further dissect the contributions of CWP’s components, we include the ablated variants CW and CWP-fix in the comparison. As shown in the last row of Fig.~\ref{fig: Burgers heatmap}, both variants improve upon the baseline methods but still exhibit larger errors near the shock compared to full CWP. These results highlight that the combination of localized convolutional weighting and adaptive resampling employed in CWP plays a vital role in stabilizing training and enhancing predictive accuracy in regions characterized by sharp gradients or discontinuities.


\begin{figure}[!htb]
\centering
    \begin{minipage}{0.45\textwidth}
     \centering
     \includegraphics[width=\linewidth]{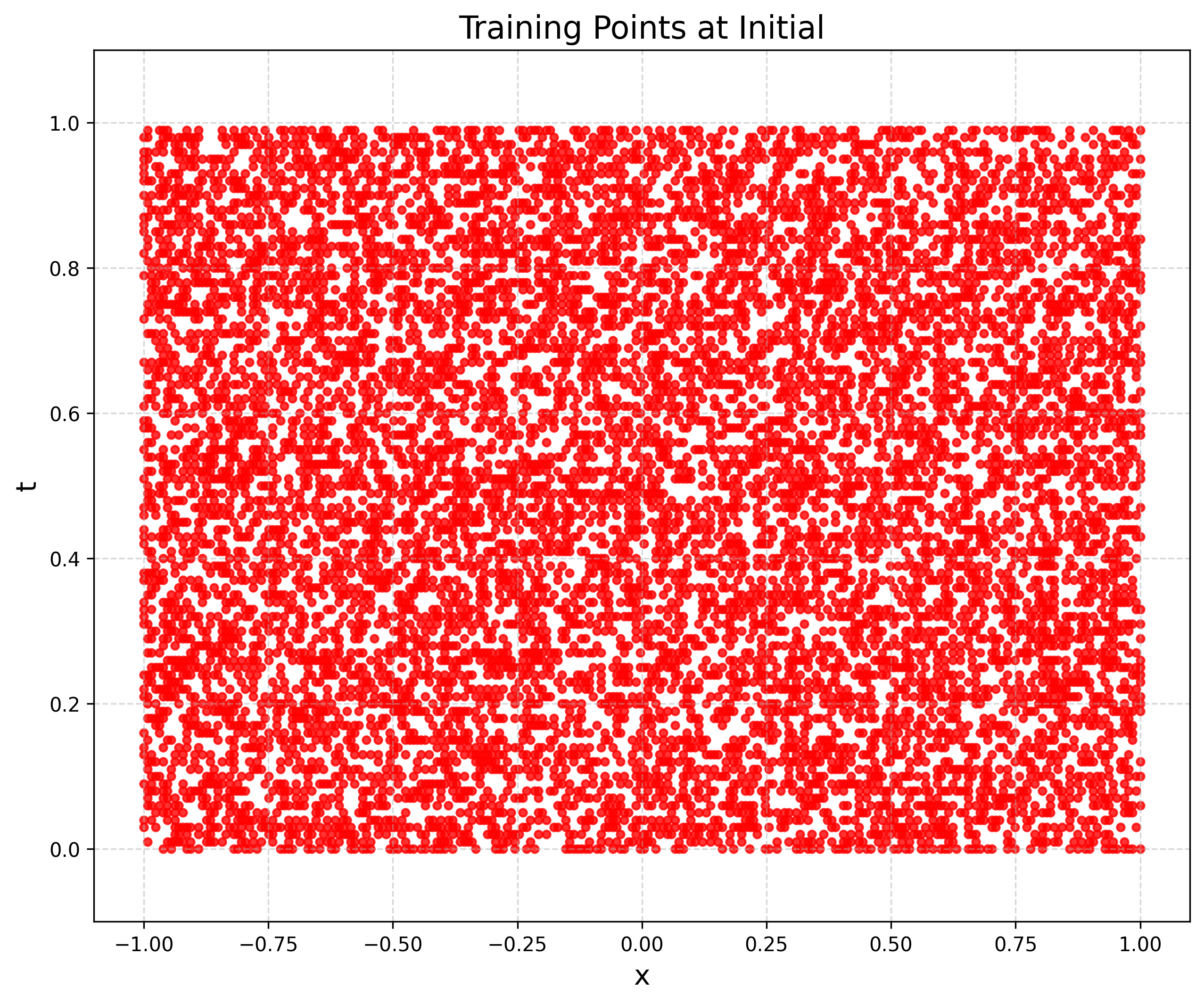} 
   \end{minipage}   
   \begin{minipage}{0.45\textwidth}
     \centering
     \includegraphics[width=\linewidth]{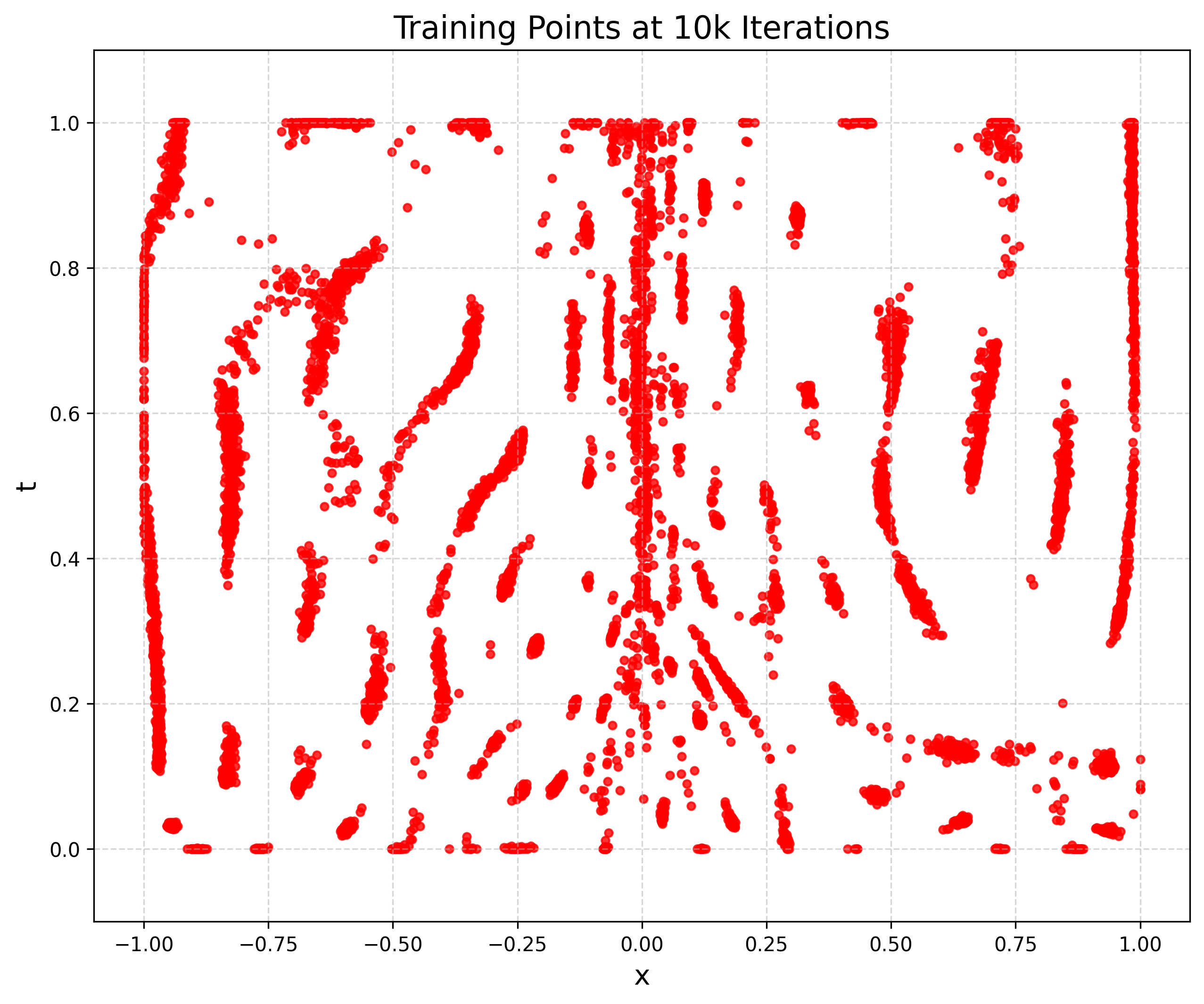} 
   \end{minipage}
   
   \begin{minipage}{0.45\textwidth}
     \centering
     \includegraphics[width=\linewidth]{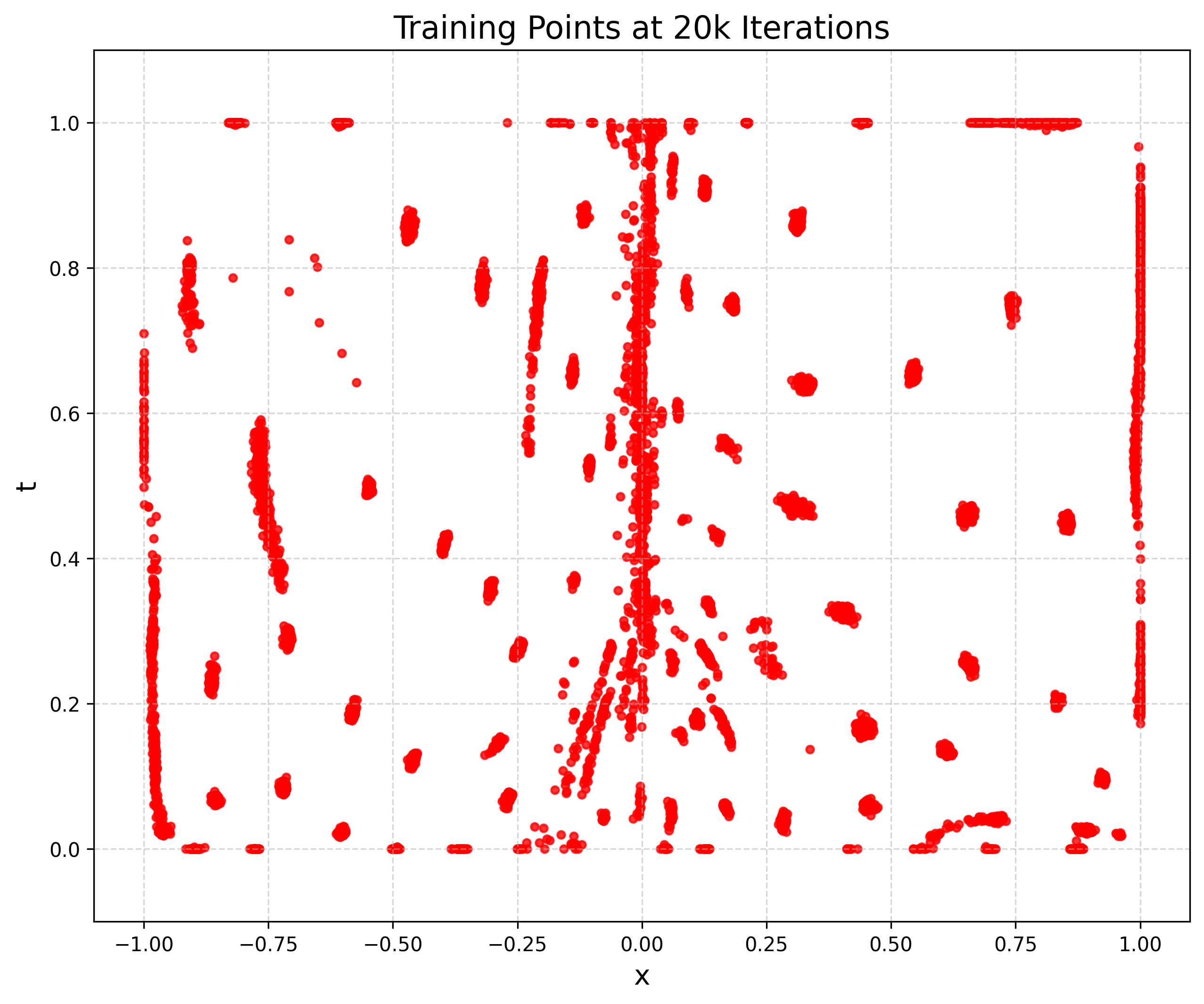} 
   \end{minipage}
   \begin{minipage}{0.45\textwidth}
     \centering
     \includegraphics[width=\linewidth]{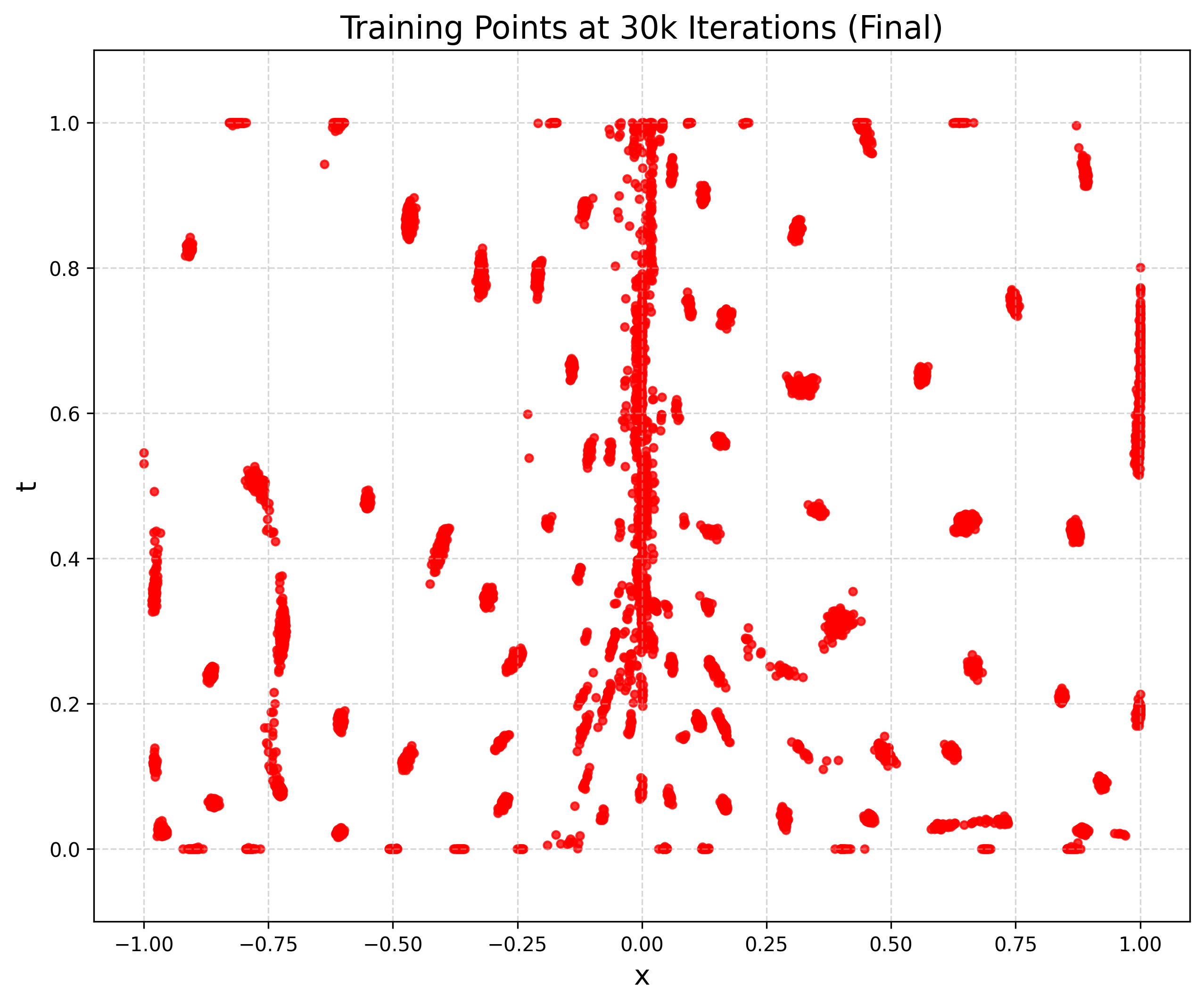} 
   \end{minipage}
   \caption{Resampling trajectory of collocation points for the viscous Burgers equation at four training stages: (a) initial random distribution; (b) after 10k iterations, showing early concentration near the shock at \(x = 0\); (c) after 20k iterations, with increasingly dense clustering in high-residual regions; (d) after 30k iterations, illustrating the final adaptive focus on the shock while reducing point density in smooth areas. These dynamics highlight the effect of the CWP's convolutional weighting strategy, which promotes spatially coherent sampling in regions with persistently high error.} \label{fig: Burgers training points}
\end{figure}

The resampling trajectory plots in Fig.~\ref{fig: Burgers training points} provide visual confirmation of CWP’s adaptive sampling mechanism. Initially, training points are randomly distributed, but by iteration 30,000, a significant concentration emerges near the shock location at $x = 0$. This spatial shift aligns well with the point-wise error distribution shown in Fig.~\ref{fig: Burgers heatmap}. It explains why CWP achieves nearly an order of magnitude lower $L^2$ error than the baseline methods RBA, SA, and LA. Notably, CWP’s convolution-based weighting ensures that resampling is informed rather than heuristic—sampling decisions are guided by both the residual magnitude and its spatial neighborhood, effectively suppressing high-frequency noise while maintaining sensitivity to discontinuities. These results underscore the value of continuous, physics-aware weighting in PINNs, especially for problems characterized by multiscale features such as shocks.


\begin{figure}[!htb]
\centering
    \begin{minipage}{0.45\textwidth}
     \centering
     \includegraphics[width=\linewidth]{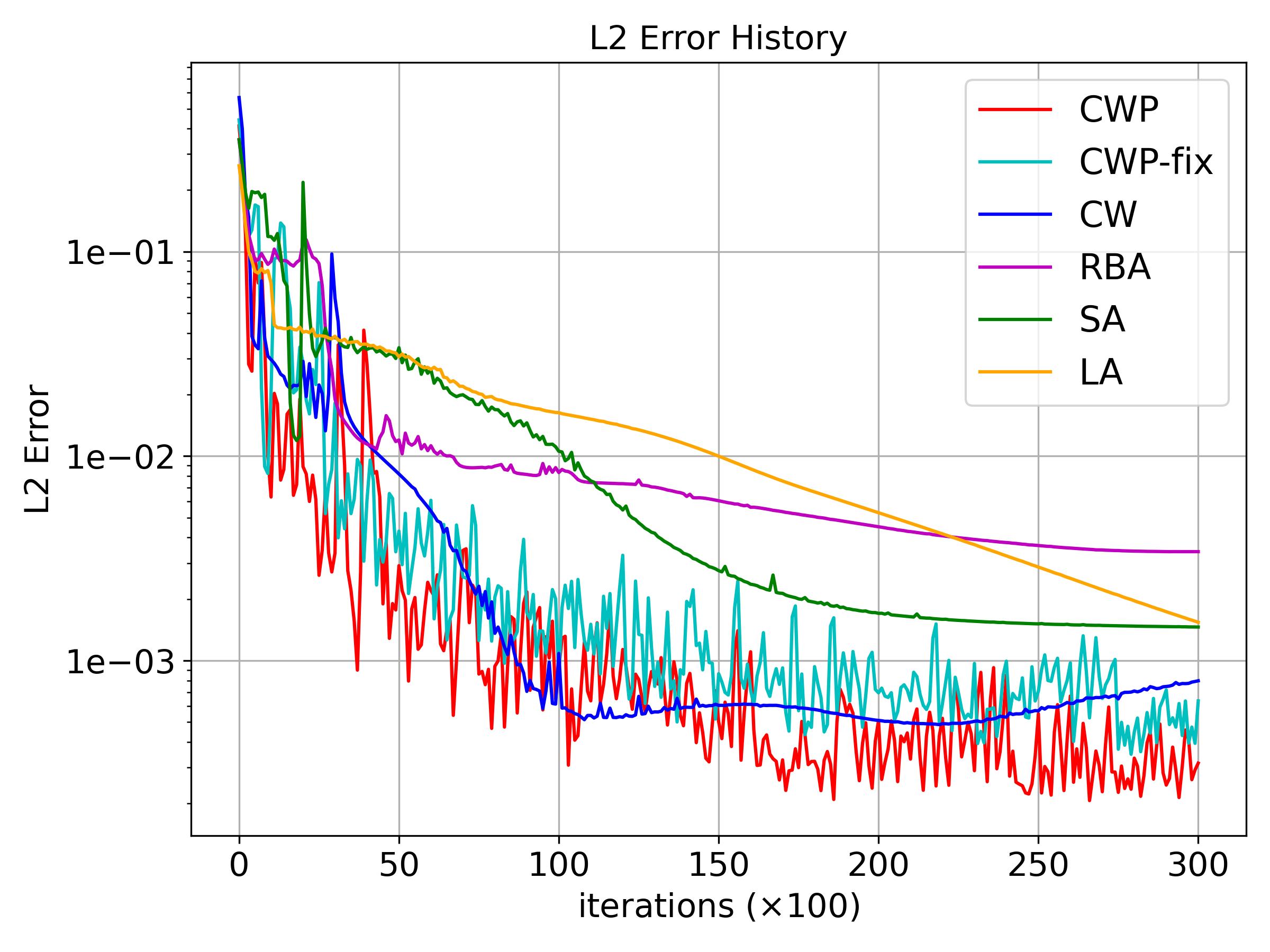} 
   \end{minipage}
   \caption{Training history of the relative \( L^2 \) error for different weighting algorithms on the viscous Burgers equation. }\label{fig: Burgers HIST}
\end{figure}

The convergence behavior across methods is further detailed in Fig.~\ref{fig: Burgers HIST}, which depicts the relative $L^2$ error over training iterations. Among all models, CWP reaches the lowest final error of approximately $3\times 10^{-4}$, while all baseline methods remain above $1\times 10^{-3}$ under identical training settings and architectures.

\begin{figure}[!htb]
\centering
    \begin{minipage}{0.85\textwidth}
     \centering
     \includegraphics[width=\linewidth]{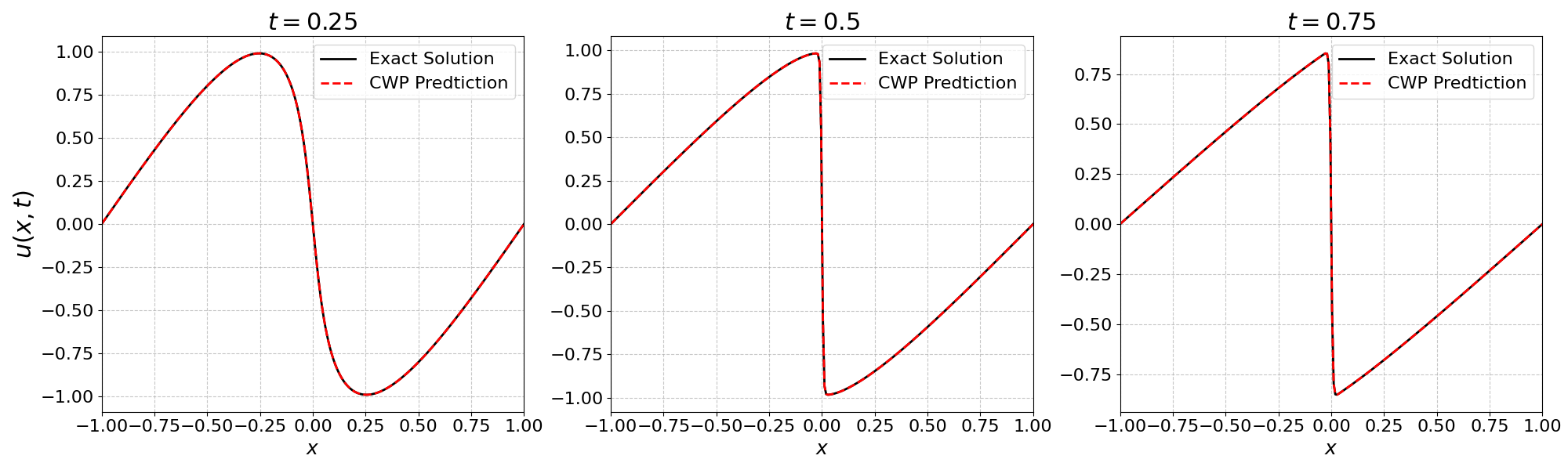} 
   \end{minipage}
   
   \begin{minipage}{0.85\textwidth}
     \centering
     \includegraphics[width=\linewidth]{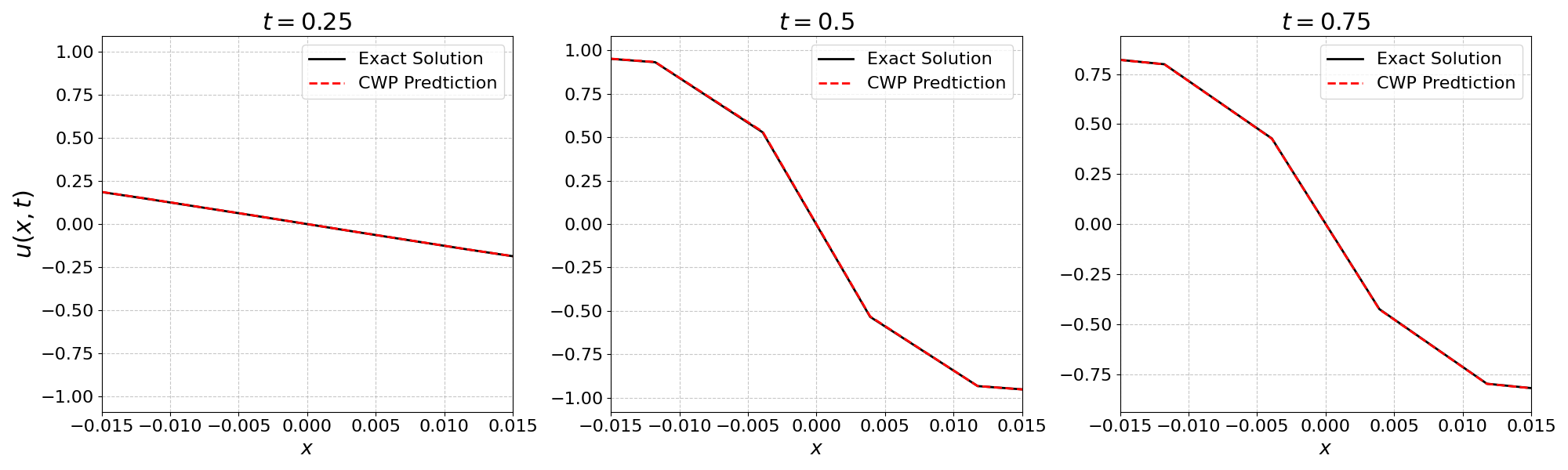} 
   \end{minipage}
    \caption{Comparison of the exact solution (black solid) and CWP predictions (red dashed) at $t = 0.25$, $0.5$, and $0.75$. \textbf{Top row:} Full-domain views capturing overall solution trends. \textbf{Bottom row:} Zoomed-in views around $x = 0$, highlighting prediction accuracy in the shock region. The visual comparison demonstrates CWP’s effectiveness in accurately resolving sharp gradients in the viscous Burgers equation.}\label{fig: Burgers slices}
\end{figure}

To further assess the quality of CWP’s predictions, Fig.~\ref{fig: Burgers slices} compares its output to the exact solution at three time instances: $t = 0.25$, $0.5$, and $0.75$. The top row displays the predicted and true solutions over the entire spatial domain, highlighting global accuracy. The bottom row provides zoomed-in views near $x = 0$, focusing on the model’s performance in capturing the shock. The solid black line denotes the ground truth, while the dashed red line represents CWP’s prediction. These plots clearly show that CWP tracks the sharp gradients and discontinuities with high fidelity, further validating the method’s robustness in handling nonlinear PDEs with shock features.


\label{Sec Burgers}

\subsection{Unsteady Cylinder Flow}
\label{Sec NS}
The Navier–Stokes (NS) equations stand as cornerstone frameworks for tackling an extensive array of real-world challenges rooted in the study of fluid motion. In this section, we focus on the incompressible flows passing through a cylinder, which is a unsteady 2D NS equation as follows:
\begin{align}
\frac{\partial \mathbf{u}}{\partial t} + \mathbf{u} \cdot \nabla \mathbf{u} &= -\frac{1}{\rho} \nabla p + \nu \nabla^{2} \mathbf{u},  \\
\nabla \cdot \mathbf{u} &= 0,
\end{align}
where $\mathbf{u} = (u, v)$ represents the time-dependent velocity field. We choose the fluid density $\rho = 1$ and $\nu = \dfrac{1}{Re}$, targeting a high Reynolds number $Re = 3,900$. The Reynolds number, defined as $Re \equiv \frac{U_{\infty} D}{\nu} = 3900$, with free-stream velocity ${U_{\infty}} = 1$ and cylinder diameter $D = 1$, characterizes flow turbulence and complexity. In this case, the flow exhibits strongly unsteady, chaotic behavior with irregular vortex shedding and transitional turbulence—a regime notoriously challenging for PINNs.

\begin{figure}[!htb]
\centering
    \begin{minipage}{0.8\textwidth}
     \centering
     \includegraphics[width=\linewidth]{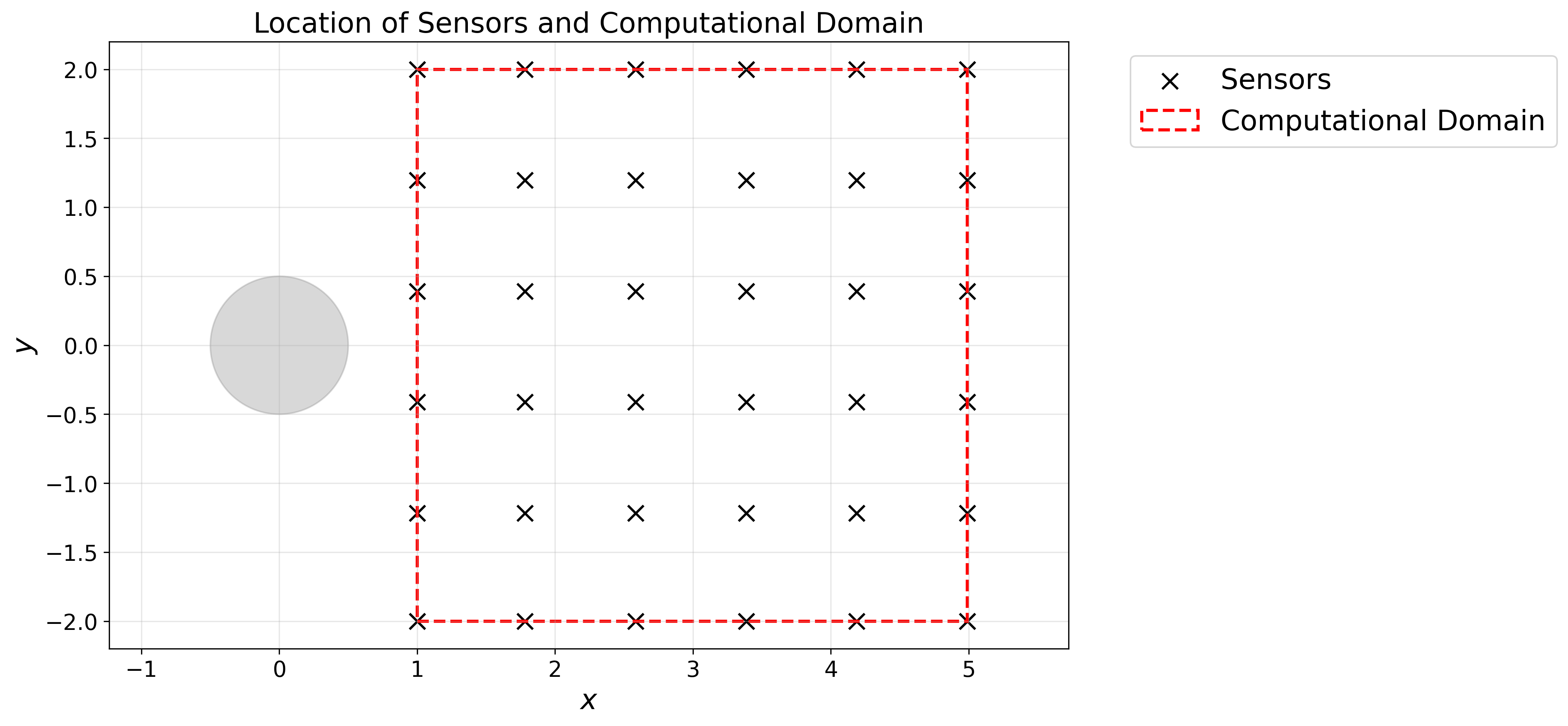} 
   \end{minipage}
    \caption{Location of sensors (marked with black crosses) in the computational domain ($\Omega = [1,5]\times [-2,2]$). The gray circle represents a cylinder with a diameter of $D = 1$ and is centered at the origin. These sensors capture observed data across 100 snapshots, which are used for training the model.}\label{fig: NS sensors}
\end{figure}

We follow the same experimental settings from the public repository \cite{Shengfenggithub}: The computational spatial domain is set to be $(x.y)\in\Omega = [1,5]\times [-2,2]$ and the temporal domain $t\in[0,42.93]$. The cylinder is centered at $(0,0)$ with diameter $D = 1$. Rather than relying on boundary or initial condition data, often exploited in PINNs to simplify training, we use sparse in-domain measurements: 36 sensors collect 100 snapshots each, yielding 3,600 labeled data points for the data loss term. This setup mimics real-world scenarios where boundary data may be inaccessible, testing the model’s ability to generalize from limited internal observations. The sensor placement is visualized in Fig. \ref{fig: NS sensors}.

For the physics-informed loss, we set $N_f = 10,000$ collocation points to enforce the NS equations. The global weights for data loss and residual loss are $\lambda_{\text{data}} = 10$ and $\lambda_F = 1$ to balance the contribution of the two loss terms. The network architecture consists of 10 hidden layers, each with 32 neurons. We train the model using the Adam optimizer with an initial learning rate of $0.001$, paired with an exponential learning rate scheduler. The scheduler applies a decay factor of $0.999$ every $100$ iterations.

\begin{figure}[!htb]
\centering
    \begin{minipage}{0.31\textwidth}
     \centering
     \includegraphics[width=\linewidth]{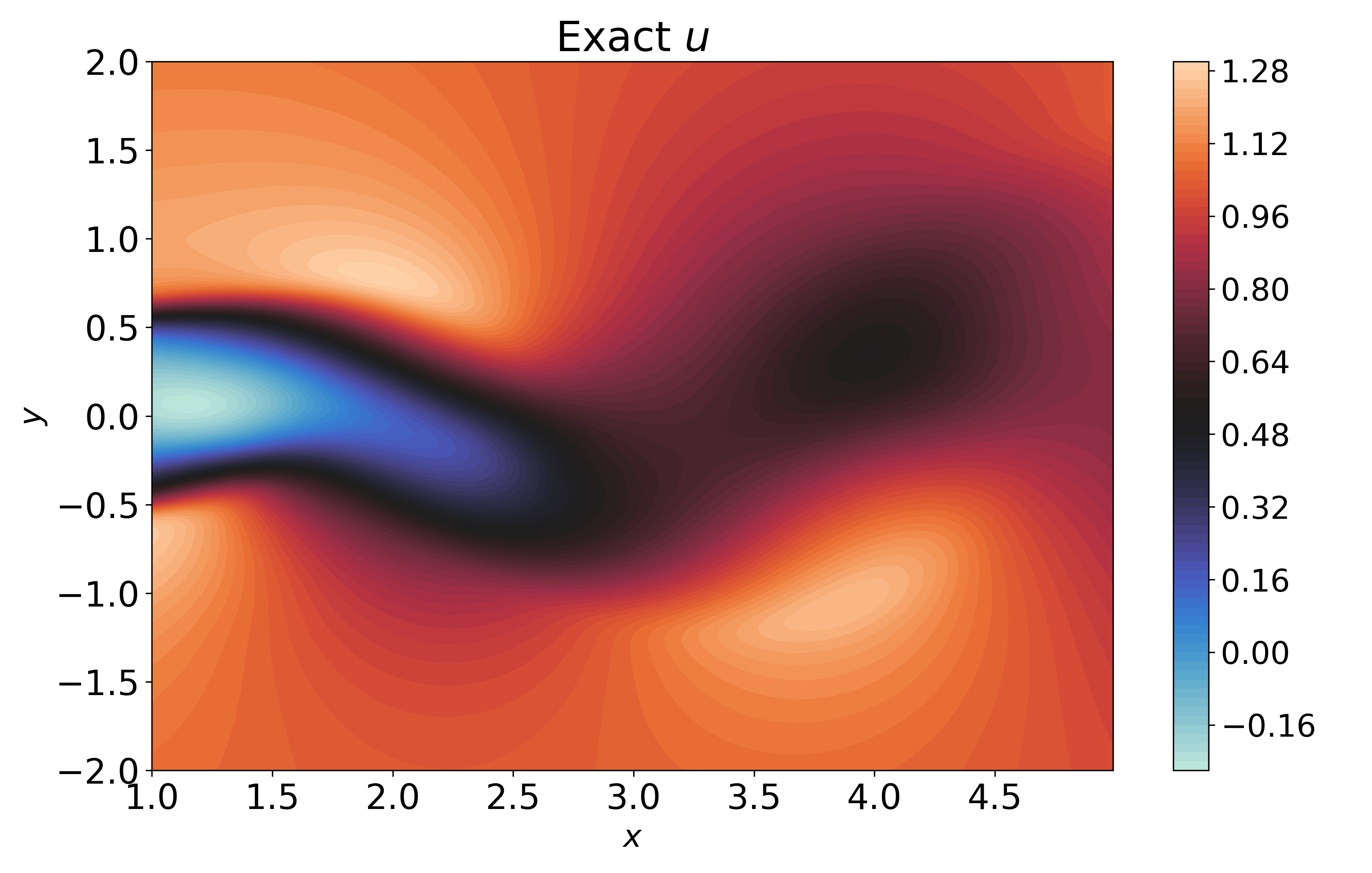} 
   \end{minipage}
   \begin{minipage}{0.31\textwidth}
     \centering
     \includegraphics[width=\linewidth]{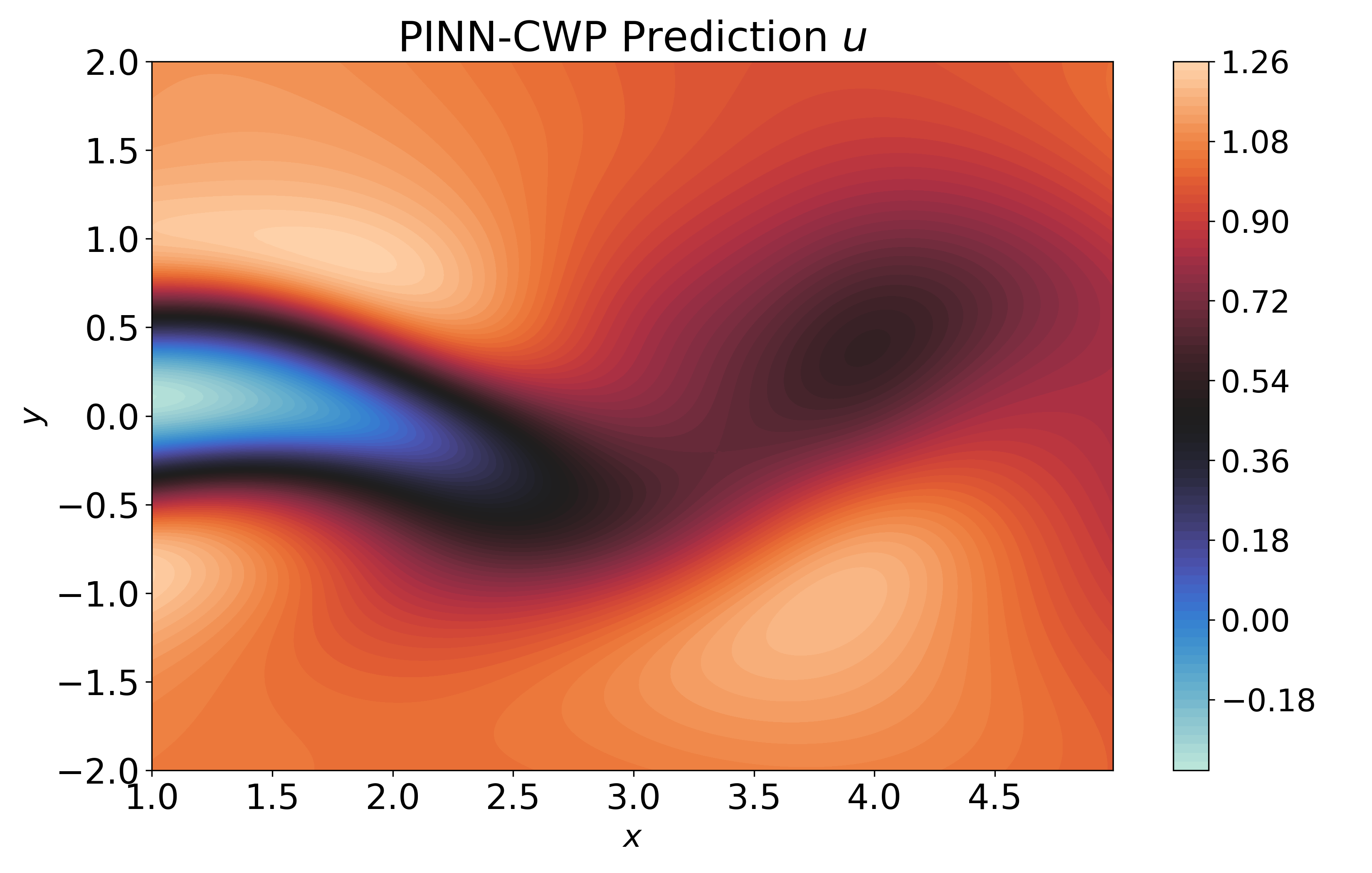} 
   \end{minipage}
   \begin{minipage}{0.31\textwidth}
     \centering
     \includegraphics[width=\linewidth]{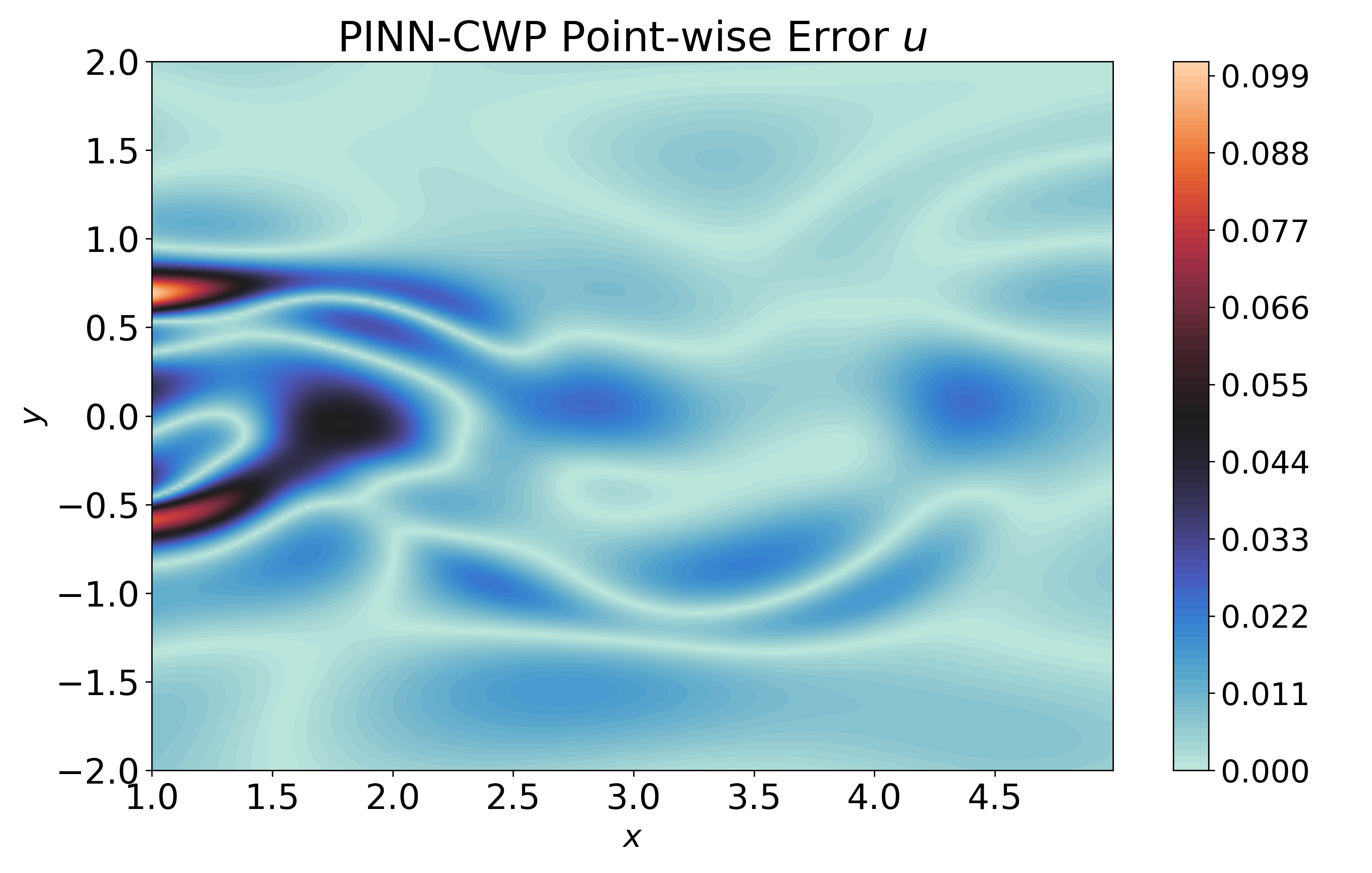} 
   \end{minipage}

   \begin{minipage}{0.31\textwidth}
     \centering
     \includegraphics[width=\linewidth]{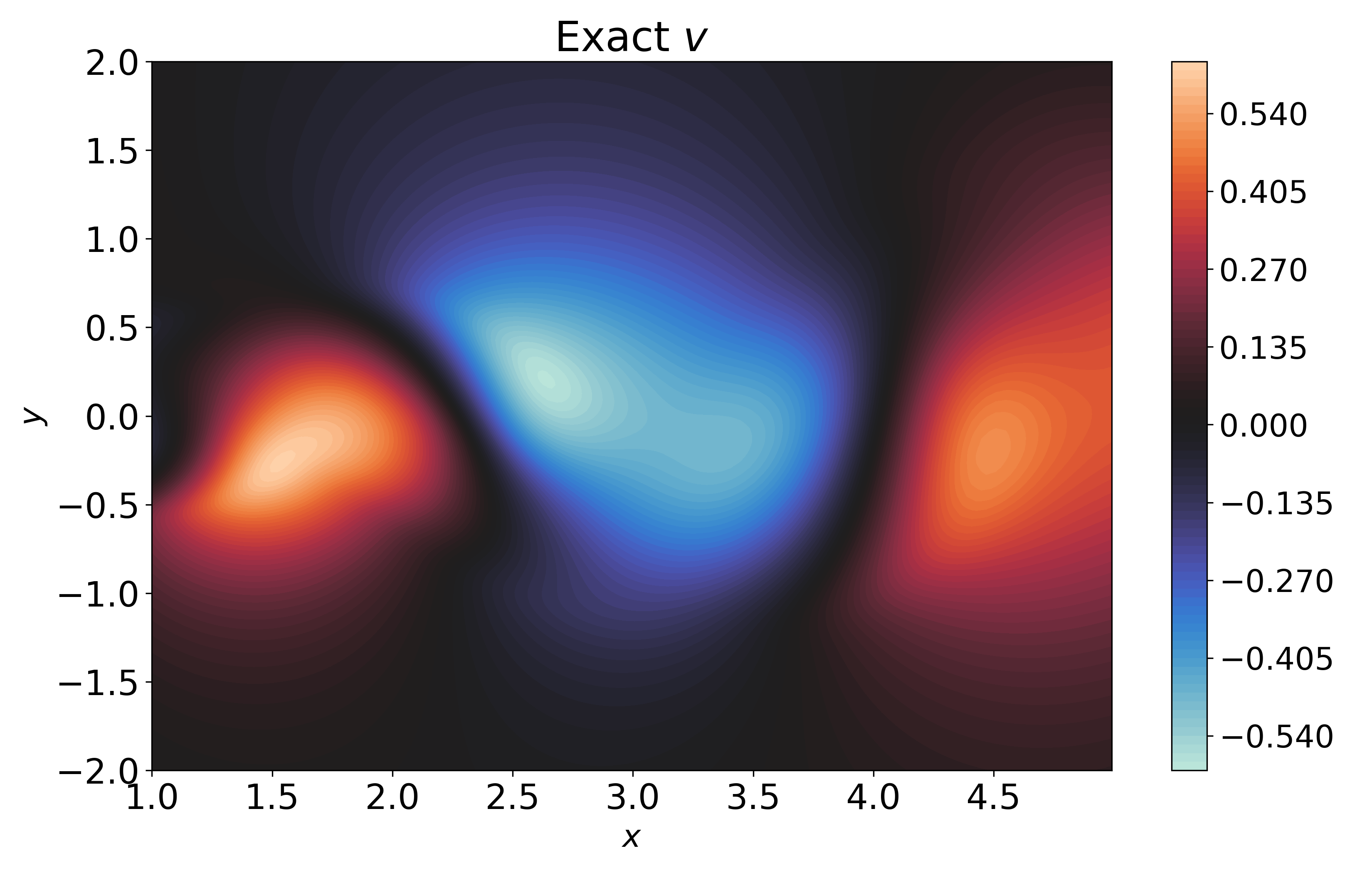} 
   \end{minipage}
   \begin{minipage}{0.31\textwidth}
     \centering
     \includegraphics[width=\linewidth]{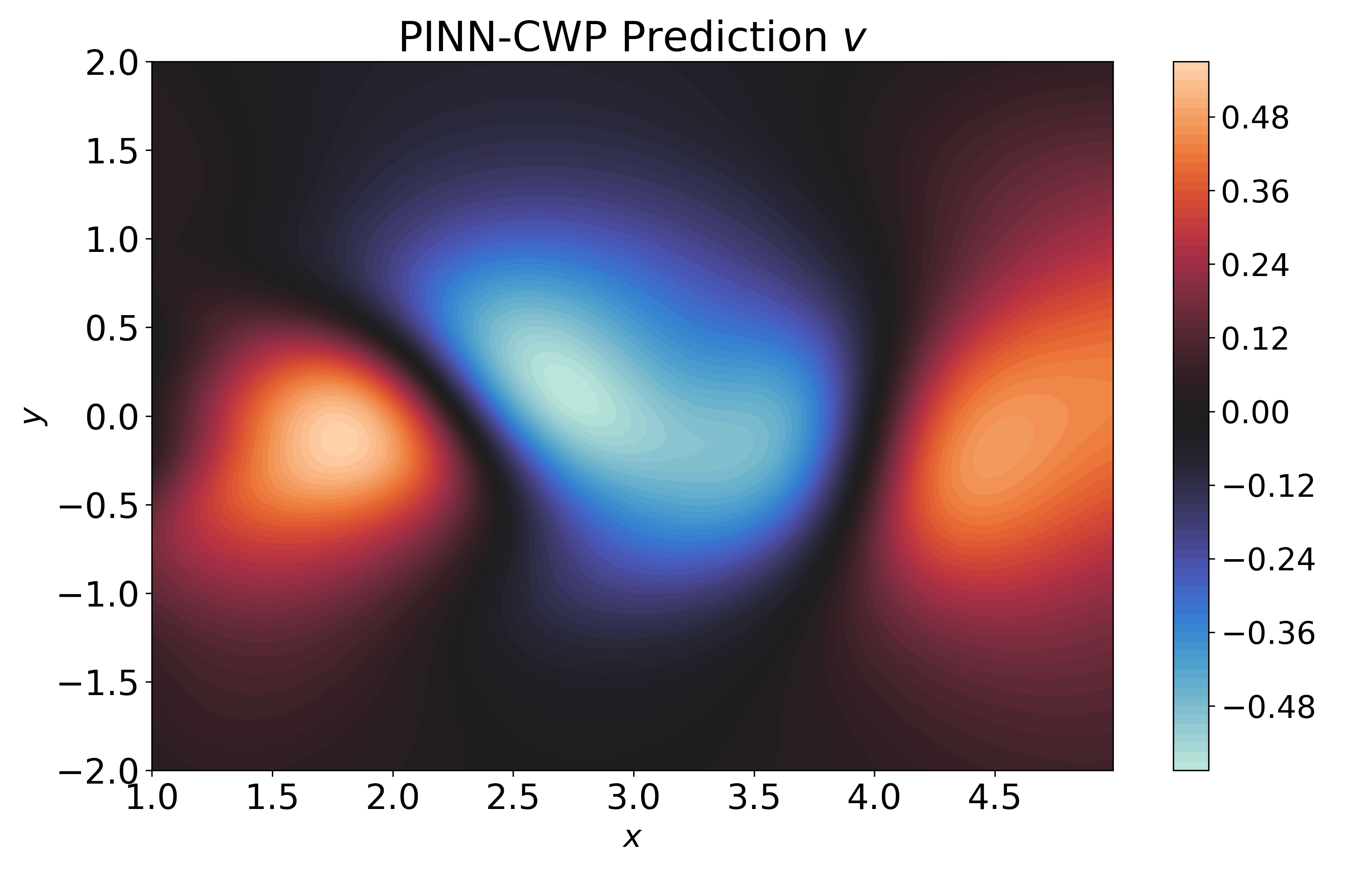} 
   \end{minipage}
   \begin{minipage}{0.31\textwidth}
     \centering
     \includegraphics[width=\linewidth]{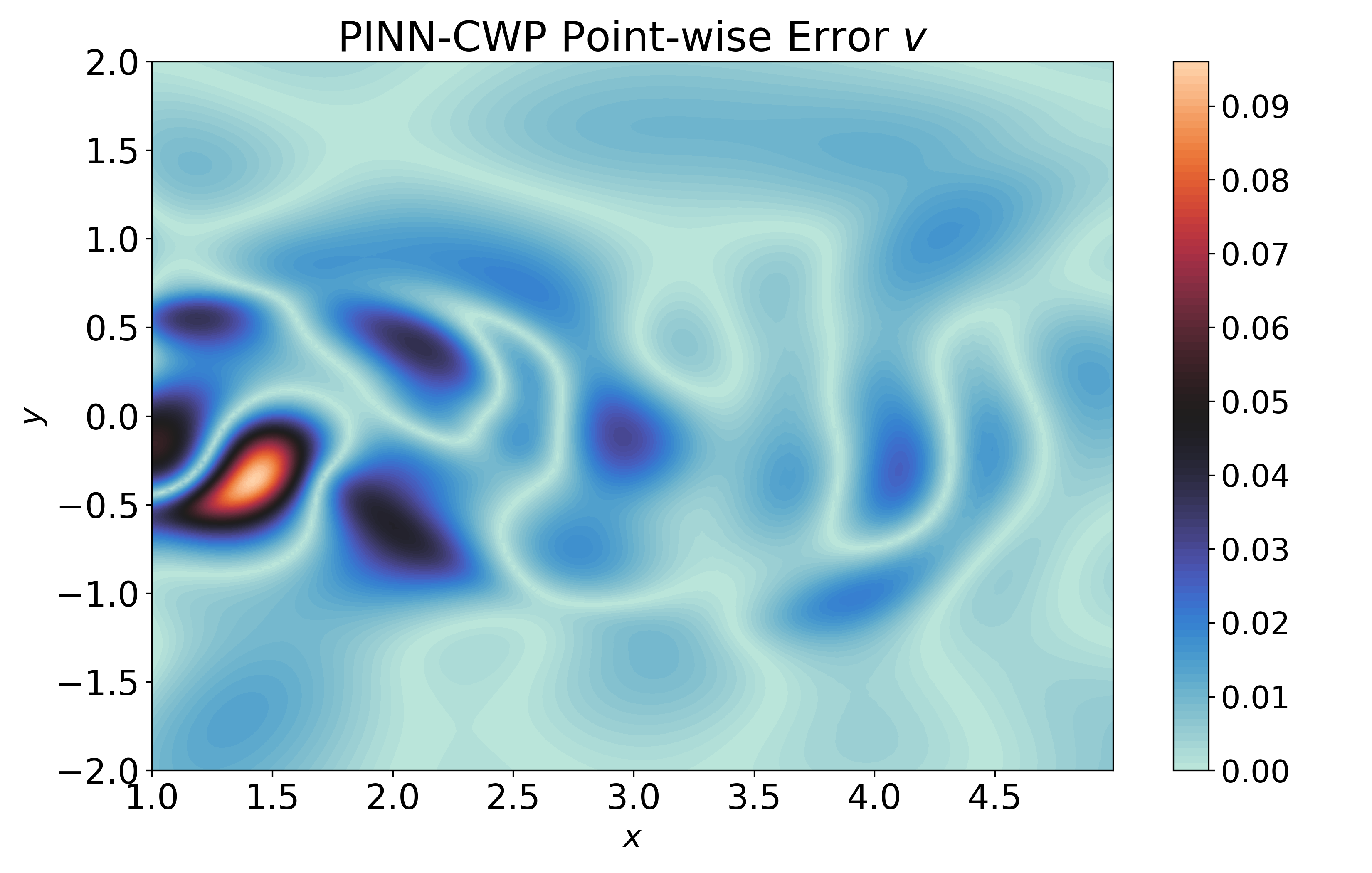} 
   \end{minipage}

   \begin{minipage}{0.31\textwidth}
     \centering
     \includegraphics[width=\linewidth]{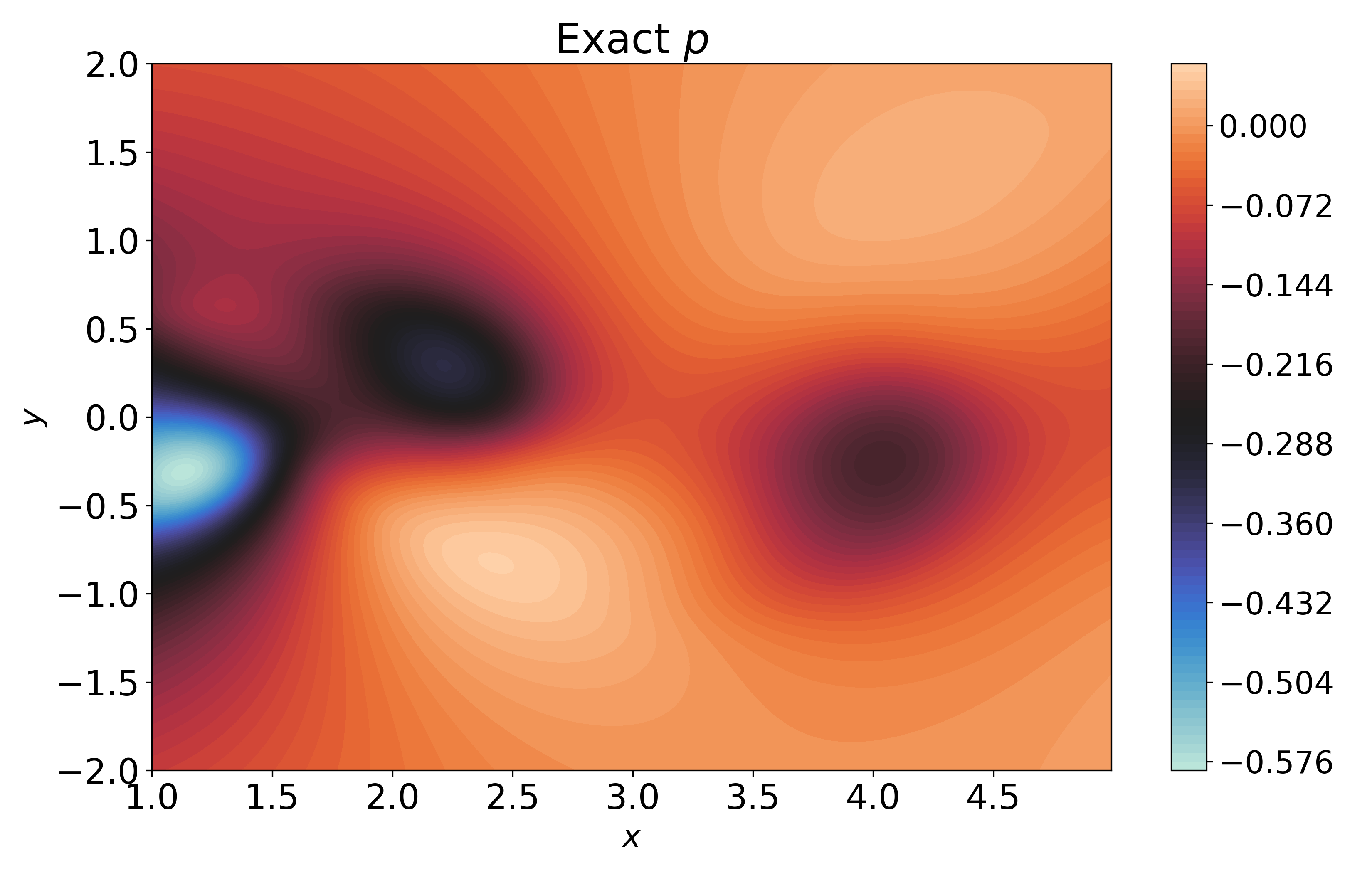} 
   \end{minipage}
   \begin{minipage}{0.31\textwidth}
     \centering
     \includegraphics[width=\linewidth]{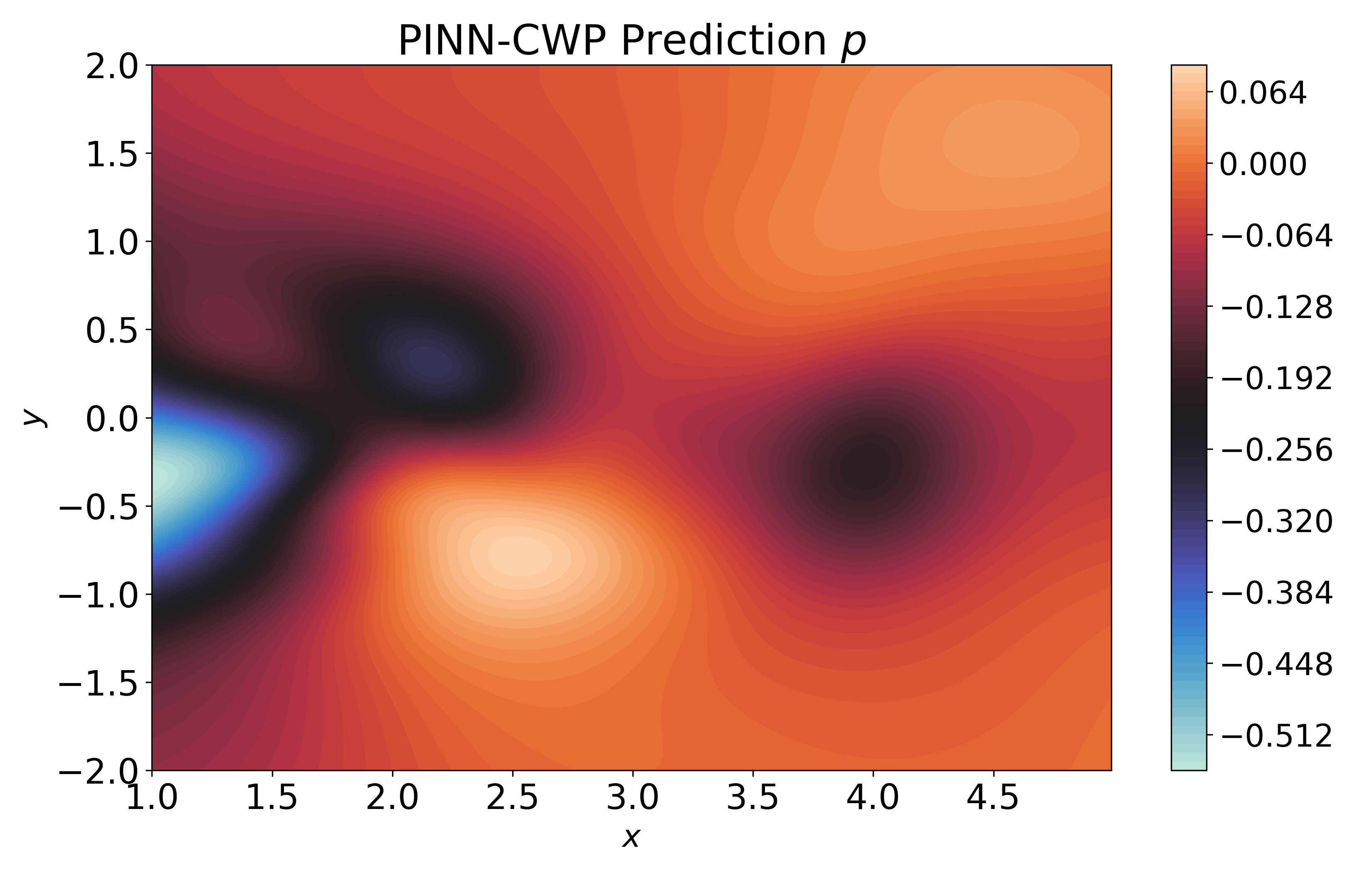} 
   \end{minipage}
   \begin{minipage}{0.31\textwidth}
     \centering
     \includegraphics[width=\linewidth]{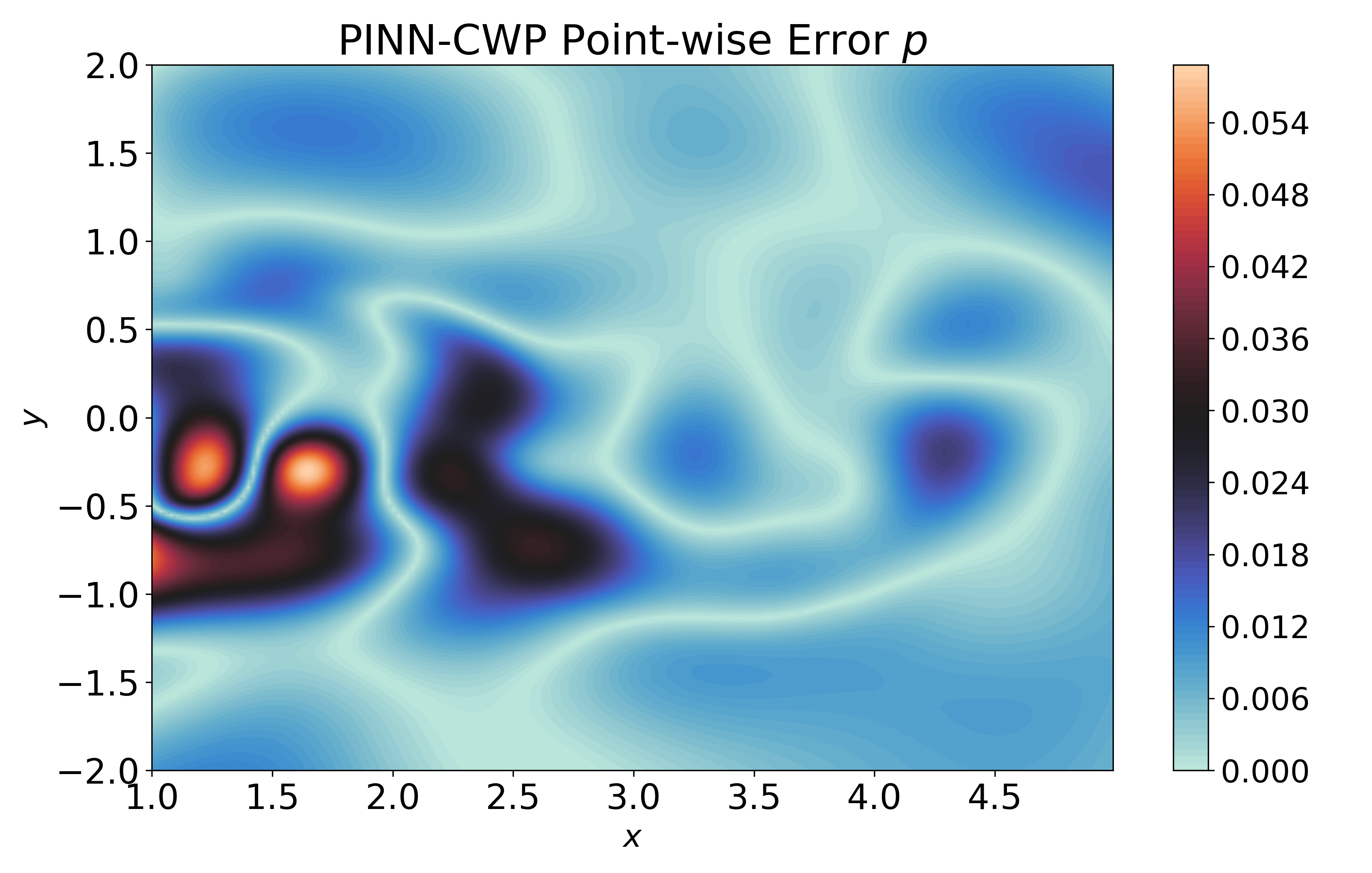} 
   \end{minipage}

    \caption{Heatmaps at the final time slice $t = 42.93$s of the exact solution (left column), predicted solution (middle column, obtained via CWP), and point-wise error (right column) for $u(x,y)$ (top row), $v(x,y)$ (middle row), and $p(x,y)$ (bottom row) for the unsteady NS equation.}\label{fig: NS Heatmap}
\end{figure}

To evaluate performance, we randomly select 90,000 validation points from the full computational domain (not limited to sensor locations) provided in \cite{Shengfenggithub}, ensuring we assess generalization beyond training data. As shown in Fig. \ref{fig: NS Heatmap}, CWP allows PINN to achieve accurate predictions for the high-Reynolds-number NS equations at the final time slice ($t=42.93$s).

We further experiment with the other three algorithms under the same setting. The numerical results for the comparison are listed in Table \ref{table: NS}. Consistently, CWP outperforms the other methods across all metrics. For the velocity components ($u$ and $v$), CWP reduces errors by nearly $50\%$ compared to RBA, SA, and LA. Notably, pressure $p$ errors remain higher than velocity errors across all methods, a known challenge in PINNs due to pressure’s indirect coupling to momentum in incompressible flows \cite{raissi2019physics,eivazi2022physics,jin2021nsfnets}. However, CWP still achieves a significant reduction in pressure error relative to the other three methods. These comparisons highlight the effectiveness of our approach in addressing the unique challenges of high-Reynolds-number flows.

\begin{table}[!h]
    \caption{Relative $L^2$ error of various algorithms on the 2D unsteady NS equation after 20,000 training iterations. }
    \label{table: NS}
    \centering
    \begin{tabular}{|c|c|c|c|c|c|c|}
        \hline
        & CWP & RBA & SA & LA \\ \hline
        x-velocity $u$ & \bm{$2.06\times 10^{-2}$} & $4.31\times 10^{-2}$ & $5.84\times 10^{-2}$ & $5.01\times 10^{-2}$ \\ \hline
        y-velocity $v$ & \bm{$5.64\times 10^{-2}$} & $7.10\times 10^{-2}$ & $8.31\times 10^{-2}$ & $7.66\times 10^{-2}$ \\ \hline
        pressure $p$ & \bm{$8.15\times 10^{-2}$} & $1.15\times 10^{-1}$ & $1.01\times 10^{-1}$ & $1.07\times 10^{-1}$ \\ \hline
        velocity $\sqrt{u^2 + v^2}$ & \bm{$2.29\times 10^{-2}$} & $5.25\times 10^{-2}$ & $5.74\times 10^{-2}$ & $4.93\times 10^{-2}$ \\ \hline
    \end{tabular}
\end{table}

\subsection{Inverse problem: Poisson equation}
In this section, we evaluate the performance of our CWP method on an inverse problem, specifically targeting the reconstruction of spatially varying coefficients in a 2D Poisson equation. We follow the benchmark setup proposed in~\cite{zhongkai2024pinnacle}. The governing PDE is given by:
\begin{align}
-\nabla (a\nabla u) = f, \quad (x, y)\in\Omega,
\end{align}
where $\Omega = [0,1]^2$, and the solution is prescribed as $u(x, y) = \sin(\pi x) \sin(\pi y)$. The source term $f$ is derived accordingly:
\begin{align}
f = \frac{2\pi^2\sin(\pi x)\sin(\pi y)}{1 + x^2 + y^2 + (x-1)^2 + (y-1)^2} + \frac{2\pi ((2x-1)\cos(\pi x)\sin(\pi y)+(2y-1)\cos(\pi y)\sin(\pi x))}{(1 + x^2 + y^2 + (x-1)^2 + (y-1)^2)^2}.
\end{align}

The objective is to infer the unknown diffusion coefficient $a(x, y)$ from the given solution $u(x, y)$ and source $f(x, y)$. The ground truth for the coefficient is:
\begin{align}
a(x,y) = \frac{1}{1 + x^2 + y^2 + (x-1)^2 + (y-1)^2}.
\end{align}

Moreover, as indicated in Ref.~\cite{zhongkai2024pinnacle}, enforcing the boundary condition for \( a(x, y) \) is essential to ensure the uniqueness of the inverse solution. The boundary condition is prescribed as:
\begin{align}
a(x,y) = \frac{1}{1 + x^2 + y^2 + (x-1)^2 + (y-1)^2},\quad (x,y)\in\partial\Omega.
\end{align}

In this experiment, we configure all models with \( N_f = 100 \) collocation points and \( N_{\text{obs}} = 60 \) observation points for $u$. Additionally, for each boundary of \( a(x, y) \), we set \( N_b = 10 \) boundary points. To simulate realistic conditions, Gaussian noise with a variance of \( 0.01 \) is added to the observed data \( u_{\text{obs}} \). We set \(\lambda_{\text{obs}} = \lambda_B = 10\).

The neural network architecture consists of a 4-layer fully connected network with 50 neurons in each hidden layer. Since the diffusion coefficient \( a(x, y) \) is spatially varying and not a constant, we introduce a separate neural network to approximate \( a(x, y) \), alongside the main network used for solving \( u(x, y) \). 
All networks are trained using the Adam optimizer with an initial learning rate of \( 0.01 \), controlled by an exponential decay scheduler with a decay rate of \( 0.85 \) every 1,500 iterations. We report the predicted solution $u(x,y)$ and the reconstruction results of \( a(x, y) \) for each method after 20,000 training iterations.

\begin{figure}[!htb]
\centering
    \begin{minipage}{0.45\textwidth}
     \centering
     \includegraphics[width=\linewidth]{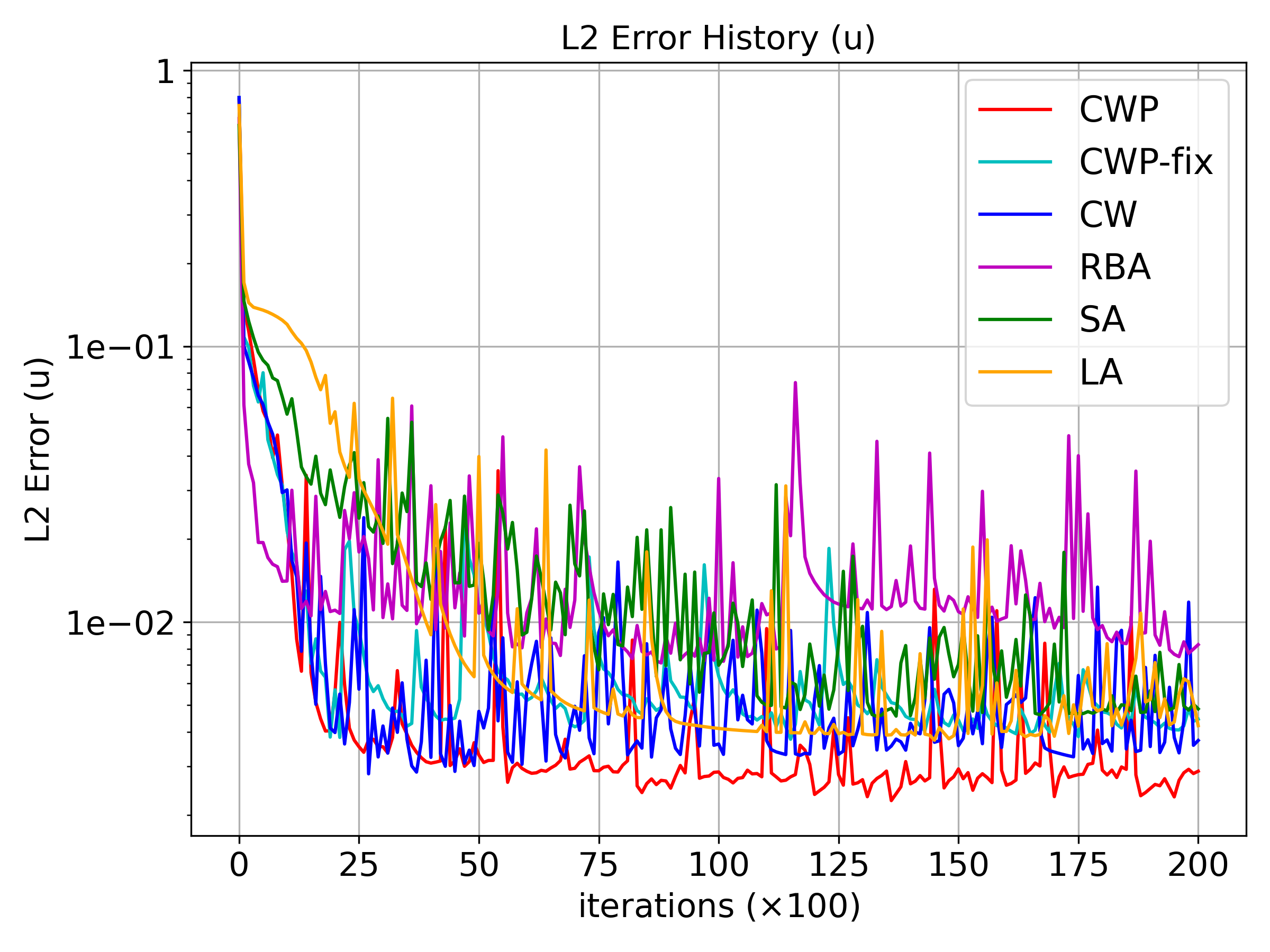} 
   \end{minipage}
   \begin{minipage}{0.45\textwidth}
     \centering
     \includegraphics[width=\linewidth]{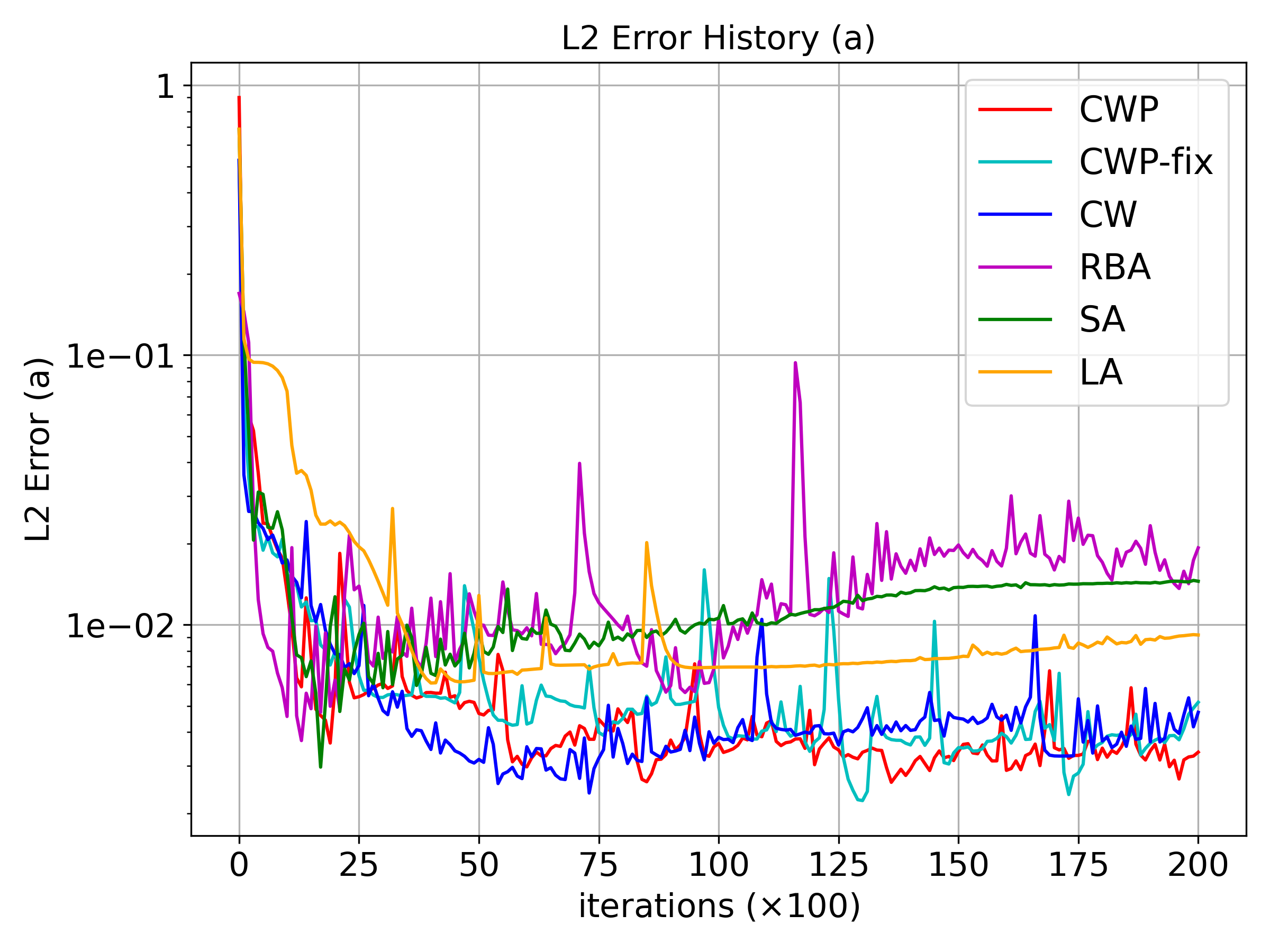} 
   \end{minipage}
    \caption{Error history for learning $u$ (left) and reconstructing $a$ (right) using different weighting algorithms for the inverse problem of the 2D Poisson equation. }\label{fig: Pinv HIST}
\end{figure}
\begin{figure}[!htb]
\centering
    \begin{minipage}{0.33\textwidth}
     \centering
     \includegraphics[width=\linewidth]{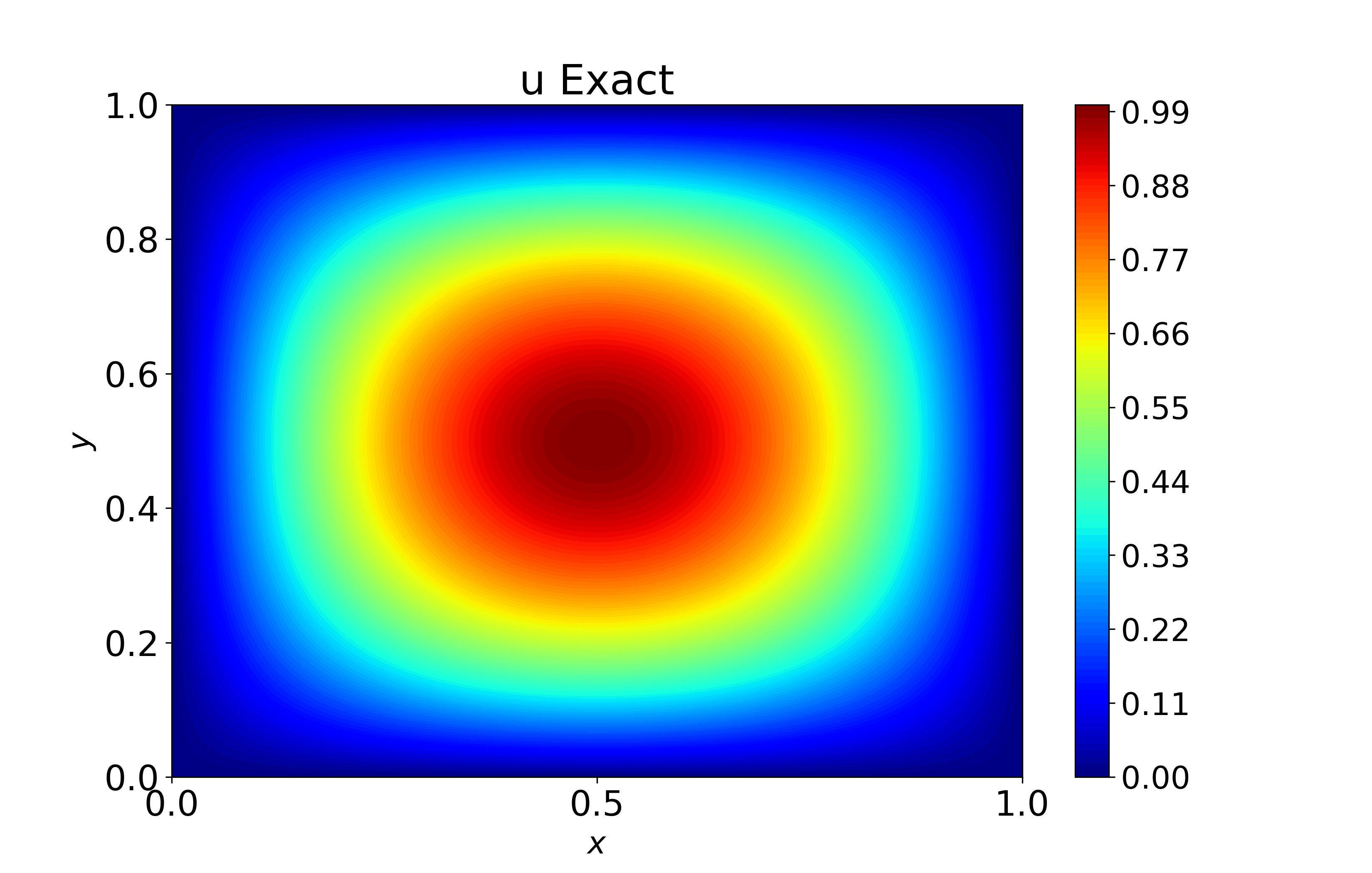} 
   \end{minipage}
   \begin{minipage}{0.33\textwidth}
     \centering
     \includegraphics[width=\linewidth]{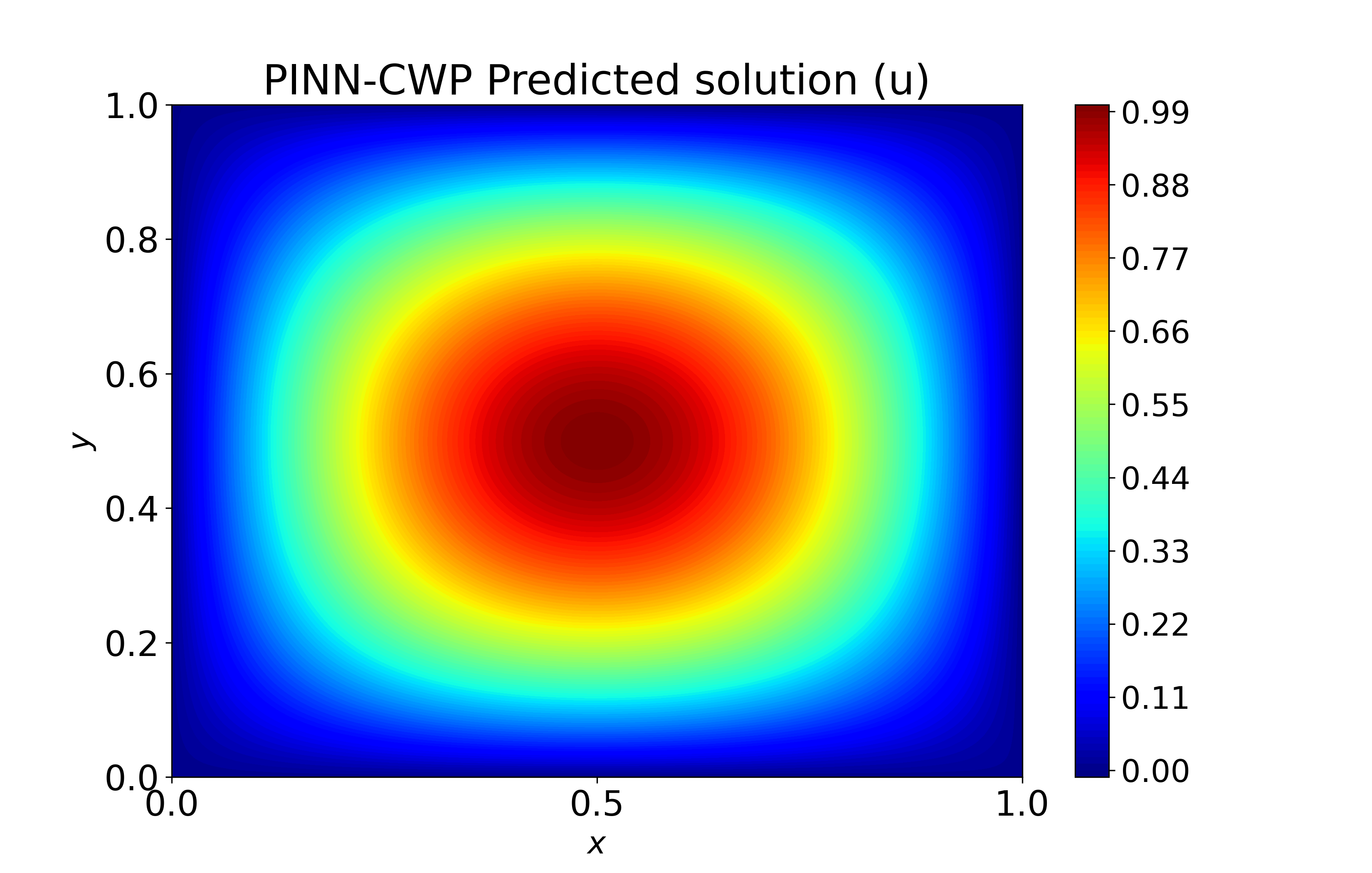} 
   \end{minipage}
   \begin{minipage}{0.33\textwidth}
     \centering
     \includegraphics[width=\linewidth]{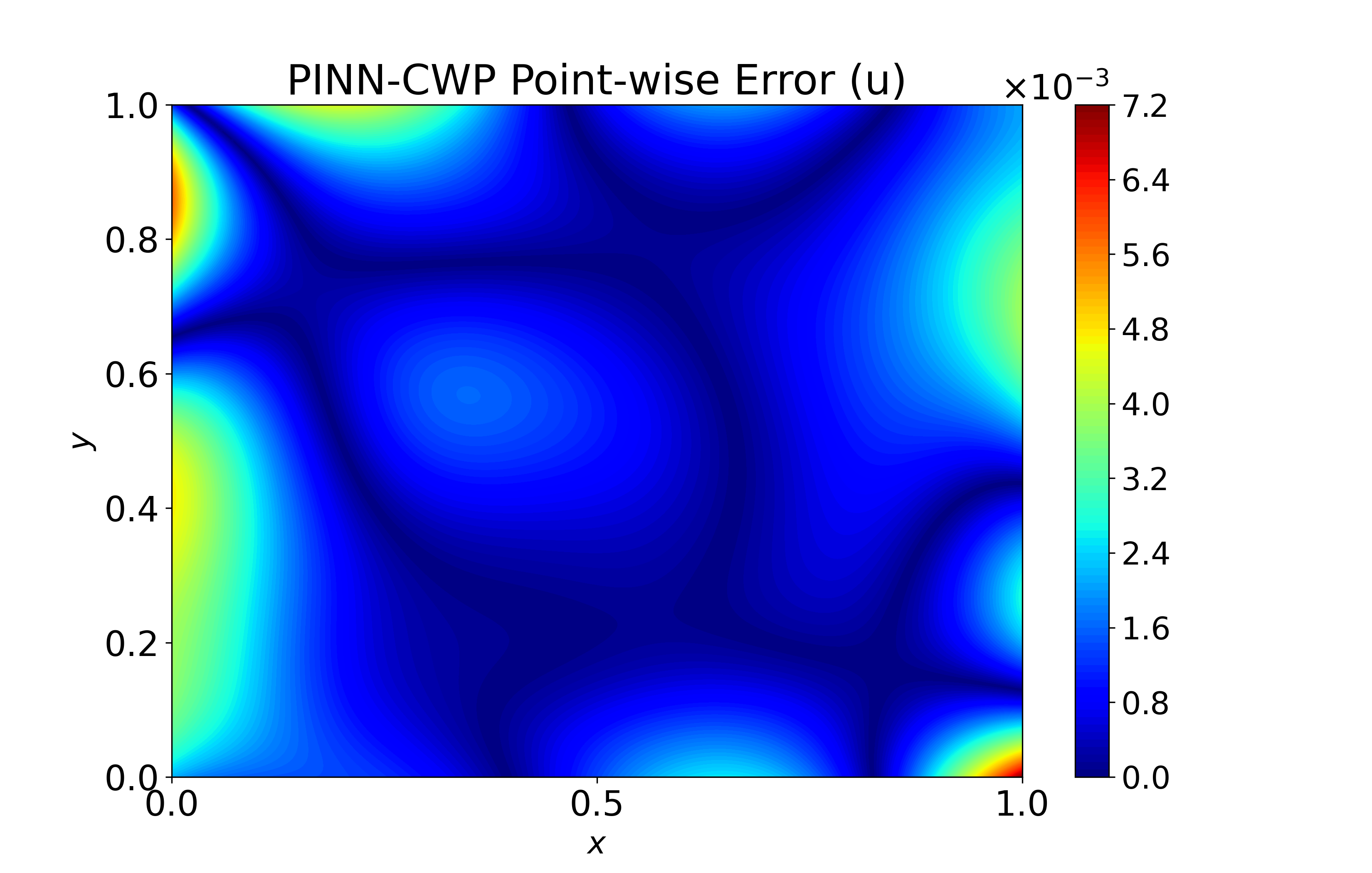} 
   \end{minipage}

   \begin{minipage}{0.33\textwidth}
     \centering
     \includegraphics[width=\linewidth]{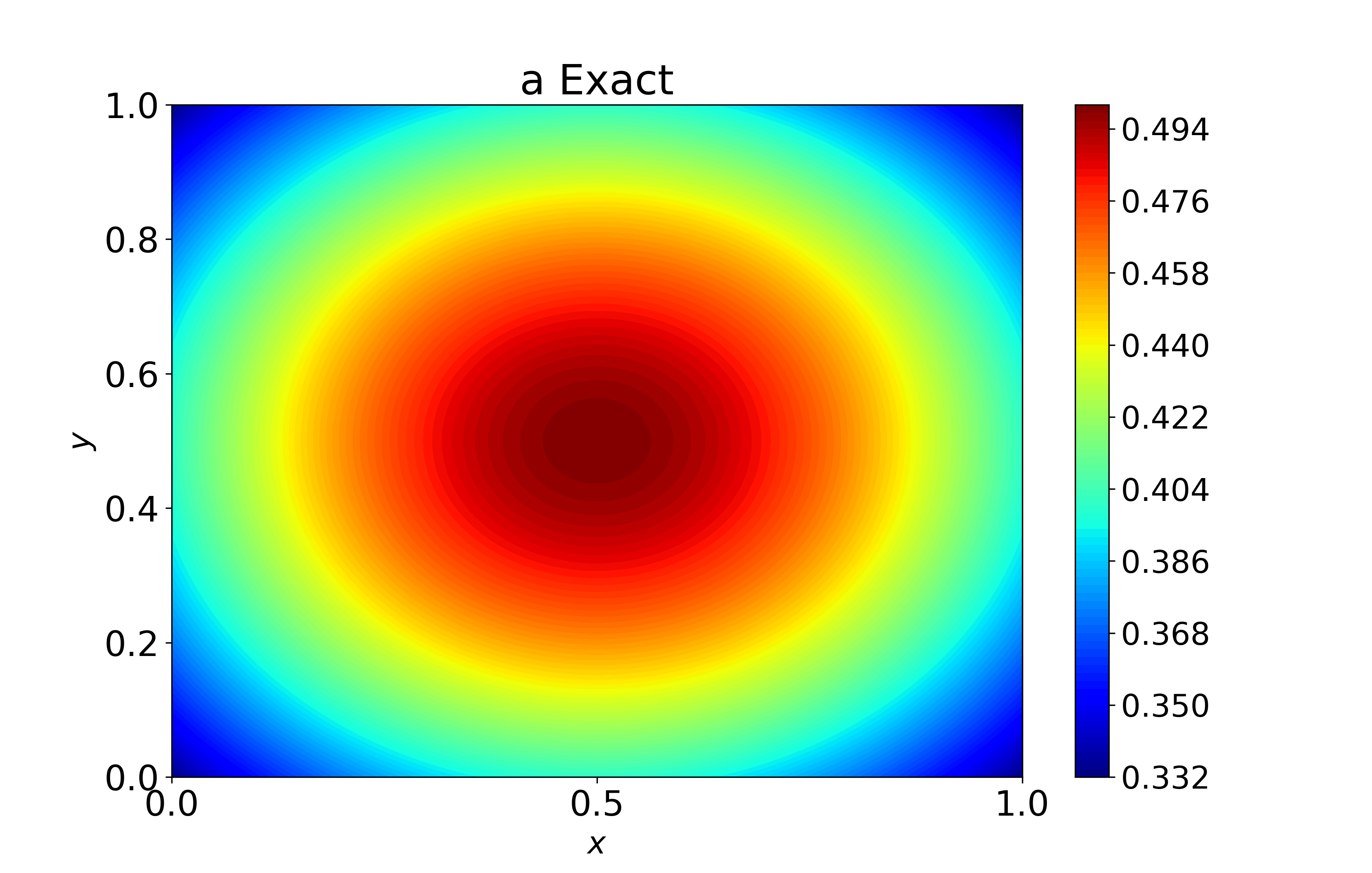} 
   \end{minipage}
   \begin{minipage}{0.33\textwidth}
     \centering
     \includegraphics[width=\linewidth]{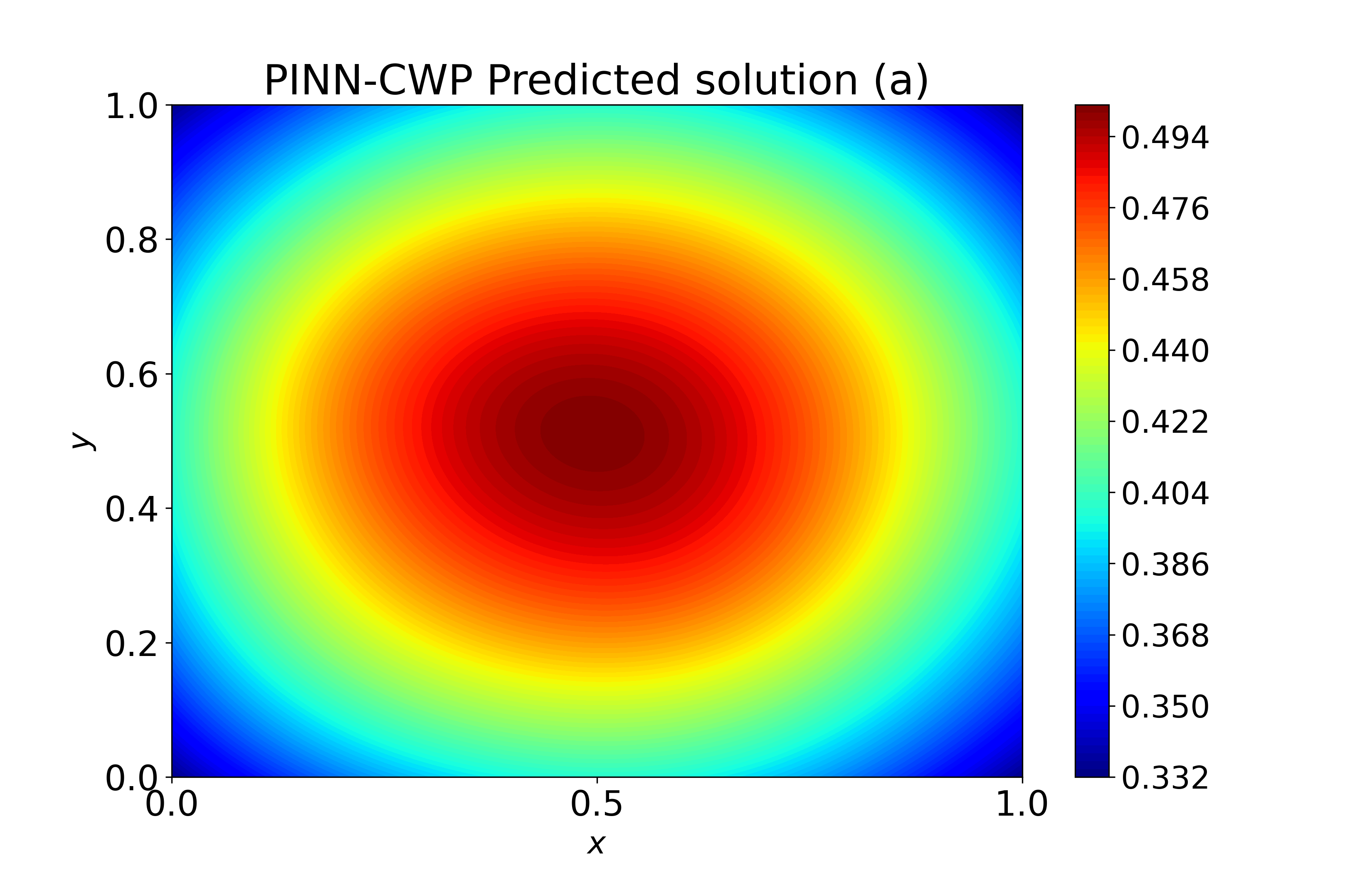} 
   \end{minipage}
   \begin{minipage}{0.33\textwidth}
     \centering
     \includegraphics[width=\linewidth]{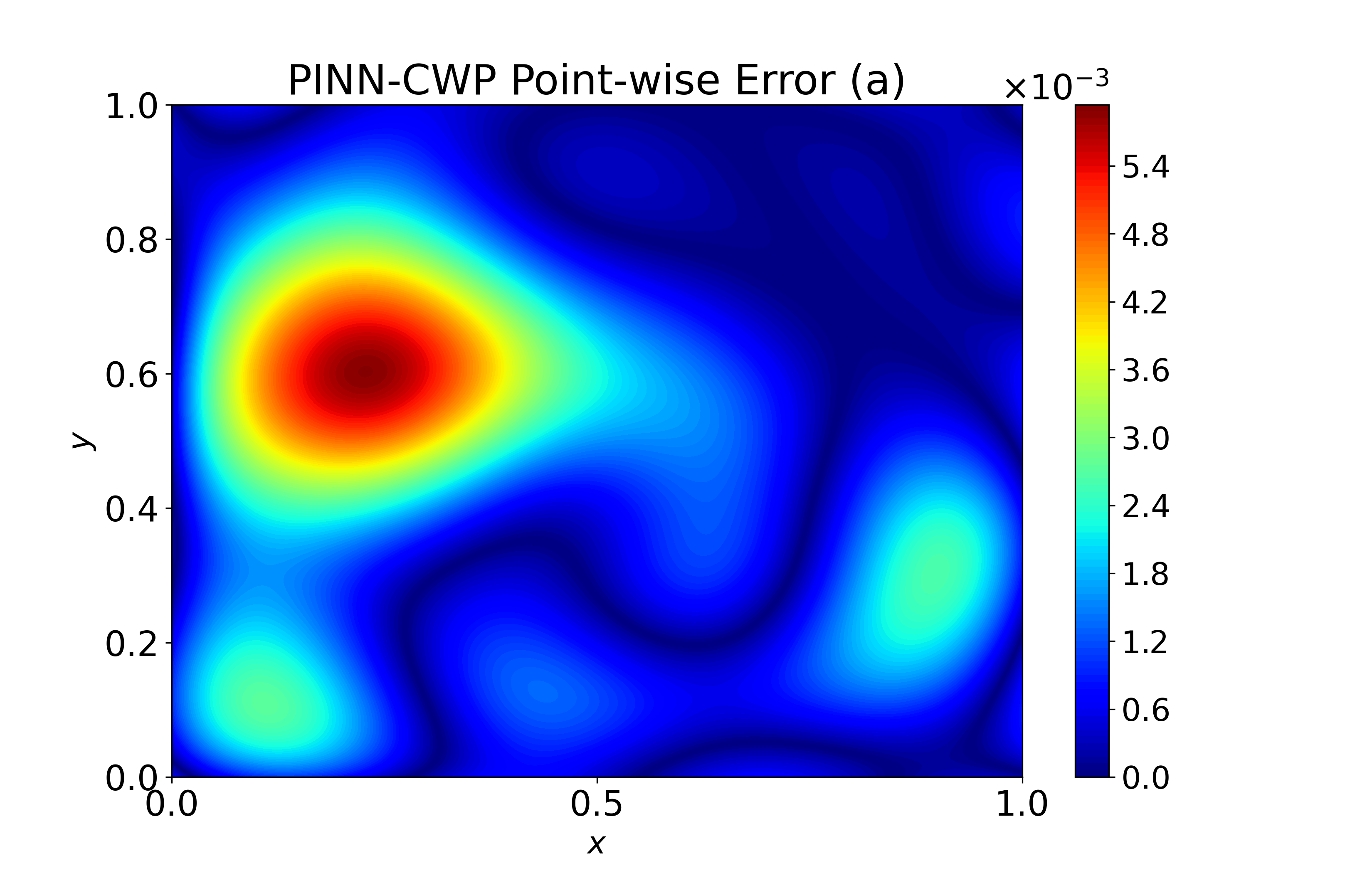} 
   \end{minipage}
    \caption{Heatmaps of the exact solution (left column), predicted solution (middle column), and point-wise error (right column) for $u(x,y)$ (top row) and $a(x,y)$ (bottom row).}\label{fig: Pinv Heatmap}
\end{figure}

The relative \( L^2 \) error history is shown in Fig.~\ref{fig: Pinv HIST}. While CWP achieves the lowest relative \( L^2 \) error in learning the solution \( u(x, y) \), all algorithms are able to reduce the error below \( 1 \times 10^{-2} \). However, in reconstructing the diffusion coefficient \( a(x, y) \), CWP and its variant demonstrate substantially higher accuracy compared to the baseline methods, highlighting their advantages in inverse problem settings. Visualizations of the predicted solution \( u(x, y) \) and the reconstructed coefficient \( a(x, y) \) obtained using CWP are provided in Fig.~\ref{fig: Pinv Heatmap}.


\label{Sec Poisson}

\section{Discussion}
\label{Discussion}
In this work, we propose a convolution-weighting method for Physics-Informed Neural Networks (CWP-PINN) from a primal-dual optimization perspective. By applying convolutional techniques to weight residual losses, CWP-PINN leverages the spatiotemporal smoothness of physical systems, overcoming the limitation of traditional PINNs that focus on isolated collocation points. The method demonstrates distinct strengths: it significantly improves accuracy compared to state-of-the-art adaptive weighting methods, particularly in scenarios with limited training data or stiff PDEs involving shocks or high-frequency dynamics. Additionally, CWP-PINN maintains computational efficiency by avoiding extra backward computations, ensuring training costs remain comparable to conventional approaches. Its integration of convolutional regularization and adaptive resampling enables robust handling of complex spatiotemporal features and enhances performance in inverse problem settings, making it a powerful framework for data-constrained scientific machine learning.

\section*{CRediT authorship contribution statement}
\textbf{Chenhao Si:} Conceptualization, Methodology, Investigation, Software, Writing – original draft
\textbf{Ming Yan:} Conceptualization, Methodology, Supervision, Writing – review and editing

\section*{Declaration of competing interest}
The authors declare that they have no known competing financial interests or personal relationships that could have appeared to influence the work reported in this paper.

\section*{Data availability}
The codes generated during the current study will be available upon reasonable request after the article is published.

\section*{Acknowledgments}
This work was partially supported by the National Natural Science Foundation of China (72495131, 82441027), Guangdong Provincial Key Laboratory of Mathematical Foundations for Artificial Intelligence (2023B1212010001), Shenzhen Stability Science Program, and the Shenzhen Science and Technology Program under grant number ZDSYS20211021111415025.

\bibliographystyle{unsrt}  
\bibliography{ref}

\begin{thebibliography}{10}

\bibitem{raissi2019physics}
Maziar Raissi, Paris Perdikaris, and George~E Karniadakis.
\newblock Physics-informed neural networks: A deep learning framework for solving forward and inverse problems involving nonlinear partial differential equations.
\newblock {\em Journal of Computational physics}, 378:686--707, 2019.

\bibitem{karniadakis2021physics}
George~Em Karniadakis, Ioannis~G Kevrekidis, Lu~Lu, Paris Perdikaris, Sifan Wang, and Liu Yang.
\newblock Physics-informed machine learning.
\newblock {\em Nature Reviews Physics}, 3(6):422--440, 2021.

\bibitem{xu2023physics}
Jiaxuan Xu, Han Wei, and Hua Bao.
\newblock Physics-informed neural networks for studying heat transfer in porous media.
\newblock {\em International Journal of Heat and Mass Transfer}, 217:124671, 2023.

\bibitem{cai2021physics}
Shengze Cai, Zhicheng Wang, Sifan Wang, Paris Perdikaris, and George~Em Karniadakis.
\newblock Physics-informed neural networks for heat transfer problems.
\newblock {\em Journal of Heat Transfer}, 143(6):060801, 2021.

\bibitem{majumdar2025hxpinn}
Ritam Majumdar, Vishal Jadhav, Anirudh Deodhar, Shirish Karande, Lovekesh Vig, and Venkataramana Runkana.
\newblock Hxpinn: A hypernetwork-based physics-informed neural network for real-time monitoring of an industrial heat exchanger.
\newblock {\em Numerical Heat Transfer, Part B: Fundamentals}, 86(6):1910--1931, 2025.

\bibitem{hu2024physics}
Haoteng Hu, Lehua Qi, and Xujiang Chao.
\newblock Physics-informed neural networks ({PINN}) for computational solid mechanics: Numerical frameworks and applications.
\newblock {\em Thin-Walled Structures}, page 112495, 2024.

\bibitem{faroughi2024physics}
Salah~A Faroughi, Nikhil~M Pawar, C{\'e}lio Fernandes, Maziar Raissi, Subasish Das, Nima~K Kalantari, and Seyed Kourosh~Mahjour.
\newblock Physics-guided, physics-informed, and physics-encoded neural networks and operators in scientific computing: {F}luid and solid mechanics.
\newblock {\em Journal of Computing and Information Science in Engineering}, 24(4):040802, 2024.

\bibitem{zhang2020learning}
Dongkun Zhang, Ling Guo, and George~Em Karniadakis.
\newblock Learning in modal space: Solving time-dependent stochastic {PDE}s using physics-informed neural networks.
\newblock {\em SIAM Journal on Scientific Computing}, 42(2):A639--A665, 2020.

\bibitem{chen2021solving}
Xiaoli Chen, Liu Yang, Jinqiao Duan, and George~Em Karniadakis.
\newblock Solving inverse stochastic problems from discrete particle observations using the {F}okker--{P}lanck equation and physics-informed neural networks.
\newblock {\em SIAM Journal on Scientific Computing}, 43(3):B811--B830, 2021.

\bibitem{yang2019adversarial}
Yibo Yang and Paris Perdikaris.
\newblock Adversarial uncertainty quantification in physics-informed neural networks.
\newblock {\em Journal of Computational Physics}, 394:136--152, 2019.

\bibitem{zhang2019quantifying}
Dongkun Zhang, Lu~Lu, Ling Guo, and George~Em Karniadakis.
\newblock Quantifying total uncertainty in physics-informed neural networks for solving forward and inverse stochastic problems.
\newblock {\em Journal of Computational Physics}, 397:108850, 2019.

\bibitem{yang2021b}
Liu Yang, Xuhui Meng, and George~Em Karniadakis.
\newblock B-{PINN}s: {B}ayesian physics-informed neural networks for forward and inverse {PDE} problems with noisy data.
\newblock {\em Journal of Computational Physics}, 425:109913, 2021.

\bibitem{tan2024physics}
Kaiyuan Tan, Peilun Li, and Thomas Beckers.
\newblock Physics-constrained learning of {PDE} systems with uncertainty quantified port-hamiltonian models.
\newblock In {\em 6th Annual Learning for Dynamics \& Control Conference}, pages 1753--1764. PMLR, 2024.

\bibitem{tan2025plug}
Kaiyuan Tan, Peilun Li, Jun Wang, and Thomas Beckers.
\newblock Plug-and-play physics-informed learning using uncertainty quantified port-hamiltonian models.
\newblock {\em arXiv preprint arXiv:2504.17966}, 2025.

\bibitem{zhao2023pinnsformer}
Zhiyuan Zhao, Xueying Ding, and B~Aditya Prakash.
\newblock Pinnsformer: A transformer-based framework for physics-informed neural networks.
\newblock {\em arXiv preprint arXiv:2307.11833}, 2023.

\bibitem{huang2024diffusionpde}
Jiahe Huang, Guandao Yang, Zichen Wang, and Jeong~Joon Park.
\newblock Diffusion{PDE}: Generative {PDE}-solving under partial observation.
\newblock {\em arXiv preprint arXiv:2406.17763}, 2024.

\bibitem{bastek2024physics}
Jan-Hendrik Bastek, WaiChing Sun, and Dennis~M Kochmann.
\newblock Physics-informed diffusion models.
\newblock {\em arXiv preprint arXiv:2403.14404}, 2024.

\bibitem{lu2019deeponet}
Lu~Lu, Pengzhan Jin, and George~Em Karniadakis.
\newblock Deeponet: Learning nonlinear operators for identifying differential equations based on the universal approximation theorem of operators.
\newblock {\em arXiv preprint arXiv:1910.03193}, 2019.

\bibitem{xu2025velocity}
Ruichen Xu, Zongyu Wu, Luoyao Chen, Georgios Kementzidis, Siyao Wang, Haochun Wang, Yiwei Shi, and Yuefan Deng.
\newblock Velocity-inferred hamiltonian neural networks: Learning energy-conserving dynamics from position-only data.
\newblock {\em arXiv preprint arXiv:2505.02321}, 2025.

\bibitem{gao2024coordinate}
Wenhan Gao, Ruichen Xu, Hong Wang, and Yi~Liu.
\newblock Coordinate transform fourier neural operators for symmetries in physical modelings.
\newblock {\em Transactions on Machine Learning Research}, 2024.

\bibitem{cao2024laplace}
Qianying Cao, Somdatta Goswami, and George~Em Karniadakis.
\newblock Laplace neural operator for solving differential equations.
\newblock {\em Nature Machine Intelligence}, 6(6):631--640, 2024.

\bibitem{wang2021understanding}
Sifan Wang, Yujun Teng, and Paris Perdikaris.
\newblock Understanding and mitigating gradient flow pathologies in physics-informed neural networks.
\newblock {\em SIAM Journal on Scientific Computing}, 43(5):A3055--A3081, 2021.

\bibitem{RBA-PINN}
Sokratis~J Anagnostopoulos, Juan~Diego Toscano, Nikolaos Stergiopulos, and George~Em Karniadakis.
\newblock Residual-based attention in physics-informed neural networks.
\newblock {\em Computer Methods in Applied Mechanics and Engineering}, 421:116805, 2024.

\bibitem{abbasi2025challenges}
Jassem Abbasi, Ameya~D Jagtap, Ben Moseley, Aksel Hiorth, and P{\aa}l~{\O}steb{\o} Andersen.
\newblock Challenges and advancements in modeling shock fronts with physics-informed neural networks: A review and benchmarking study.
\newblock {\em arXiv preprint arXiv:2503.17379}, 2025.

\bibitem{liu2024discontinuity}
Li~Liu, Shengping Liu, Hui Xie, Fansheng Xiong, Tengchao Yu, Mengjuan Xiao, Lufeng Liu, and Heng Yong.
\newblock Discontinuity computing using physics-informed neural networks.
\newblock {\em Journal of Scientific Computing}, 98(1):22, 2024.

\bibitem{arzani2023theory}
Amirhossein Arzani, Kevin~W Cassel, and Roshan~M D'Souza.
\newblock Theory-guided physics-informed neural networks for boundary layer problems with singular perturbation.
\newblock {\em Journal of Computational Physics}, 473:111768, 2023.

\bibitem{bararnia2022application}
Hassan Bararnia and Mehdi Esmaeilpour.
\newblock On the application of physics informed neural networks ({PINN}) to solve boundary layer thermal-fluid problems.
\newblock {\em International Communications in Heat and Mass Transfer}, 132:105890, 2022.

\bibitem{lu2021deepxde}
Lu~Lu, Xuhui Meng, Zhiping Mao, and George~Em Karniadakis.
\newblock Deep{XDE}: {A} deep learning library for solving differential equations.
\newblock {\em SIAM review}, 63(1):208--228, 2021.

\bibitem{wu2023comprehensive}
Chenxi Wu, Min Zhu, Qinyang Tan, Yadhu Kartha, and Lu~Lu.
\newblock A comprehensive study of non-adaptive and residual-based adaptive sampling for physics-informed neural networks.
\newblock {\em Computer Methods in Applied Mechanics and Engineering}, 403:115671, 2023.

\bibitem{gao2023failure}
Zhiwei Gao, Liang Yan, and Tao Zhou.
\newblock Failure-informed adaptive sampling for {PINN}s.
\newblock {\em SIAM Journal on Scientific Computing}, 45(4):A1971--A1994, 2023.

\bibitem{gao2023active}
Wenhan Gao and Chunmei Wang.
\newblock Active learning based sampling for high-dimensional nonlinear partial differential equations.
\newblock {\em Journal of Computational Physics}, 475:111848, 2023.

\bibitem{bischof2025multi}
Rafael Bischof and Michael~A Kraus.
\newblock Multi-objective loss balancing for physics-informed deep learning.
\newblock {\em Computer Methods in Applied Mechanics and Engineering}, 439:117914, 2025.

\bibitem{wang2022NTK}
Sifan Wang, Xinling Yu, and Paris Perdikaris.
\newblock When and why {PINN}s fail to train: {A} neural tangent kernel perspective.
\newblock {\em Journal of Computational Physics}, 449:110768, 2022.

\bibitem{xiang2022self}
Zixue Xiang, Wei Peng, Xu~Liu, and Wen Yao.
\newblock Self-adaptive loss balanced physics-informed neural networks.
\newblock {\em Neurocomputing}, 496:11--34, 2022.

\bibitem{hua2023physics}
Jiaqi Hua, Yingguang Li, Changqing Liu, Peng Wan, and Xu~Liu.
\newblock Physics-informed neural networks with weighted losses by uncertainty evaluation for accurate and stable prediction of manufacturing systems.
\newblock {\em IEEE Transactions on Neural Networks and Learning Systems}, 2023.

\bibitem{song2024loss}
Yanjie Song, He~Wang, He~Yang, Maria~Luisa Taccari, and Xiaohui Chen.
\newblock Loss-attentional physics-informed neural networks.
\newblock {\em Journal of Computational Physics}, 501:112781, 2024.

\bibitem{LUO2025114010}
Jiaqi Luo, Yahong Yang, Yuan Yuan, Shixin Xu, and Wenrui Hao.
\newblock An imbalanced learning-based sampling method for physics-informed neural networks.
\newblock {\em Journal of Computational Physics}, 534:114010, 2025.

\bibitem{lu2021physics}
Lu~Lu, Raphael Pestourie, Wenjie Yao, Zhicheng Wang, Francesc Verdugo, and Steven~G Johnson.
\newblock Physics-informed neural networks with hard constraints for inverse design.
\newblock {\em SIAM Journal on Scientific Computing}, 43(6):B1105--B1132, 2021.

\bibitem{sukumar2022exact}
Natarajan Sukumar and Ankit Srivastava.
\newblock Exact imposition of boundary conditions with distance functions in physics-informed deep neural networks.
\newblock {\em Computer Methods in Applied Mechanics and Engineering}, 389:114333, 2022.

\bibitem{dong2021method}
Suchuan Dong and Naxian Ni.
\newblock A method for representing periodic functions and enforcing exactly periodic boundary conditions with deep neural networks.
\newblock {\em Journal of Computational Physics}, 435:110242, 2021.

\bibitem{SA-PINN}
Levi~D McClenny and Ulisses~M Braga-Neto.
\newblock Self-adaptive physics-informed neural networks.
\newblock {\em Journal of Computational Physics}, 474:111722, 2023.

\bibitem{wu2024ropinn}
Haixu Wu, Huakun Luo, Yuezhou Ma, Jianmin Wang, and Mingsheng Long.
\newblock Ro{PINN}: Region optimized physics-informed neural networks.
\newblock {\em arXiv preprint arXiv:2405.14369}, 2024.

\bibitem{Shengfenggithub}
Xu~Shengfeng.
\newblock \url{https://github.com/Shengfeng233/PINN-for-NS-equation}.

\bibitem{eivazi2022physics}
Hamidreza Eivazi, Mojtaba Tahani, Philipp Schlatter, and Ricardo Vinuesa.
\newblock Physics-informed neural networks for solving {R}eynolds-averaged {Navier-Stokes} equations.
\newblock {\em Physics of Fluids}, 34(7), 2022.

\bibitem{jin2021nsfnets}
Xiaowei Jin, Shengze Cai, Hui Li, and George~Em Karniadakis.
\newblock {NSFnets (Navier-Stokes flow nets): Physics-informed neural networks for the incompressible Navier-Stokes equations}.
\newblock {\em Journal of Computational Physics}, 426:109951, 2021.

\bibitem{zhongkai2024pinnacle}
Hao Zhongkai, Jiachen Yao, Chang Su, Hang Su, Ziao Wang, Fanzhi Lu, Zeyu Xia, Yichi Zhang, Songming Liu, Lu~Lu, et~al.
\newblock Pinnacle: A comprehensive benchmark of physics-informed neural networks for solving {PDEs}.
\newblock {\em Advances in Neural Information Processing Systems}, 37:76721--76774, 2024.

\end{thebibliography}
\end{document}